\documentclass[runningheads]{llncs}

\usepackage{eccv}

\usepackage{eccvabbrv}

\usepackage[cjk]{kotex}
\usepackage[dvipsnames]{xcolor}
\usepackage{colortbl}

\newcommand*{\mb}[1]{\mathbf{#1}}

\def\be#1\ee{\begin{align}#1\end{align}}
\def\bea#1\eea{\begin{eqnarray}#1\end{eqnarray}}
\def\ba#1\ea{\begin{align*}#1\end{align*}}
\def\bs#1\es{\begin{equation}\begin{split}#1\end{split}\end{equation}}
\newcommand{\stimes}{{\times}}

\definecolor{deemph}{gray}{0.6}

\definecolor{baselinecolor}{gray}{.9}
\usepackage[position=top,font=scriptsize]{subcaption}

\usepackage{graphicx}
\usepackage{booktabs}

\usepackage[accsupp]{axessibility}  %

\usepackage[pagebackref,breaklinks,colorlinks,citecolor=eccvblue]{hyperref}

\usepackage{orcidlink}

\usepackage{multirow}
\usepackage{svg}
\usepackage{wrapfig,lipsum,booktabs}
\usepackage{url}
\usepackage{pifont}
\usepackage{enumitem}
\usepackage{amsmath}
\usepackage{lmodern}
\usepackage{makecell}

\newcommand\ours{\text{RDNet}\xspace}
\newcommand\gr{\rowcolor{Gray!30}}

\begin{document}

\title{DenseNets Reloaded:\\Paradigm Shift Beyond ResNets and ViTs} 

\titlerunning{DenseNets Reloaded (RDNet)}

\author{Donghyun Kim\inst{1}\thanks{Equal contribution. Correspondence to Dongyoon Han.} \and Byeongho Heo\inst{2} \and
Dongyoon Han\inst{2}$^\star$}

\authorrunning{Kim et al.}

\institute{$^1$NAVER Cloud AI, $^2$NAVER AI Lab}

\maketitle
\newlist{coloritemize}{itemize}{1}
\setlist[coloritemize]{label=\textcolor{eccvblue}{\textbullet}}
\newcolumntype{C}[1]{>{\centering\let\newline\\\arraybackslash\hspace{0pt}}m{#1}}

\begin{abstract} 
This paper revives Densely Connected Convolutional Networks (DenseNets) and reveals the underrated effectiveness over predominant ResNet-style architectures.
We believe DenseNets' potential was overlooked due to untouched training methods and traditional design elements not fully revealing their capabilities. Our pilot study shows dense connections through concatenation are strong, demonstrating that DenseNets can be revitalized to compete with modern architectures. We methodically refine suboptimal components - architectural adjustments, block redesign, and improved training recipes towards widening DenseNets and boosting memory efficiency while keeping concatenation shortcuts. Our models, employing simple architectural elements, ultimately surpass Swin Transformer, ConvNeXt, and DeiT-III — key architectures in the residual learning lineage. Furthermore, our models exhibit near state-of-the-art performance on ImageNet-1K, competing with the very recent models and downstream tasks, ADE20k semantic segmentation, and COCO object detection/instance segmentation. Finally, we provide empirical analyses that uncover the merits of the concatenation over additive shortcuts, steering a renewed preference towards DenseNet-style designs. Our code is available at \url{https://github.com/naver-ai/rdnet}.
\end{abstract}

\section{Introduction}
\label{sec:intro}

The ``ImageNet moment'' was sparked by the emergence of Convolutional Neural Networks (ConvNets), starting with the milestone AlexNet~\cite{Krizhevsky2012ImageNetCW}. Subsequently, VGG~\cite{simonyan2014very} and GoogleNet~\cite{szegedy2015going} further highlighted the benefits of stacking multiple convolutional layers in ConvNets. In the same era, a monumental architecture ResNet~\cite{resnet} and its family~\cite{preresnet,resnext} stands out for introducing a groundbreaking concept - additive skip connections (also known as additive shortcuts or identity mapping~\cite{preresnet}), which allowed for the stacking of up to 1,000 layers. The introduction of residual learning with it was a game-changer, diminishing the gradient vanishing problem by ensuring the input gradient always remained at one from the derivative of the identity mapping. This innovation sparked a series of successors, including the milestone ConvNets - EfficientNet~\cite{icml2019efficientnet} and ConvNeXt~\cite{cvpr2022convnext}; it paved the way for the next leap, such as Transformers~\cite{vaswani2017attention}, Vision Transformers (ViTs)~\cite{vit}, and Hierarchical ViTs~\cite{liu2021swin}, which accentuates the lasting influence of additive shortcuts.

In the early stage of this period governed by residual learning, Densely Connected Convolutional Networks (DenseNets~\cite{densenet}) introduced a novel approach: maintaining shortcut connections through feature concatenation instead of using additive shortcuts. This led to the concept of feature reuse~\cite{densenet}, allowing more compact models and reducing overfitting through explicit supervision propagation to the early layers. DenseNets showcased efficiency and superior performance in tasks like semantic segmentation~\cite{tiramisu}. The evolution of architectural designs post-DenseNet appeared to challenge the dominance of ResNets but saw a decline in their popularity, shaded by the advantages of additive shortcuts. 

Successors of DenseNets~\cite{wang2018pelee,Lee19vovnet,wang2020cspnet} revisited DenseNets to advance its design spirit but struggled against more dominant architectural trends again. We argue the potential of DenseNets still remained underexplored due to low accessibility, being gradually hindered by outdated training methods and the limitations of low-capacity components; they struggled to keep pace with the advancements in modern architectures that benefited from years of evolutionary refinements. Furthermore, we presume DenseNets requires an overhaul due to its limited applicability and memory challenges caused by increasing feature dimensionality. While the authors addressed memory concerns~\cite{densenet,pleiss2017memoryefficient}, these issues continue to restrict the expansion of the architecture, particularly for width scaling. Despite the drawbacks, we conjecture the core design concept is still highly potent.

Bearing this in mind, this paper revitalizes DenseNets by highlighting the undervalued efficacy of concatenations. Through a comprehensive pilot study training with over 10k random networks across varied setups, we validate our claim that concatenation can surpass the additive shortcut. Afterward, we modernize DenseNet with a more memory-efficient design to widen it, abandoning ineffective components and enhancing architectural and block designs, while preserving the essence of dense connectivity via concatenation. We employ contemporary strategies that synergize with DenseNets as well. 
Our methodology eventually exceeds strong modern architectures~\cite{nips2022hornet,zhu2023biformer,cvmj2023VAN, hassani2023NAT, Weifeng2023smt, anonymous2024moganet} and some milestones like Swin Transformer~\cite{liu2021swin}, ConvNeXt~\cite{cvpr2022convnext}, and DeiT-III~\cite{eccv2022deit3} in performance trade-offs on ImageNet-1K~\cite{imagenet}. Our models demonstrate competitive performance on downstream tasks such as ADE20K semantic segmentation and COCO object detection/instance segmentation. Remarkably, our models do not exhibit slowdown or degradation as the input size increases. %
Ultimately, our empirical analyses shed light on the unique benefits of concatenation.  %

\section{Related Work}
\label{sec:related_works}
\subsubsection{Densely Connected Neural Networks} (DenseNets)~\cite{densenet} pioneered dense connections within Convolutional Neural Networks (ConvNets) beyond additive shortcuts, highlighted by parameter efficiency and enhanced precisions. Building on this framework, several variants were proposed. PeleeNet~\cite{wang2018pelee} successfully proposed modifications %
to achieve real-time inference capabilities upon DenseNet. VovNet~\cite{Lee19vovnet} departed from DenseNets' dense feature reuse in favor of a sparser one-shot aggregation aimed at real-time object detection. CSPNet~\cite{wang2020cspnet}, by omitting features and later concatenating them at cross-stage layers, reduces computational demands, barely affecting precision. 
DenseNets were further highlighted with the effectiveness on dense prediction tasks; for example, Jegou~\etal~\cite{tiramisu} showed the effectiveness of DenseNets on semantic segmentation. MDU-Net~\cite{jiawei2023mdunet} exploited the dense connectivity for enhanced biomedical image segmentation. DCCT~\cite{PARIHAR2023103722} integrated dense connections into a Transformer architecture~\cite{vaswani2017attention} to facilitate image dehazing. 
For video snapshot compressive imaging, EfficientSCI~\cite{Wang2023EfficientSCIDC} also leveraged the benefits of dense connectivity. Wang \etal~\cite{Wang2021SmallObjectDB} utilized dense connections to improve the detection of small objects. %

We believe these references demonstrate DenseNet-based designs' potential, to our knowledge, but none recently challenged the ImageNet benchmarks using the principle of dense connections.

\vspace{-1em}
\subsubsection{Modern architectures.}
DeiT~\cite{icml2021deit} and AugReg~\cite{steiner2021train} exhibited modernized training recipes~\cite{icml2019efficientnet,bello2021revisiting,wightman2021rsb} could replace massive training data for  ViT~\cite{vit} training. Descendant hierarchical ViTs~\cite{liu2021swin,cvpr2022CSWin,nips2021Focal,nips2022focalnet,iccv2021PVT,cvmj2022PVTv2}, which got closer to ConvNets, showed locality offers efficacy along with computational efficiency. Hybrid architectures~\cite{iccv2021coat, nips2021coatnet,eccv2022MaxViT,zhu2023biformer,hassani2023NAT} then explicitly equip convolutions for the locality. Ironically, incomers have become closer to ConvNets, aiming not to forsake the proven effectiveness of simple convolution, albeit using Transformers. 

ConvNets~\cite{resnet,icml2019efficientnet,cvpr2020regnet,brock2021high} initially predominated due to strong capability along with efficiency.
Interestingly, advancements from the ViT side have also contributed to modernizing ConvNets; 
many recent architectures~\cite{han2022learning,2021patchconvnet, trockman2022patches,yu2022metaformer} 
were inspired by ViT's designs but armed with locality, demonstrating the continued high competitiveness of convolutions.  %
Successors like RepLKNet~\cite{cvpr2022replknet} and SLaK~\cite{Liu2022SLak} employed large-scale kernel convolutions built upon the predecessor's legacy to emulate the globality of attention~\cite{vit}, offering to learn enhanced global representations. 
RevCol~\cite{cai2023reversible} introduced a new concept to mix multi-level features repeatedly through multiple columns. 
InceptionNeXt~\cite{yu2023inceptionnext} adopted the inception module~\cite{szegedy2015going} inside ConvNeXt to show improved performance. HorNet~\cite{nips2022hornet} and MogaNet~\cite{anonymous2024moganet} both have presented remarkable performance by employing multiple gated convolution and multi-level features, respectively, which also took advantage of multi-scale features for globality.  %

Those architectures surpassed ViTs on ImageNet and dense prediction tasks as well, but similarities like using additive shortcuts and architectural complexity continue to restrict architectural diversity and innovation. Furthermore, network modernization methods~\cite{wang2018pelee,cvpr2022convnext,bello2021revisiting,wightman2021rsb} have successfully revisited existing architectures but did not handle beyond baselines using additive shortcuts. This work follows a general direction but ensures our starts from a distinct baseline, acknowledging uncertainties about the effectiveness of existing roadmaps.

\section{Methodology: Revitalizing DenseNets}
\label{sec:method}
\vspace{-.5em}
This section starts with our conjecture that DenseNet may not fall behind modern architectures and proves it by substantiating revised DenseNet architectures. Based on our conjecture with our pilot experiments, we propose our methodology, which encompasses some modernized materials to revive DenseNet.

\vspace{-1em}
\subsection{Preliminary}
\label{sec:preliminary}
\subsubsection{Motivation.} 
ResNets~\cite{resnet} have renowned due to the simple formulation at $l$-th layer: $\mb{X_{l+1}}=\mb{X_l}+f(\mb{X_l}\mb{W})$, where the input $\mb{X}_l$ and the weight $\mb{W}$. %
A pivotal element is the \textit{residual connection} (\ie, additive shortcut, +), which facilitates modularized architectural designs as evidenced in Swin~\cite{liu2021swin}, ConvNeXt~\cite{cvpr2022convnext}, and ViT~\cite{vit}. While DenseNets~\cite{densenet} follow the formulation: $\mb{X_{l+1}}=[\mb{X_l}, \ f(\mb{X_l}\mb{W})]$ based on the \textit{dense connection} through concatenation having explicit parameter efficiency. However, this formulation should constrain feature dimensionality due to memory concerns, making it challenging to scale in width.

DenseNets~\cite{densenet} initially outperformed ResNets~\cite{resnet} but failed to realize a complete paradigm shift, losing initial momentum due to their applicability. %
In particular, despite the efforts for memory~\cite{pleiss2017memoryefficient, Lee19vovnet}, width scaling remains problematic for DenseNets, with wider models like DenseNet-161/-233 consuming more memory~\cite{densenet}. %
Nonetheless, inspired by the prior works~\cite{wightman2021rsb, bello2021revisiting,cvpr2022convnext} and motivated by successes in dense prediction tasks such as semantic segmentation~\cite{jiawei2023mdunet}, we believe DenseNets would outperform popular architectures and warrant further exploration of their potential: 1) feature concatenation merits strong capability; 2) the above concerns in DenseNets can be mitigated through a strategic design. %

\vspace{-1em}
\subsubsection{Our conjecture.}\textit{Concatenation shortcut is an effective way of increasing rank.} Consider the layer output $f(\mb{X}\mb{W})$ with the weight $\mb{W}{\in}\mathbb{R}^{d_{in}\stimes d_{out}}$ and the input $\mb{X}{\in}\mathbb{R}^{N \stimes d_{in}}$ with a nonlinearity $f$, where we assume the number of instances $N{\gg}d_{in}$. As focusing on the matrix rank of $f$, $\mathrm{rank}(f(\mb{X}\mb{W}))$ generally gets closer to $d_{out}$ due to the nonlinearity when $d_{in}$ is not that small~\cite{han2021rethinking}. Literature~\cite{han2017deep,chen2017dual, han2021rethinking} manifested the layer $\mb{W}$ with $d_{out}>d_{in}$ offers increased representational capacity. Intriguingly, DenseNets enjoy a similar aspect because we can decompose $\mb{W}=[\mb{W_P}, \mb{I}]$, where $\mb{W_P}$ and $\mb{I}$ denote the weights in the building block and concatenation. We further argue that increasing rank like this frequently would be more beneficial. %
The output dimension of $\mb{W_P}$ is called \textit{growth rate}.%

\textit{A strategic design mitigates memory concerns.}
Consider the output of stacked layers $\mb{X}\mb{W_1}f(\mb{W_2})$, where the weights $\mb{W_1}{\in}\mathbb{R}^{d_{in}\stimes d_{r}}$, $\mb{W_2}{\in}\mathbb{R}^{d_{r}\stimes d_{out}}$, $d_{r}<d_{in}, d_{out}$, and a nonlinearity $f$ after  $\mb{W_2}$. Likewise, the rank is likely preserved as $d_{r}$ is not that small. This suggests that using intermediate dimension reducers like $\mb{W}_1$ (\ie, \textit{transition layer}) may not impact the rank significantly. We argue a frequent application would effectively address memory concerns.

\vspace{-1em}
\subsubsection{Pilot study.} We conduct a pilot study to verify our conjecture by sampling over 15k networks on Tiny-ImageNet~\cite{Le2015TinyIV}, where their shortcuts are either additive like ResNets~\cite{resnet} or concatenation in DenseNets~\cite{densenet}. We carefully control experiments regarding computational costs and involve diverse training setups to ensure a balanced and comprehensive comparison. Intriguingly, concatenation shortcuts all outperform additive ones with the averaged Tiny-ImageNet accuracy - \textbf{56.8±3.9 (concat) vs. 55.9±4.1 (add)}, thereby empirically supporting our claim. Detailed results and setups are provided in \S\ref{sec:potential_concat}.

\vspace{-1em}
\subsection{Revitalizing DenseNets}
\label{sec:method_revital}
We revisit DenseNets while maintaining its core principle via concatenation. Our strategy explores ways to widen DenseNets and identify effective elements. Elements that contribute to the performance improvements are detailed in Table~\ref{tab:densenet_modifications}.

\vspace{-1em}
\subsubsection{Baseline.}
\label{sec:baseline}
As the series of revisiting ResNets~\cite{bello2021revisiting,wightman2021rsb} showed, refined training recipes bring significant improvements. Likewise, we train DenseNet-201 with a modern training setup, establishing it as our baseline. Following the well-explored setups~\cite{icml2021deit,cvpr2022convnext,liu2021swin,icml2021EfficientNetV2,wightman2021rsb}, we include Label Smoothing~\cite{cvpr2016inceptionv3}, RandAugment~\cite{cubuk2020randaugment}, Random Erasing~\cite{zhong2020random,devries2017improved}, Mixup~\cite{zhang2017mixup}, CutMix~\cite{yun2019cutmix}, and Stochastic Depth~\cite{huang2016deep}; we use AdamW~\cite{iclr2019AdamW} with the cosine learning rate schedule~\cite{loshchilov2016sgdr} and linear warmup~\cite{goyal2017accurate} with a popular large epochs training setup (\ie, 300). %

\newcommand\up{\textcolor{WildStrawberry}}
\newcommand\down{\textcolor{Violet}}

\begin{table}[t]
\small
\caption{\textbf{ImageNet-1K performance progressions.} Beginning from the baseline - DenseNet-201~\cite{densenet}, we report every performance change throughout progressions.  We uphold DenseNet's principle of feature reuse through concatenation as the core of the model progression. $+\alpha$ denotes a new element $\alpha$ was added to each prior model. Both enhancements in efficiency or accuracy are colored in \up{red}, while degradations are marked in \down{blue}.; GR denotes the growth rate, the amount of feature concatenation~\cite{densenet}.} %
\label{tab:densenet_modifications}
\vspace{-1em}
\setlength{\tabcolsep}{0.2em}
\resizebox{\linewidth}{!}{
\begin{tabular}{@{}c|l|ccccccc@{}}
\toprule
\multicolumn{1}{c|}{} & \multirow{2}{*}{Elements}   & Top-1                  & Param                & \: FLOPs                & Lat (ms)                   & Lat (ms)            & Lat (ms)                & Mem (GB)      \\
                      &                             & Acc (\%)               &  (M)                  &  \: (G)                 & (b1, Infer)                    & (b128, Infer)            &  (Train)                &  (Train)      \\ \midrule
   (a) & DenseNet-201~\cite{densenet}               & 79.7                 & 20.0                & \: 4.3                &    38.4                       & 190                & 131                    & 3.9  \\
   \midrule
   (b) & (a) + Wider \& shallower                   & 79.5 (\down{$-$0.2}) & 21.8 (\down{+1.8})  & 11.1 (\down{+6.8})    &     \: 8.5 (\up{$-$29.9})                     & 170 (\up{$-$20})    & \: 85 (\up{$-$46})     & 3.2 (\up{$-$0.7}) \\
   (c) & (b) + Modernized blocks                    & 80.4 (\up{+0.9})     & 12.9 (\up{$-$8.9})  & \: 4.8 (\up{$-$6.3})  &     10.4 (\down{+\:\:2.9})                     & 230 (\down{+60})       & 112 (\down{+27})     & 3.4 (\down{+0.2}) \\
   (d) & (c) + Channel dim\big\uparrow (GR\big\downarrow)         & 80.8 (\up{+0.4})     & 19.9 (\down{+7.0})  & \: 4.7 (\up{$-$0.1})  &     11.8 (\down{+\:\:1.4})        & 184 (\up{$-$46})       & \: 88 (\up{$-$24})     & 3.1 (\up{$-$0.3}) \\
   (e) & (d) + Trans. layers\big\uparrow (GR\big\uparrow)        & 82.3 (\up{+1.5})     & 21.2 (\down{+1.3})  & \: 5.0 (\down{+0.3})  &    11.0 (\down{+\:\:0.8})        & 183 (\up{$-$\:\:1})     & \:\:\: 90 (\down{+\:\:2})  \:  & 3.4 (\down{+0.3}) \\
   (f) & (e) + Patchification stem                  & 82.4 (\up{+0.1})          & 21.2 (\up{$-$0.0})         & \: 4.9 (\up{$-$0.1})  &    11.0 (\up{$-$\:\:0.0})                       & 179 (\up{$-$\:\:6})     & \:\:\: 88 (\up{$-$\:\:2})  \:  & 3.2 (\up{$-$0.2}) \\
   (g) & (f) + Refined Trans. layers     & 82.6 (\up{+0.2})     & 22.4 (\down{+1.2})  & \: 4.9 (\down{+0.0})             &    13.6 (\down{+\:\:2.6})                       & 170 (\up{$-$\:\:9})       & \:\:\: 97 (\down{+\:\:9})  \:  & 3.1 (\up{$-$0.1}) \\
   (h) & (g) + Channel re-scaling                 & 82.8 (\up{+0.2})     & 23.9 (\down{+1.5})  & \: 5.0 (\down{+0.1})     &    14.0 (\down{+\:\:0.4})                       &  175 (\down{+\:\:5})       & \:\:\: 99  (\down{+\:\:2})  \:  & 3.1 (\down{+0.0}) \\

\bottomrule
\end{tabular}
}
\vspace{-1.5em}
\end{table}

\vspace{-1em}
\subsubsection{Going wider and shallower.}
DenseNets originally proposed exceedingly deep architectures (\eg, DenseNet-265~\cite{densenet}), which effectively showed the scalability. %
We argue that enhancing feature dimension through a high growth rate (GR) and increasing depth is hardly achieved simultaneously under resource constraints. Prior works~\cite{wang2018pelee,Lee19vovnet,cvpr2020regnet,han2021rethinking,cvpr2022convnext} designed shallower networks to achieve efficiency, particularly latency. Inspired by this, we modify DenseNet to a favorable baseline accordingly; widening the network by augmenting GR while diminishing its depth. Specifically, we vastly increase GR - from 32 to 120 here - to achieve it; we adjust the number of blocks per stage, being reduced from (6, 12, 48, 32) to a much smaller (3, 3, 12, 3) for a depth adjustment. We do not shrink the depth as much to maintain minimal nonlinearity. Table~\ref{tab:densenet_modifications}(b) shows this strategic modification has led to notable latencies and memory efficiency - around 35\% and 18\% decreases in training speed and memory, respectively. The marked increase in GFLOPs to 11.1 will be adjusted through the later elements. %
Further study supports our decision - prioritizing width while balancing depth (see Table~\ref{tab:ablation_depth_vs_wide}).
 
\vspace{-1em}
\subsubsection{Improved feature mixers.} We employ the base block~\cite{cvpr2022convnext} for our feature mixer block, which has been extensively studied to reveal its effectiveness. Before using it, we should reevaluate the studies for our case because 1) DenseNets did not use additive shortcuts, and 2) the building block was originally designed to reduce dimensions successively. We find using the following setups still holds: using 1) Layer Normalization (LN)~\cite{ln} instead of Batch Normalization (BN)~\cite{bn}; 2) post-activation; 3) depthwise convolution~\cite{2017MobileNet} 4) fewer normalizations and activations; 5) a kernel size of 7. A unique aspect of our block is that the output channel (GR) is smaller than the input channel (C); mixed features are eventually more compressed features. As can be seen in Table~\ref{tab:densenet_modifications}(c), our design improves accuracy by a large margin (+0.9\%p) while slightly increasing computational costs. We supplement factor analyses for our study here (see Table~\ref{tab:ablation_featuremixer}).

\begin{figure}[t!]
    \centering
    \includegraphics[width=.84\textwidth]{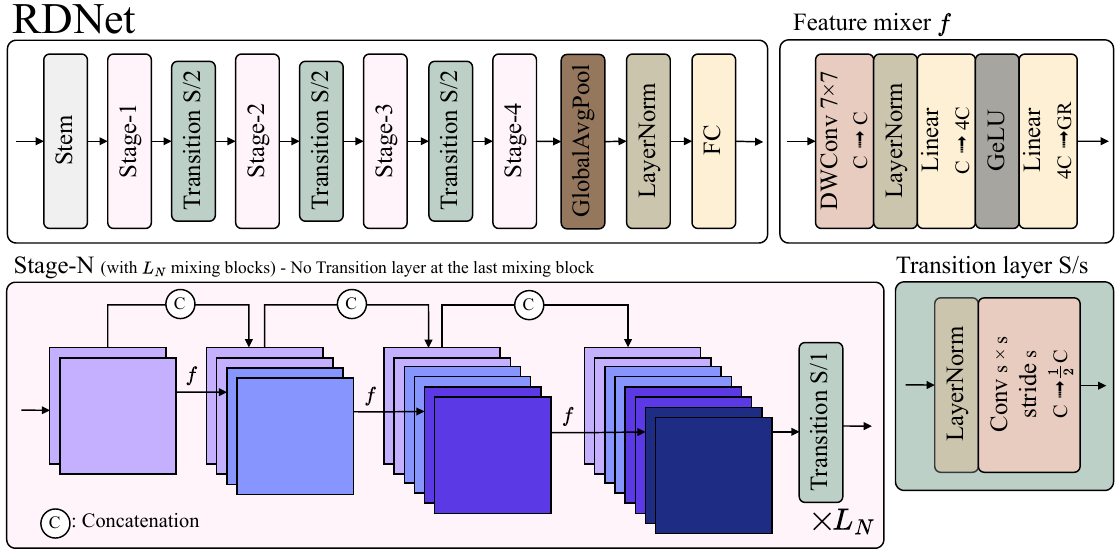} 
    \vspace{-.75em}
    \caption{\textbf{Schematic illustration of \ours.}  \ours features a unique design distinguishing it from ResNet-style architectures, primarily due to the use of feature concatenation. We design four stages in \ours across all scales, where each stage-N comprises $L_N$ mixing blocks consisting of three feature mixers and one transition layer (the last mixing block does not employ the transition layer). Feature mixer $f$ denotes our building block combines previously concatenated features to compress them into GR-dimensional features for concatenation. The growth rate (GR) adjusts the amount of concatenated features and is predetermined for each stage. Transition layers for downsampling are positioned after each stage as before. S and C denote stride and channel size. This figure illustratively sets GR to two.}
    \label{fig:rdnet_architecture}
    \vspace{-1em}
\end{figure}

\vspace{-1em}
\subsubsection{Larger intermediate channel dimensions.}
A large input dimension for the depthwise convolution is crucial~\cite{cvpr2018mobilenetv2}. By adeptly modulating expansion ratio (ER) for inverted bottlenecks in the previous works~\cite{cvpr2018mobilenetv2, icml2019efficientnet,han2021rethinking, icml2021EfficientNetV2, cvpr2022convnext} successfully achieved significant performance, by enlarging intermediate tensor size within the block beyond input dimensions (\eg, ER was tuned to 6). 

DenseNets similarly employed the ER concept; however, they distinctively applied it to the growth rate (GR) (\eg, ER=4$\times$GR) rather than to the input dimension to reduce both input and output dimensions. 
We argue that this harms the capability of encoded features through the nonlinearity~\cite{han2021rethinking}. %
Thus, we reengineer the approach by directing ER proportional to the input dimension (\ie, decoupling ER from GR). %
This change results in increased computational demands from a larger intermediate dimension; thus, halving GR (\eg, from 120 to 60) manages these demands without compromising accuracy. Namely, we enrich the features before applying nonlinearity and further compress the channels to control computational costs. %
Thereafter, we achieve both a faster training speed of 21\% and 0.4\%p improvement in accuracy shown in Table~\ref{tab:densenet_modifications}.
Additionally, we conduct a factor analysis to ascertain whether reducing ER and increasing GR is preferable, or conversely, elevating ER and decreasing GR; Table~\ref{tab:ablation_expansion_ratio} displays employing GR of 4 ultimately yields the optimal results.

\vspace{-1em}
\subsubsection{More transition layers.}
The transition layers~\cite{densenet} between stages are intended to reduce the number of channels. %
Due to the dense connections in every block, the intensified accumulation of features does not allow a high growth rate (GR). This gets worse as multiple blocks are stacked within a single stage, such as in the third stage, where numerous blocks accumulate in a single stage with low GRs. We introduce a novel aspect using more transition layers to address it. To be specific, we propose to use a transition layer in a stage, not solely after each stage, but after every three blocks with a stride of 1. These transition layers focus on dimension reduction rather than downsampling. This modification evidently reduces the computational costs substantially; therefore, we successfully increase overall GRs thanks to it\footnote{Increase in GR aims to address the overall low GR in the baseline at an \textit{architecture level}, whereas the abovementioned GR decrease was to boost ER on a \textit{block level}.}. This is further supported by the results in Table~\ref{tab:ablation_transition_interval}, which reveals using transition layers frequently often improves accuracy.

Additionally, we note that the models exhibit low parameter counts compared to their FLOPs. We remedy this by introducing variable GR at different stages (\eg, 64, 104, 128, 192) instead of a uniform GR. Our further study in Table~\ref{tab:ablation_same_gr} suggests that a uniform growth rate (GR) compromises both accuracy and efficiency. Finally, Table~\ref{tab:densenet_modifications}(e) shows our design achieves significant accuracy improvements without greatly affecting computational costs.

\vspace{-1em}
\subsubsection{Patchification stem}
Recent advancements revealed the effectiveness of using image patches as inputs within a stem~\cite{2021patchconvnet, cvpr2022convnext, nips2022hornet}. We use the identical setup of a patch size 4 with a stride 4 as the patchification (LN~\cite{ln} follows). Our empirical findings suggest that employing the patchification yields a notable acceleration in computational speed without loss of precision (see Table~\ref{tab:densenet_modifications}(f)).

\vspace{-1em}
\subsubsection{Refined transition layers}
Another role of the transition layers was downsampling, and extra average poolings to downsample were adopted. We refine the transition layers, removing the average pooling and replacing the convolution by adjusting the kernel size and stride with the stride (LN replaces BN as well). Therefore, our transition layers play two additional roles: 1) dimension reduction, as aforementioned; 2) downsampling. Placing the transition layer after each stage exhibits +0.2\%p gain, barely hurting efficiency (see Table~\ref{tab:densenet_modifications}(g)). For the dimension reduction ratio, we reexamine the impact, previously explored in \cite{densenet}; Table~\ref{tab:ablation_transition_ratio} reconfirms 0.5 is optimal; higher transition ratios degrade precision. 

\vspace{-1em}
\subsubsection{Channel re-scaling.} We investigate if channel re-scaling is required due to the diverse variance of concatenated features. 
We examine our proposed re-scaling approach, which has a similar formulation by merging the channel layer-scale~\cite{icml2021deit} and an effective squeeze-excitation network~\cite{Lee2020CenterMask}. Table~\ref{tab:densenet_modifications}(h) indicates it achieves a slight +0.2\%p improvement, albeit with very minor inefficiency.%
\subsection{Revitialized DenseNet (\ours)}
We finally introduce Revitalized DenseNet (dubbed \ours), illustrated in Fig.~\ref{fig:rdnet_architecture}. Our final model achieves both enhanced precision and efficiency, particularly enjoying significantly faster speed (see Table~\ref{tab:densenet_modifications}(h) vs. Table~\ref{tab:densenet_modifications}(a)). %
\ours model family aligns with the widely-adopted scales~\cite{resnet,liu2021swin,cvpr2022convnext}. %
Our models distinctively include the Growth Rate, GR=$(GR_1, GR_2, GR_3, GR_4)$, and the number of the feature mixers in each stage, B=$(B_1, B_2, B_3, B_4)$, where we assign the number of the feature mixers per each stage being a multiple of 3 (\ie, $B_N{=}3L_N$), where $L_N$ is the number of the mixing blocks. We summarize the configurations below:

\vspace{-.5em}
\begin{coloritemize}
\item \ours-T: $\text{GR} = (64, 104, 128, 224), \text{B} = (3, 3, 12, 3)$
\item \ours-S: $\text{GR} = (64, 128, 128, 240), \text{B} = (3, 3, 21, 6)$
\item \ours-B: $\text{GR} = (96, 128, 168, 336), \text{B} = (3, 3, 21, 6)$
\item \ours-L: $\text{GR} = (128, 192, 256, 360), \text{B} = (3, 3, 24, 6)$
\end{coloritemize}

\section{Experiment}
\label{sec:exps}

\vspace{-.5em}
\subsection{Image Classification}
We evaluate our model family on ImageNet-1K~\cite{imagenet}. Our models are trained following the training setups in Swin Transformer~\cite{liu2021swin} and ConvNeXt~\cite{cvpr2022convnext} to ensure a fair comparison and not aimed to finetune the setups. The models are trained using AdamW~\cite{iclr2019AdamW} with a batch size of 512 and an initial learning rate of 1e-4 for 300 epochs. As aforementioned in our baseline in \S\ref{sec:baseline}, we employed identical data augmentations/regularization techniques to ConvNeXt's; EMA is not used for our training. The comprehensive details of the training recipe are detailed in Appendix. We follow the standard evaluation protocols~\cite{resnet,liu2021swin,cvpr2022convnext}.

\begin{figure}[t!]
    \centering
    \hspace{-.5em}
    \begin{subfigure}{.34\linewidth}
    \includegraphics[width=\textwidth]{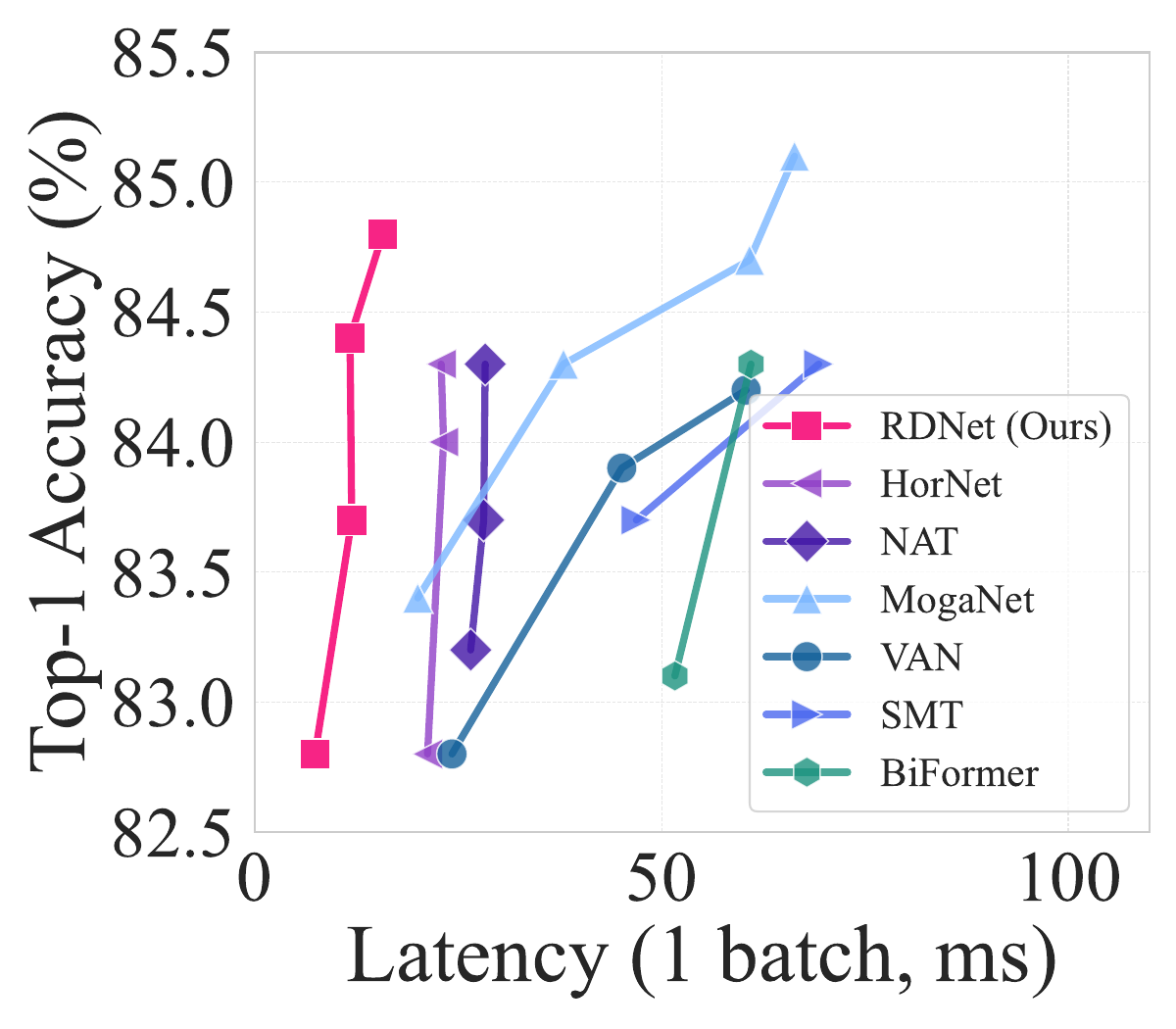} 
    \end{subfigure}
    \hspace{-1em}
    \begin{subfigure}{.34\linewidth}
    \includegraphics[width=\textwidth]{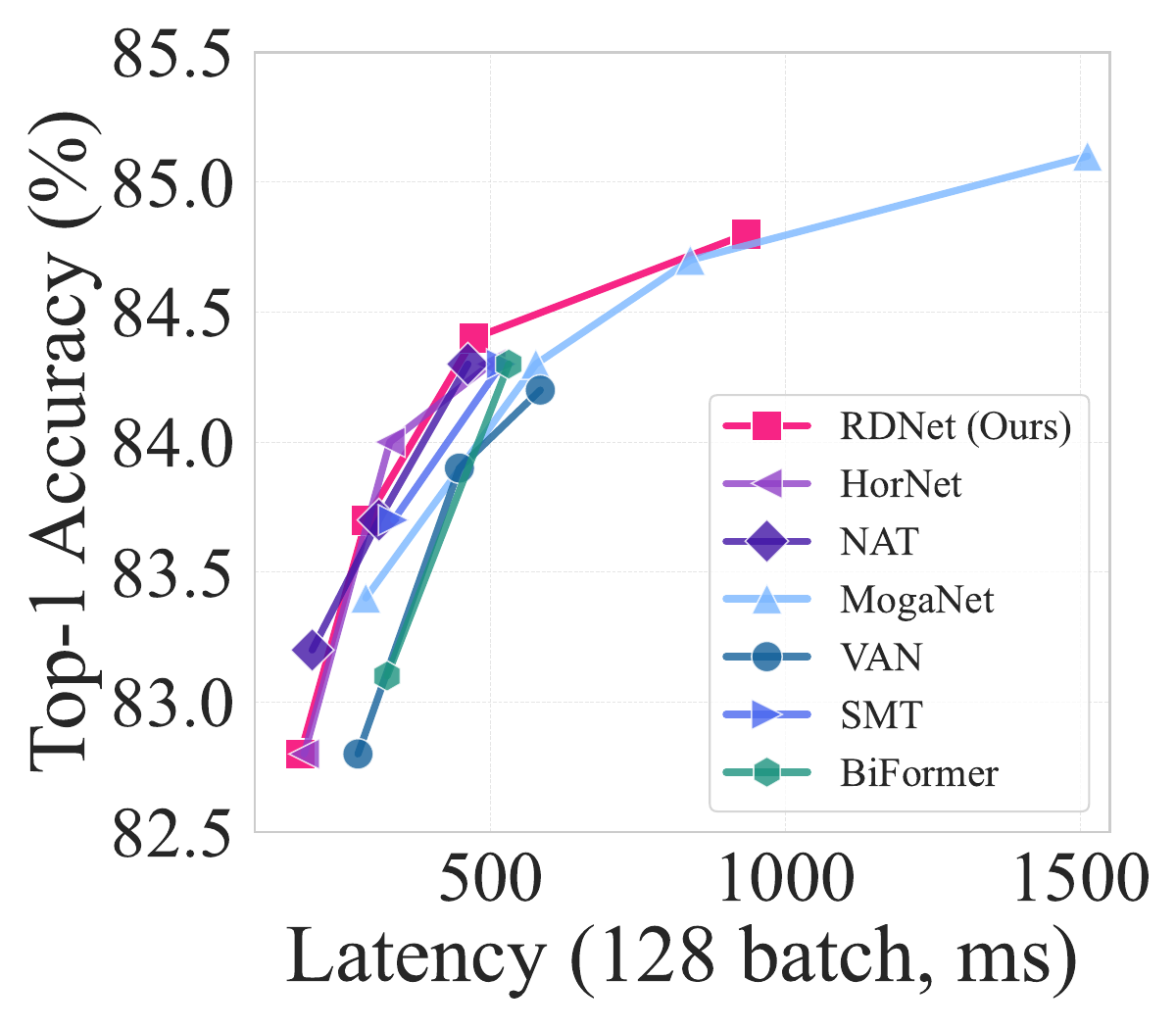}
    \end{subfigure}
    \hspace{-1em}
    \begin{subfigure}{.34\linewidth}
    \includegraphics[width=\textwidth]{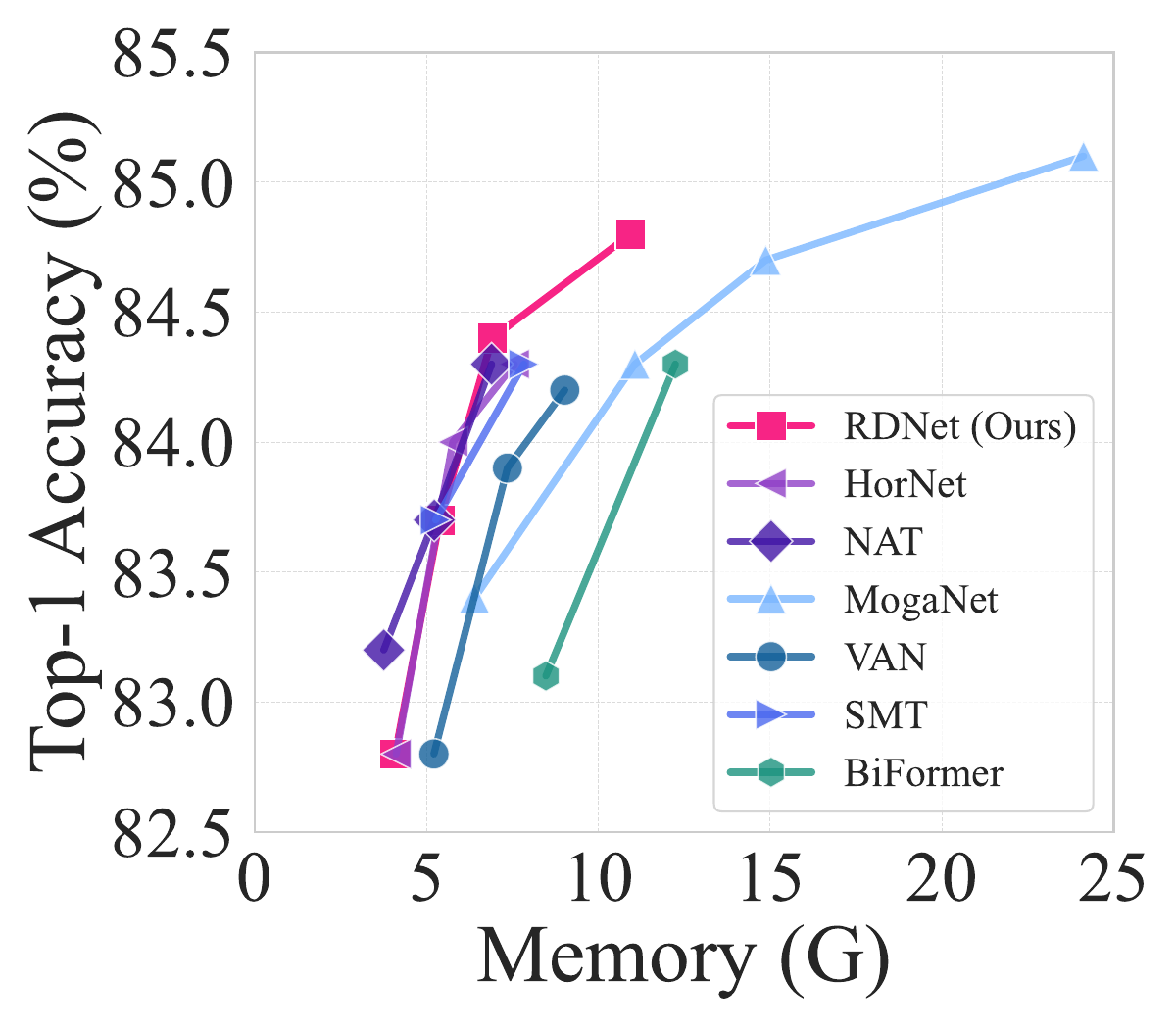} 
    \end{subfigure}
    \vspace{-2.5em}
    \caption{\textbf{ImageNet-1K performance trade-off among state-of-the-arts.} We provide comparative visualizations among \textit{state-of-the-art models}, which were known for top-performing models. It turns out that \ours is highly competitive in practice in terms of model speed and memory consumption.}
    \label{fig:latency_acc_with_sota}
    \vspace{-.5em}
\end{figure}

Our superiority is first underscored as compared with those of the current top-performing architectures~\cite{nips2022hornet,hassani2023NAT,cvmj2023VAN,Weifeng2023smt,zhu2023biformer}. We visualize the trade-off plots in Fig.~\ref{fig:latency_acc_with_sota} and detail the accuracies with diverse computational costs in Table~\ref{tab:in1k_latest}. Ours show very competitive results compared with state-of-the-art models. Table~\ref{tab:in1k_latest} exhibits that while our models slightly fall behind in accuracy, they significantly make up with speed metrics. For example, \ours-S can match with other lighter models such as SMT-S or MogaNet-S. Notably, ours do not require large memory usage as we aimed but achieve further efficiency. 

\begin{figure}[t!]
    \centering
    \hspace{-.75em}
    \begin{subfigure}{.33\linewidth}
    \includegraphics[width=\textwidth]{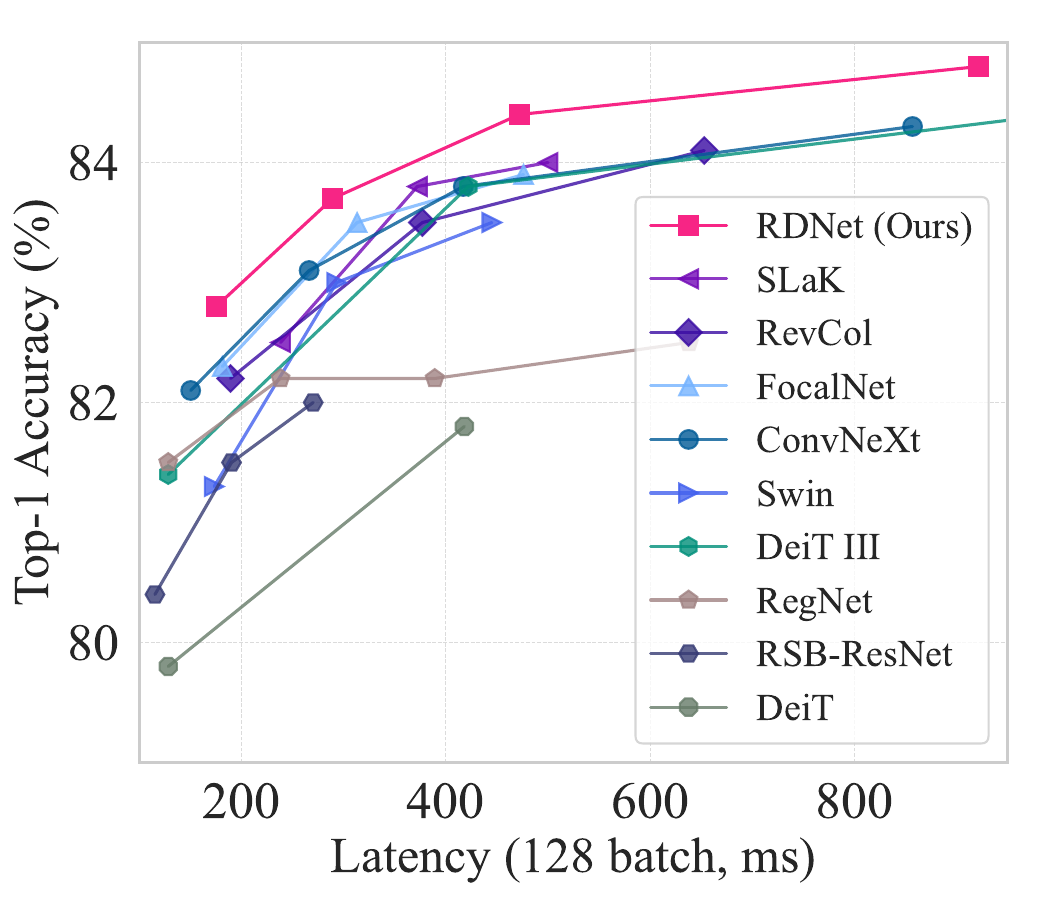} 
    \end{subfigure}
    \begin{subfigure}{.33\linewidth}
    \includegraphics[width=\textwidth]{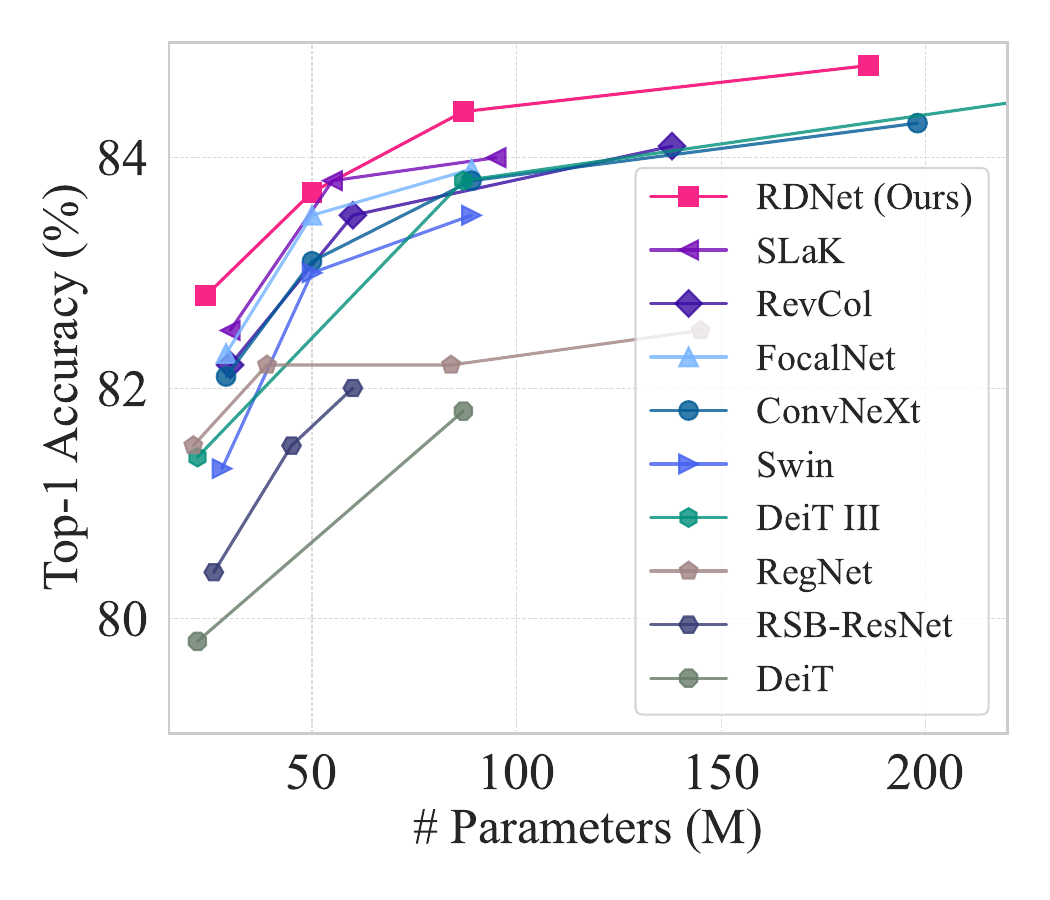}
    \end{subfigure}
    \begin{subfigure}{.33\linewidth}
    \includegraphics[width=\textwidth]{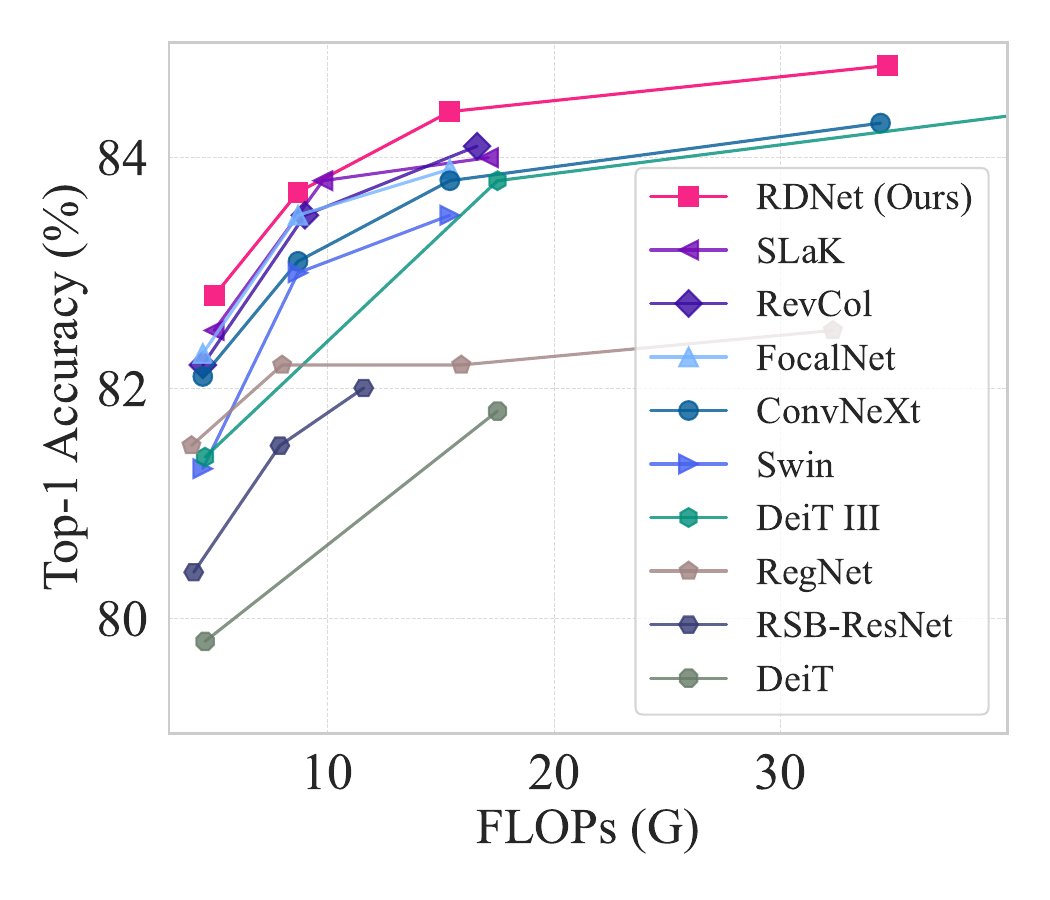} 
    \end{subfigure}
    \vspace{-2em}
    \caption{\textbf{ImageNet-1K performance trade-off among previous milestones.} We provide comparative visualizations between previous architectures and our models. Notice that we also include speed comparisons to highlight actual differences in practice. Our models outperform the competing modern architectures revealing the potential of feature concatenation in designing networks. }%
    \label{fig:latency_acc}
    \vspace{-.5em}
\end{figure}

We further exhibit a comparison with the popular models in Table~\ref{tab:in1k}. Ours surpass competitors by high precision, with decent memory usage and faster speeds. We further visualize trade-offs in Fig.~\ref{fig:latency_acc}, where \ours demonstrates competitive performance even when juxtaposed with the milestone architectures. %

\begin{table}[t!]
\small
\centering
\caption{\textbf{ImageNet-1K comparison with the latest models.} Fig.~\ref{fig:latency_acc_with_sota} visualized this table. We thoroughly compare our models against the latest architectures in practical latency and memory usage to demonstrate superiority. b$n$ denotes latency (ms), measured with a batch size of $n$. Mem denotes the memory occupation (GB) measured with a batch size of 16. Interestingly, while our models slightly lag in accuracy, they significantly compensate with superior speed metrics.  }
\label{tab:in1k_latest}
\vspace{-1em}
\tabcolsep=0.3em
\resizebox{0.94\linewidth}{!}{
\begin{tabular}{lccccrrrrrrr}
\toprule
Model                                       & Date     &Param & FLOPs & Top-1 & b1   & b8    & b16   & b32   & b64    & b128 & Mem  \\ 
\midrule
\gr \ours-T                                 & Ours      & 24    & 5.0   & 82.8  & 7.4  & 13.4& 24.6& 45.7  & 88.9  & 175.2 & 4.1 \\ 
HorNet-T$_{7\times 7}$~\cite{nips2022hornet}& NeurIPS'2022& 22    & 4.0   & 82.8  & 21.2 & 23.2  & 27.0  & 50.7  & 96.1  & 183.7 & 4.1  \\
VAN-B2~\cite{cvmj2023VAN}                   & CVMJ'2023   & 27    & 5.0   & 82.8  & 24.2 & 28.0  & 39.0  & 75.4  & 144.4 & 274.8 & 5.2 \\
BiFormer-S~\cite{zhu2023biformer}           & CVPR'2023   & 26    & 4.5   & 83.8  & 51.6 & 50.5  & 50.9  & 86.8  & 167.5 & 197.2 & 8.5 \\
NAT-T~\cite{hassani2023NAT}                 & CVPR'2023   & 28    & 4.3   & 83.2  & 26.5 & 28.0  & 33.2  & 53.0  & 102.2 & 335.3 & 3.8 \\
SMT-S~\cite{Weifeng2023smt}                 & ICCV'2023   & 21    & 4.7   & 83.7  & 46.9 & 48.2  & 55.8  & 96.0  & 176.5 & 335.3 & 5.3 \\
MogaNet-S~\cite{anonymous2024moganet}       & ICLR'2024   & 25    & 5.0   & 83.4  & 20.0 & 22.4  & 40.9  & 77.2  & 147.4 & 288.1 & 6.4 \\\midrule
\gr \ours-S                                 & Ours      & 50    & 8.7   & 83.7  & 11.9 & 21.5  & 39.8  & 74.0  & 144.2 & 289.0 & 5.4  \\ 
HorNet-S$_{7\times 7}$~\cite{nips2022hornet}& NeurIPS'2022& 50     & 8.8   & 84.0  & 23.2 & 25.7  & 46.0  & 88.3  & 171.4 & 328.9 & 5.7  \\
VAN-B3~\cite{cvmj2023VAN}                   & CVMJ'2023   & 45    & 9.0   & 83.9   & 45.1 & 49.4  & 64.1  & 123.1 & 237.2 & 446.9 & 7.4 \\
BiFormer-B~\cite{zhu2023biformer}           & CVPR'2023   & 57    & 9.8   & 84.3   & 60.4 & 67.8  & 85.8  & 161.2 & 311.9 & 584.2 & 12.2 \\
NAT-S~\cite{hassani2023NAT}                 & CVPR'2023   & 51    & 7.8   & 83.7   & 28.1 & 28.2  & 43.4  & 82.7  & 159.9 & 310.4 & 5.2 \\
SMT-B~\cite{Weifeng2023smt}                 & ICCV'2023   & 32    & 7.7   & 84.3   & 69.3 & 70.9  & 87.1  & 149.6 & 272.5 & 518.7 & 7.8 \\
MogaNet-B~\cite{anonymous2024moganet}       & ICLR'2024   & 44    & 9.9   & 84.3  & 37.9 & 43.8  & 80.9  & 152.9 & 294.0 & 576.6 & 11.1 \\
\midrule
\gr \ours-B                                 & Ours     & 87     & 15.4  & 84.4  & 11.7 & 32.2  & 61.4  & 116.6 & 233.7 & 471.6 & 6.9  \\ 
HorNet-B$_{7\times 7}$~\cite{nips2022hornet}& NeurIPS'2022& 87     & 15.6  & 84.3  & 22.9 & 37.9  & 71.5  & 134.5 & 259.6 & 500.0 & 7.7  \\
VAN-B4~\cite{cvmj2023VAN}                   & CVMJ'2023  & 60     & 12.2  & 84.2 & 60.4 & 67.8 & 85.8 & 161.2 & 311.9 & 584.2 & 9.0 \\
NAT-B~\cite{hassani2023NAT}                 & CVPR'2023  & 90     & 13.7  & 84.3 & 28.3 & 33.5 & 43.4 & 82.7 & 159.9 & 310.4 & 5.2 \\
MogaNet-L~\cite{anonymous2024moganet}       & ICLR'2024  & 83     & 15.9  & 84.7  & 60.8 & 64.9  & 118.5 & 224.3 & 429.2 & 838.6 & 14.9 \\
\midrule
\gr \ours-L                                 & Ours     & 186    & 34.7  & 84.8  & 15.7 & 63.2  & 121.0 & 233.3 & 460.7 & 933.7 & 10.9  \\ 
MogaNet-XL~\cite{anonymous2024moganet}      & ICLR'2024  & 181    & 34.5  & 85.1  & 66.3 & 112.3 & 207.5 & 394.0 & 771.9 & 1512.5 & 24.1 \\
\bottomrule
\end{tabular}
}
\vspace{-1.4em}
\end{table}

\vspace{-1em}
\subsection{Zero-shot Image Classification}
\begin{wraptable}{r}{15em}
\small
    \vspace{-5.7em}
    \caption{\textbf{ImageNet-1K zero-shot classification results.} Ours outperforms ConvNeXt further in efficiency.}
    \label{tab:openclip}
    \tabcolsep=0.4em
    \resizebox{1\linewidth}{!}{
    \begin{tabular}{cccc}
    \toprule
    Models  & Param &  Top-1 & Top-5 \\ \midrule
    ConvNeXt-B & 152 & 51.2 & 79.3 \\
    \gr \ours-B & \textbf{150} & \textbf{54.1} & \textbf{82.1} \\
    \bottomrule
    \end{tabular}
    }
    \vspace{-3em}
\end{wraptable} 

We evaluate \ours on ImageNet-1k zero-shot performance by training CLIP~\cite{clip} to verify the applicability under a different training scheme. We follow the training protocol in ConvNeXt-OpenCLIP~\cite{openclip} using 1.28B seen images from the aggregated set of CC3M~\cite{cc3m}, CC12M~\cite{cc12m}, and RedCaps~\cite{redcaps}. We use the OpenCLIP codebase\footnote{\url{https://github.com/mlfoundations/open_clip}}.

\begin{table}[ht!]
\small
\centering
\caption{\textbf{ImageNet-1K performance comparison with milestones.} We report top-1 accuracy (\%), parameter count (M), FLOPs (G), inference time (ms) for 128 images, and memory usage (GB) with a batch size of 16.  All models are (pre-)trained on ImageNet-1k from scratch.  }
\label{tab:in1k}
\vspace{-1em}
\hspace{-.55em}
\resizebox{0.5\linewidth}{!}{
\begin{tabular}{@{}llccccccccc@{}}
\toprule
Model                                 & {Res}  & {Param}  & FLOPs &  Lat     & Mem      & Top-1        \\ \midrule
RSB-ResNet50~\cite{wightman2021rsb}                  & $224^2$  & 26        & 4.1      & 115         & 2.1      & 80.4            \\
RegNetY-4GF~\cite{cvpr2020regnet}           & $224^2$  & 21        & 4.0      & 128         & 2.7      & 81.5            \\
Deit-S~\cite{icml2021deit}                     & $224^2$  & 22        & 4.6      & 128         & 1.9      & 79.8            \\
CoaT-Lite-S~\cite{iccv2021coat}                   & $224^2$  & 22        & 4.0      & 211         & 3.3      & 81.9            \\
Swin-T~\cite{liu2021swin}                         & $224^2$  & 28        & 4.5      & 173         & 2.6      & 81.3            \\
PVTv2-B2-Li~\cite{cvmj2022PVTv2}                  & $224^2$  & 23          & 3.9         &  173       & 4.4        & 82.1            \\
FocalNet-T ~\cite{nips2022focalnet}       & $224^2$  & 29      & 4.5     &  181       & 4.0     & 82.3          \\
ConvNeXt-T~\cite{cvpr2022convnext}                 & $224^2$  & 29        & 4.5      & 150         & 2.7      & 82.1            \\
CSWin-T~\cite{cvpr2022CSWin}                      & $224^2$  & 23        & 4.3      & 194         & 2.7      & 82.8            \\
Deit III-S~\cite{eccv2022deit3}                   & $224^2$  & 22        & 4.6      & 128         & 2.0      & 81.4            \\
RevCol-T~\cite{cai2023reversible}                  & $224^2$  & 30        & 4.5      & 189         & 2.0      & 82.2            \\
SLaK-T~\cite{Liu2022SLak}                          & $224^2$  & 30        & 5.0      & 238         & 3.3      & 82.5            \\
InceptionNeXt-T~\cite{yu2023inceptionnext} & $224^2$ & 28 & 4.2 & 132 & 3.3 & 82.3 \\

\gr \ours-T                                            & $224^2$  & 24        & 5.0      & 175         & 4.1      & 82.8            \\ \midrule
RSB-ResNet101~\cite{wightman2021rsb}                  & $224^2$  & 45        & 7.9      & 190         & 3.9      & 81.5            \\
RegNetY-8GF~\cite{cvpr2020regnet}           & $224^2$  & 39        & 8.0      & 238         & 4.0      & 82.2            \\
NFNet-F0~\cite{brock2021high}                    & $224^2$  & 71        & 12.4     & 235         & 3.5      & 83.6            \\
CoaT-Lite-M~\cite{iccv2021coat}                    & $224^2$  & 45        & 9.8      & 396         & 5.5      & 83.6            \\
Swin-S~\cite{liu2021swin}                          & $224^2$  & 50        & 8.7      & 293         & 3.9      & 83.0            \\
PVTv2-B4~\cite{cvmj2022PVTv2}                      & $224^2$  & 63        & 10.1      & 370            &  7.2       &   83.6          \\
ConvNeXt-S~\cite{cvpr2022convnext}                 & $224^2$  & 50        & 8.7      & 266         & 4.0      & 83.1            \\
CSWin-S~\cite{cvpr2022CSWin}                       & $224^2$  & 35        & 6.9      & 313         & 4.0      & 83.6            \\
FocalNet-S ~\cite{nips2022focalnet}           & $224^2$  & 50        & 8.7      & 313         & 4.6      & 83.5            \\
RevCol-S~\cite{cai2023reversible}                  & $224^2$  & 60        & 9.0      & 377         & 2.4      & 83.5            \\
SLaK-S~\cite{Liu2022SLak}                          & $224^2$  & 55        & 9.8      & 372         & 5.0      & 83.8            \\
InceptionNeXt-S~\cite{yu2023inceptionnext} & $224^2$ & 49 & 8.4 & 245 & 3.2 & 83.5 \\
\gr \ours-S                                            & $224^2$  & 50        & 8.7      & 289         & 5.4      & 83.7            \\ \midrule
\end{tabular}
}
\hspace{-.8em}
\resizebox{0.51\linewidth}{!}{
\begin{tabular}{@{}llccccccccc@{}}
\toprule
Model                                 & {Res}  & {Param}  & FLOPs &  Lat     & Mem      & Top-1        \\ \midrule
RSB-ResNet152~\cite{wightman2021rsb}  & $224^2$  & 60       & 11.6  &  270     & 4.7    & 82.0            \\
RegNetY-16GF~\cite{cvpr2020regnet}          & $224^2$  & 84        & 15.9     & 389         & 5.4      & 82.2            \\
DeiT-B~\cite{icml2021deit}                      & $224^2$  & 87        & 17.5     & 418         & 3.8      & 81.8            \\
Swin-B~\cite{liu2021swin}                          & $224^2$  & 89        & 15.4     & 445         & 5.4      & 83.5            \\ %
PVTv2-B5~\cite{cvmj2022PVTv2}                      & $224^2$  &  82         & 11.8         & 414       &  7.0       &  83.8           \\
ConvNeXt-B~\cite{cvpr2022convnext}                 & $224^2$  & 89        & 15.4     & 417         & 5.4      & 83.8            \\
CSWin-B~\cite{cvpr2022CSWin}                       & $224^2$  & 78        & 15.0     & 543         & 6.5      & 84.2            \\ %
RepLKNet-31B~\cite{cvpr2022replknet}               & $224^2$  & 79        & 15.3     & 461         & 2.7      & 83.5            \\
DeiT III-B~\cite{eccv2022deit3}                    & $224^2$  & 87        & 17.5     & 422         & 4.0      & 83.8            \\ %
FocalNet-B ~\cite{nips2022focalnet}           & $224^2$  & 89        & 15.4     & 476         & 6.1      & 83.9            \\
RevCol-B~\cite{cai2023reversible}                  & $224^2$  & 138       & 16.6     & 653         & 3.5      & 84.1            \\
SLaK-B~\cite{Liu2022SLak}                          & $224^2$  & 95        & 17.1     & 558         & 6.9      & 84.0            \\
InceptionNeXt-B~\cite{yu2023inceptionnext} & $224^2$ & 87 & 14.9 & 405 & 6.1 & 84.0 \\
\gr \ours-B                                            & $224^2$  &  87       & 15.4     & 472         & 6.9      & 84.4            \\ \midrule %
RegNetY-32GF~\cite{cvpr2020regnet}          & $224^2$  & 145       & 32.3     & 638         & 7.3      & 82.5            \\
NFNet-F1~\cite{brock2021high}                    & $320^2$  & 133       & 35.5     & 421         & 5.9      & 84.7            \\
DeiT III-L~\cite{eccv2022deit3}                    & $224^2$  & 304       & 61.6     & 1375        & 10.5     & 84.9            \\
DeiT III-L~\cite{eccv2022deit3}                    & $384^2$  & 304       & 191.2    & 4586        & 28.1     & 85.8            \\
ConvNeXt-L~\cite{cvpr2022convnext}                 & $224^2$  & 198       & 34.4     & 857         & 8.6      & 84.3            \\ 
ConvNeXt-L~\cite{cvpr2022convnext}                 & $384^2$  & 198       & 101.1    & 2550        & 19.0     & 85.5            \\
\gr \ours-L                                            & $224^2$  & 186       & 34.7     & 934         & 10.9     & 84.8            \\ %
\gr \ours-L                                            & $384^2$  & 186       & 101.9    & 2714        & 24.3     & 85.8            \\
\bottomrule
\vspace{5em}
\end{tabular}
}
\vspace{-1em}
\end{table}

\vspace{-1em}
\subsection{Semantic Segmentation}
We employ ImageNet-1K pre-trained weights to perform semantic segmentation on the ADE20K~\cite{Zhou2018ADE20k} dataset using UperNet~\cite{eccv2018upernet}. We use a learning rate of 8e-5 with a weight decay of \(0.03\), and utilize stochastic depth rate \(0.1\), \(0.2\), and \(0.3\) for the \ours-T, -S, and -B, respectively. The remainder of the training settings follows ConvNeXt~\cite{cvpr2022convnext}. As demonstrated in Table~\ref{tab:ade20k}, \ours exhibits strong performance, which reveals the effectiveness on dense prediction tasks.

\begin{table}[t!]
    \small
    \centering
    \caption{\textbf{ADE20K semantic segmentation results.} All trained with the unified head UperNet (160K) on ADE20K. FLOPs (G) are measured at $512\times2048$ resolutions.}
    \label{tab:ade20k}
    \vspace{-1em}
    \tabcolsep=0.5em
    \resizebox{0.65\linewidth}{!}{
    \begin{tabular}{lccccc}
        \toprule
        Architecture                                      & Crop      & Param & FLOPs   & mIoU$^{ss}$ & mIoU$^{ms}$ \\ \midrule
        
        Swin-T~\cite{liu2021swin}                      & 512$^2$   & 60     & 945     & 44.5        & 46.1        \\
        ConvNeXt-T~\cite{cvpr2022convnext}             & 512$^2$   & 60     & 939     & 46.0        & 46.7        \\
        RevCol-T~\cite{cai2023reversible}              & 512$^2$   & 60     & 937     & 47.4        &  47.8                 \\
        NAT-T~\cite{hassani2023NAT}                    & 512$^2$   & 58     & 934     & 47.1        & 48.4       \\
        \gr \ours-T                                            & 512$^2$   & 58 & 961 & \textbf{47.6}        & \textbf{48.6}        \\ \midrule
        Swin-S~\cite{liu2021swin}                      & 512$^2$   & 81     & 1038    & 47.6        & 49.5        \\
        ConvNeXt-S~\cite{cvpr2022convnext}             & 512$^2$   & 82     & 1027    & \textbf{48.7}        & 49.6        \\
        RevCol-S~\cite{cai2023reversible}              & 512$^2$   & 90     & 1031     & 47.9        &  49.0                 \\
        NAT-S~\cite{hassani2023NAT}                    & 512$^2$   & 82     & 1010    & 48.0         & 49.5                  \\
        \gr \ours-S                                            & 512$^2$   & 86 & 1040 & \textbf{48.7}        & \textbf{49.8}        \\ \midrule
        Swin-B~\cite{liu2021swin}                      & 512$^2$   & 121    & 1188    & 48.1        & 49.7        \\   %
        ConvNeXt-B~\cite{cvpr2022convnext}             & 512$^2$   & 122    & 1170    & 49.1        & 49.9        \\
        DeiT III-B~\cite{eccv2022deit3}                & 512$^2$   & 128    & 1283    & 49.3        &  50.2  \\ %
        RevCol-B~\cite{cai2023reversible}              & 512$^2$   & 122     & 1169     & 49.0        &  50.1                 \\
        NAT-B~\cite{hassani2023NAT}                    & 512$^2$   & 123     & 1137     & 48.5        & 49.7 \\
        \gr \ours-B                                            & 512$^2$   &  127  & 1187   & \textbf{49.6}        & \textbf{50.5}        \\
        \bottomrule
    \end{tabular}
    }
    \vspace{-.5em}
\end{table}

\vspace{-1em}
\subsection{Object Detection}
We evaluate object detection performance on COCO~\cite{2014MicrosoftCOCO} using Mask-RCNN~\cite{2017iccvmaskrcnn}. We use a learning rate of 3e-5 with a stochastic depth rate of 0.2. The remainder of the training settings follows ConvNeXt~\cite{cvpr2022convnext} again. As demonstrated in Table~\ref{tab:coco}, \ours exhibits competitive performance.

\begin{table}[b]
\centering
\caption{\textbf{COCO object detection and segmentation results.} We utilize Mask-RCNN with 3x schedule. FLOPs (G) are calculated with image size (1280, 800). The result of Swin-T is from the official repository~\cite{github_swin_detection}. }
\vspace{-1em}
\tabcolsep=0.35em
\resizebox{.85\linewidth}{!}{
\begin{tabular}{@{}lcccccccc@{}}
\toprule
Backbone & Param & FLOPs & $\text{AP}^{\text{box}}$ & $\text{AP}^{\text{box}}_{50}$ & $\text{AP}^{\text{box}}_{75}$ & $\text{AP}^{\text{mask}}$ & $\text{AP}^{\text{mask}}_{\text{50}}$ & $\text{AP}^{\text{mask}}_{75}$  \\ \hline
PVT-S~\cite{iccv2021PVT}                     & 44M & 304G & 43.0 & 65.3 & 46.9 & 39.9 & 62.5 & 42.8 \\
Swin-T~\cite{liu2021swin}                    & 48M & 267G & 46.0 & 68.1 & 50.3 & 41.6 & 65.1 & 44.9 \\
ConvNeXt-T~\cite{cvpr2022convnext}           & 48M & 262G & 46.2 & 67.9 & 50.8 & 41.7 & 65.0 & 44.9 \\
\gr \ours-T                                  & 43M & 278G & \textbf{47.5} & 68.5 & 52.1 & \textbf{42.4}& 65.6 & 45.7 \\ 
\bottomrule
\end{tabular}
 }
\label{tab:coco}
\vspace{-2em}
\end{table}

\section{Discussions}
\label{sec:discussion}

\subsection{Pilot Study - Random Network Experiments}
\label{sec:potential_concat}
This study aims to reveal the effectiveness of dense connections over residual connections. We train tons of random networks across various scenarios, which include 1) multiple network scales; 2) multiple types of building blocks; 3) a range of network architectural elements; and 4) different training setups. 
\vspace{-1.5em}
\subsubsection{Parameter spaces and cost constraints.} 
Table~\ref{tab:randnet_setups_results} (left) shows our parameter spaces for three individual scales, where RandNet$_{\mathcal{A}, \mathcal{B}, \mathcal{C}}$ are trained. We diversify the search space with respect to the budgets, such as parameter count, FLOPs, and memory consumption. We expand space from $\mathcal{C}$ to $\mathcal{D}$ by incorporating data augmentation and further to $\mathcal{E}$ with both data augmentation and a different optimizer~\cite{iclr2019AdamW}. Only randomly generated networks that meet the predefined budget are trained. We use the 90-epochs training setup~\cite{resnet} trained on Tiny-ImageNet~\cite{eccv2022tinyvit}. %
For $\mathcal{C,E}$ spaces using data augmentation~\cite{cvpr2016inceptionv3,zhang2017mixup,yun2019cutmix,zhong2020random,huang2016deep,cubuk2020randaugment}, training is done for 180 epochs. %
 Overall, the cumulative number of trained networks reach over 15k.

\begin{table}[t]
\centering
\small
\footnotesize
\caption{\textbf{Random network experiments.} We present our experimental setups (left) and results (right). Five parameter spaces guide random network generations for two distinct shortcuts. We sample random networks within each parameter space, ensuring similar computational costs. Each parameter space varies in 1) architectural elements - channel sizes, activations, normalizations, and convolution kernel sizes in $\mathcal{A, B, C, D, E}$; 2) data augmentations in $\mathcal{D, E}$; 3) optimizers in $\mathcal{E}$. $\mathcal{D, E}$ is based on the architectural space $\mathcal{C}$.  [$a_1, \dots, a_n$] and ($a$, $b$, $c$) denote a closed interval: a list of $n$ elements and a range of elements from $a$ to $b$ with a step of $c$, respectively.  %
All results are averaged. %
}
\label{tab:randnet_setups_results}
\vspace{-1em}
\tabcolsep=0.2em
  \resizebox{0.45\linewidth}{!}{
    \begin{tabular}{l|ccccccc}
    \toprule
    {Parameter space}        & \multicolumn{2}{c}{${\mathcal{A}}$}     &\multicolumn{2}{c}{\hspace{.5em} ${\mathcal{B}}$}      & \multicolumn{2}{c}{${\mathcal{C}}$}                 \\ \midrule
    Param ($x$M) & \multicolumn{2}{c}{ $2{<}x{<}2.5$}      & \multicolumn{2}{c}{\hspace{.5em} $4{<}x{<}5$}         & \multicolumn{2}{c}{$9{<}x{<}10$}       \\
    FLOPs ($x$G)    & \multicolumn{2}{c}{$2{<}x{<}{2.5}$}      & \multicolumn{2}{c}{\hspace{.5em} $4{<}x{<}5$}       & \multicolumn{2}{c}{$9{<}x{<}10$}         \\ 
    \midrule
    {Depth}          & \multicolumn{2}{c}{{(}3, 6, 1{)}}       & \multicolumn{2}{c}{\hspace{.5em} {(}3, 8, 1{)}}        & \multicolumn{2}{c}{{(}3, 12, 1{)}}       \\
    {Inter. channel dim} & \multicolumn{2}{c}{{(}32, 96, 8{)}}  & \multicolumn{2}{c}{\hspace{.5em} {(}64, 128, 8{)}}  & \multicolumn{2}{c}{{(}64, 192, 8{)}}  \\
    {Output channel dim}   & \multicolumn{2}{c}{{(}32, 96, 8{)}}  & \multicolumn{2}{c}{\hspace{.5em} {(}64, 128, 8{)}}  & \multicolumn{2}{c}{{(}64, 192, 8{)}}  \\
    Activations &  \multicolumn{6}{c}{[ReLU, SiLU, Mish, GELU, LeakyReLU]} \\
    Normalization layers & \multicolumn{6}{c}{[BatchNorm, LayerNorm]} \\
    Kernel sizes & \multicolumn{6}{c}{[3, 5, 7, 9]} \\ 
    \bottomrule 
    \toprule
    {Parameter space}        & \multicolumn{3}{c}{\hspace{4em} ${\mathcal{D}}$}     & \multicolumn{3}{c}{${\mathcal{E}}$}                       \\ \midrule
    Base space &  \multicolumn{3}{c}{\hspace{4em} $\mathcal{C}$} &  \multicolumn{3}{c}{$\mathcal{C}$}   \\
    Optimizer &  \multicolumn{3}{c}{\hspace{4em} -} &  \multicolumn{3}{c}{AdamW}   \\
    Data augmentation &  \multicolumn{3}{c}{\hspace{4em} \ding{51}} &  \multicolumn{3}{c}{\ding{51}}  \\ %
    \bottomrule
    \end{tabular}       
   }
\resizebox{.502\linewidth}{!}{

\begin{tabular}{c|cccc|c}
    \toprule
    Model                                 & Skip          & FLOPs  & Param & Mem  & Top-1 (\%)               \\ \midrule
    RandNet$_{\mathcal{A}}$               & add           & 2.25±0.13&2.21±0.13&0.65±0.03&45.8±2.0       \\ 
    \gr RandNet$_{\mathcal{A}}$           & concat        & 2.24±0.14&2.24±0.13&0.75±0.08&\textbf{47.5±2.1}   \\ \midrule
    RandNet$_{\mathcal{B}}$               & add           & 4.53±0.27&4.44±0.26&0.78±0.05&50.1±2.1    \\
    \gr RandNet$_{\mathcal{B}}$           & concat        & 4.53±0.29&4.51±0.28&0.90±0.11&\textbf{51.2±2.1}        \\ \midrule
    RandNet$_{\mathcal{C}}$               & add           & 9.61±0.23&9.41±0.23&1.01±0.09&53.2±2.2                    \\
    \gr RandNet$_{\mathcal{C}}$           & concat        & 9.53±0.26&9.44±0.26&1.24±0.17&\textbf{54.3±2.0}                \\ \midrule
    RandNet$_{\mathcal{D}}$               & add           & 9.60±0.23&9.40±0.23&1.02±0.09&57.4±1.5   \\
    \gr RandNet$_{\mathcal{D}}$           & concat        & 9.54±0.26&9.44±0.26&1.24±0.16&\textbf{58.1±1.4}     \\  \midrule
    RandNet$_{\mathcal{E}}$                & add          & 9.59±0.23&9.38±0.23&1.02±0.09&58.2±1.5     \\
    \gr RandNet$_{\mathcal{E}}$            & concat       & 9.54±0.26&9.44±0.26&1.25±0.17&\textbf{58.9±1.6}      \\
    \bottomrule
\end{tabular}
}
\vspace{-1em}
\end{table}

\begin{figure}[t!]
    \centering
    \small
    \hspace{-.75em}
    \begin{subfigure}{.195\linewidth}\hfill
        \includegraphics[width=\textwidth]{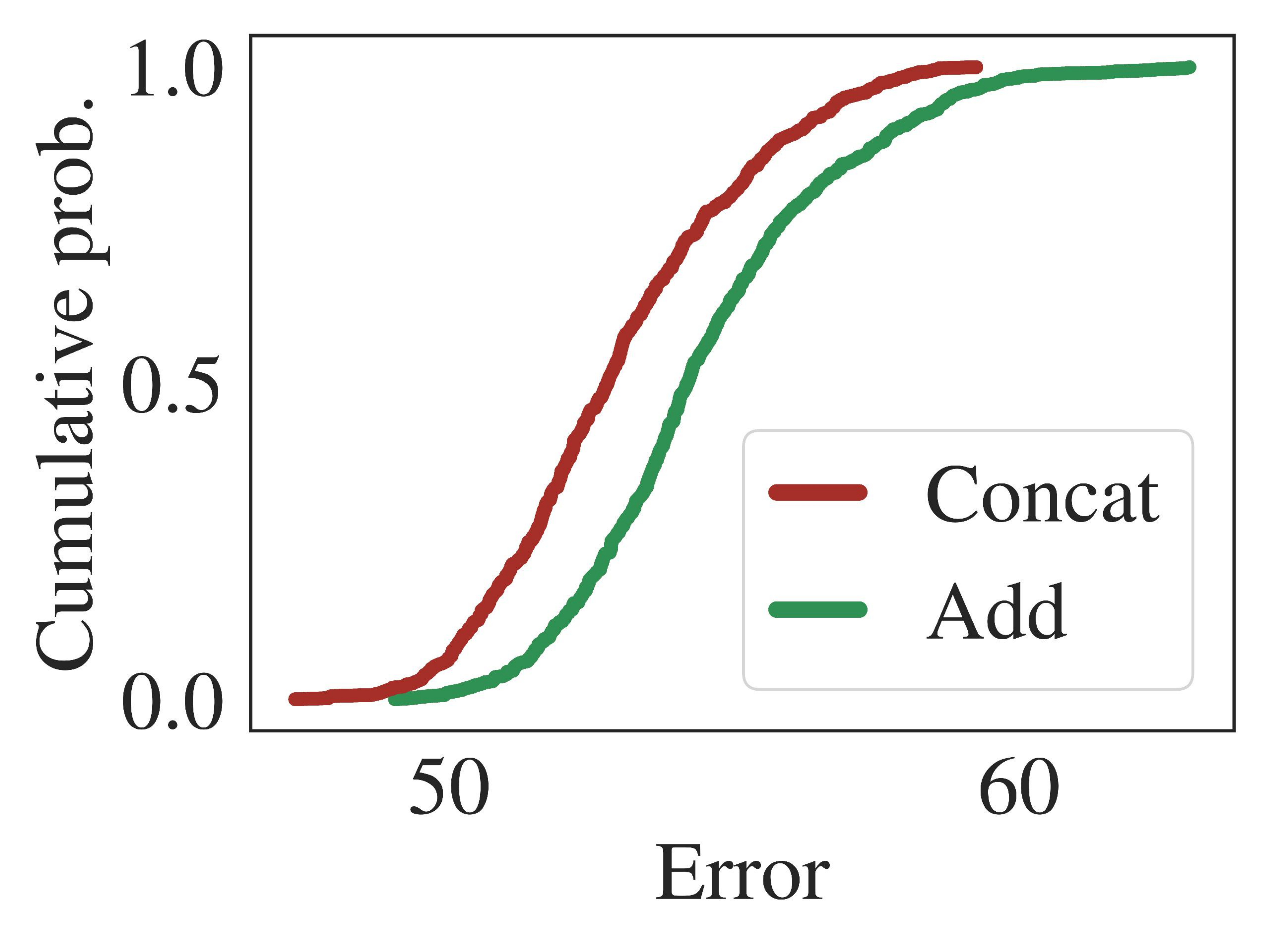} 
        \vspace{-1.75em}
        \caption{RandNet$_{\mathcal{A}}$}
    \end{subfigure}
    \begin{subfigure}{.195\linewidth}\hfill
        \includegraphics[width=\textwidth]{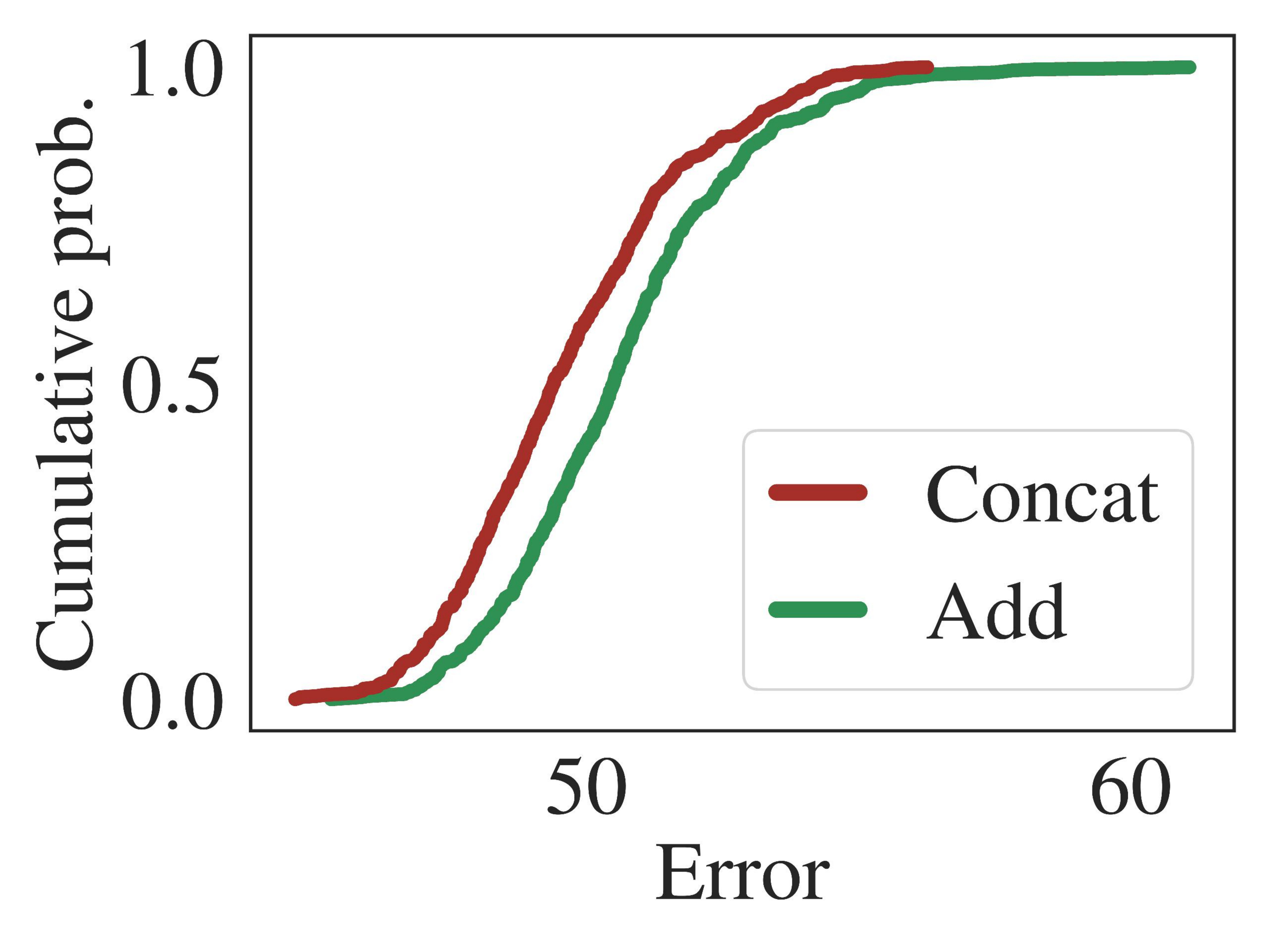}
        \vspace{-1.75em}
        \caption{RandNet$_{\mathcal{B}}$}
    \end{subfigure}
    \begin{subfigure}{.195\linewidth}\hfill
        \includegraphics[width=\textwidth]{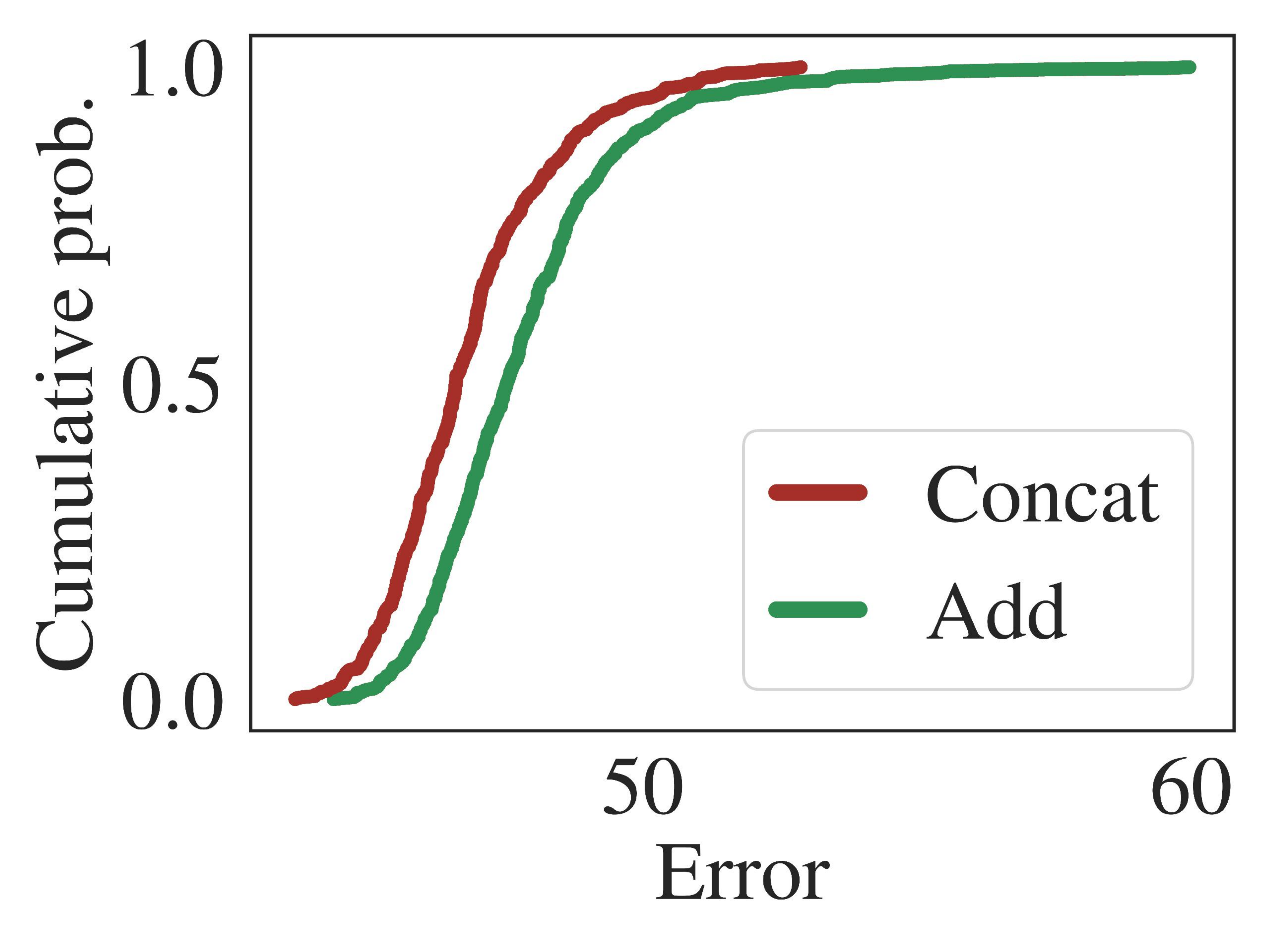} 
        \vspace{-1.75em}
        \caption{RandNet$_{\mathcal{C}}$}
    \end{subfigure}
    \begin{subfigure}{.195\linewidth}\hfill
        \includegraphics[width=\textwidth]{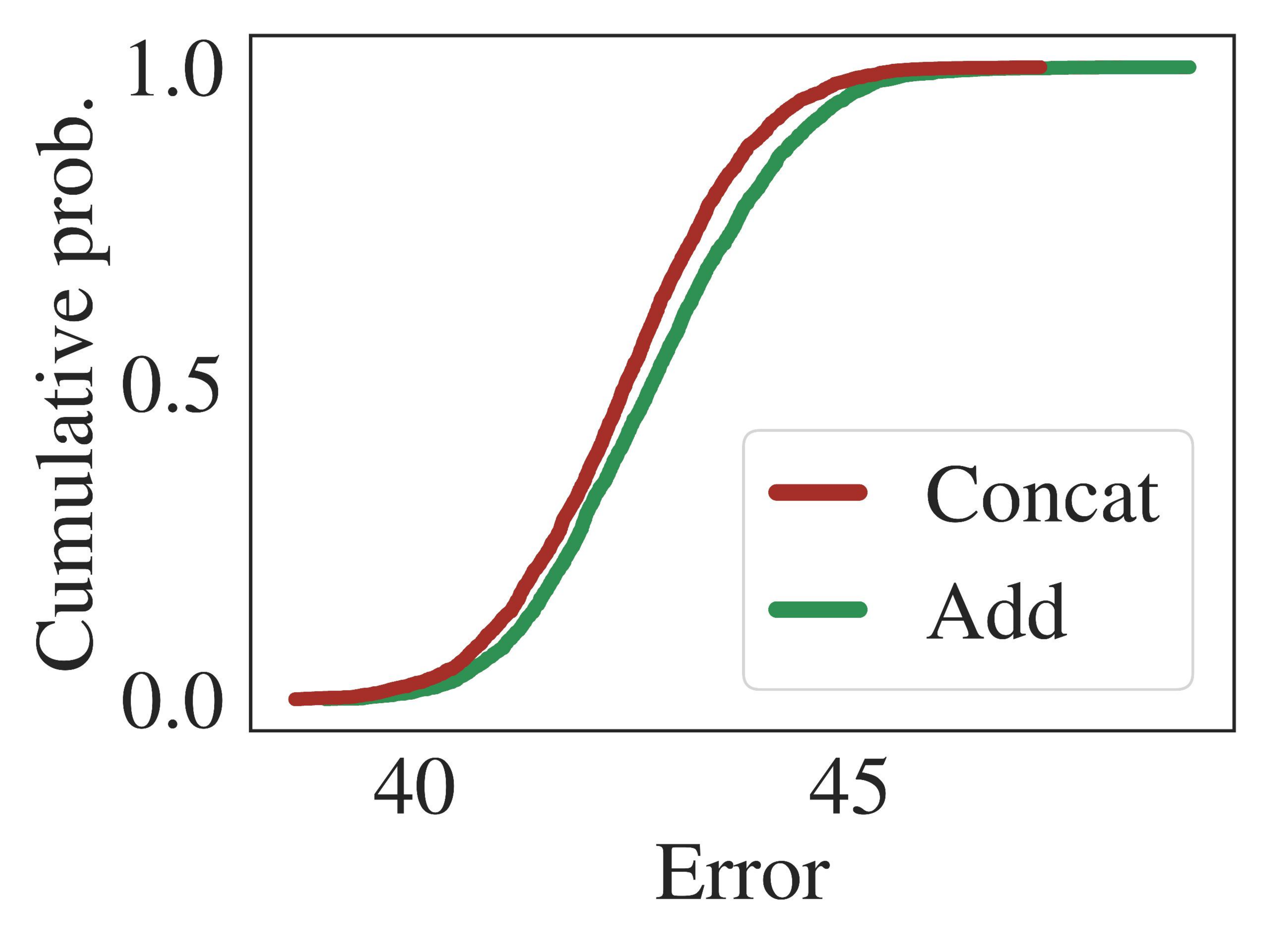} 
        \vspace{-1.75em}
        \caption{RandNet$_{\mathcal{D}}$}
    \end{subfigure}
    \begin{subfigure}{.195\linewidth}
        \includegraphics[width=\textwidth]{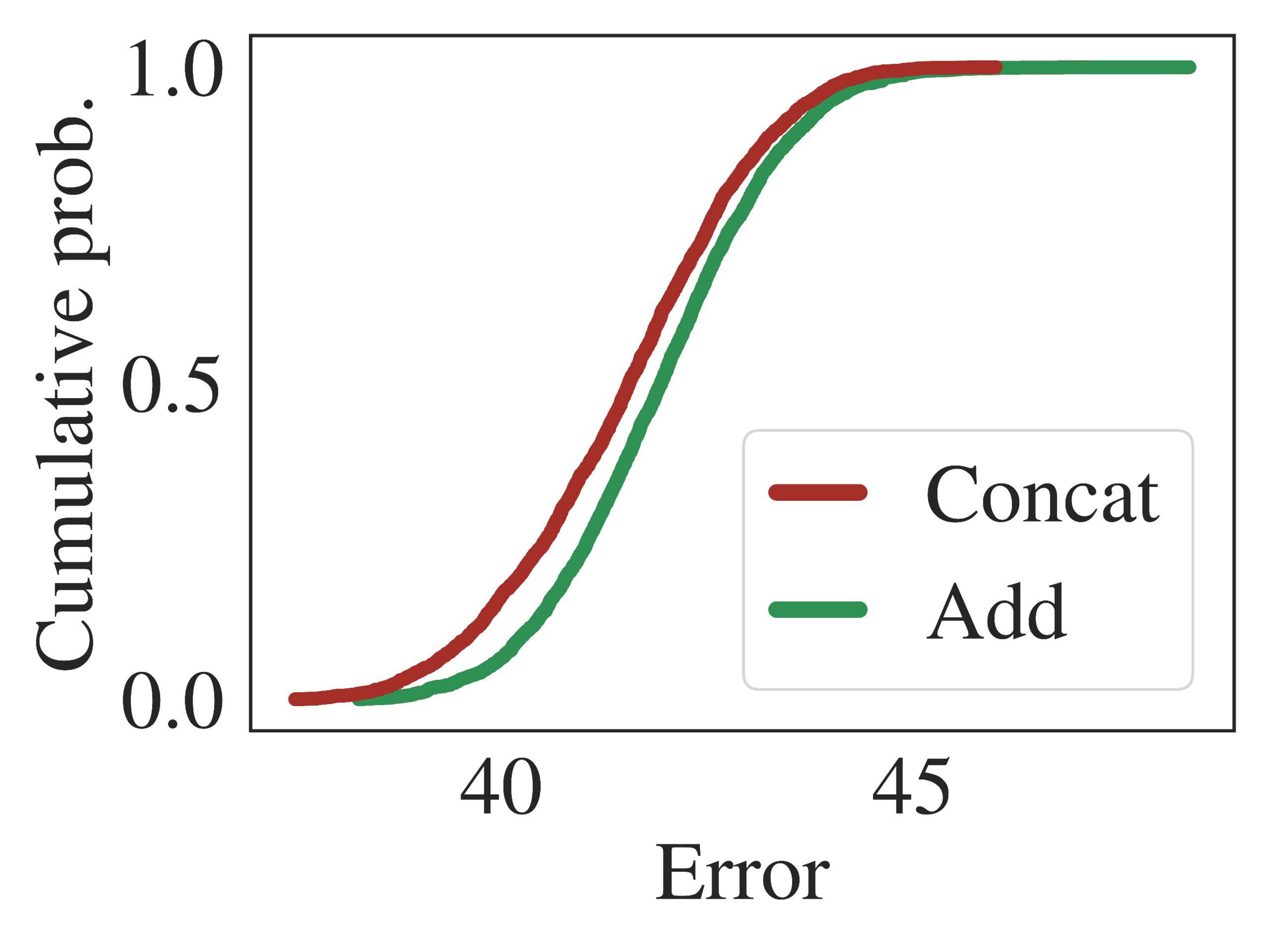} 
        \vspace{-1.75em}
        \caption{RandNet$_{\mathcal{E}}$}
    \end{subfigure}
    \vspace{-2em}
    \caption{\textbf{Cumulative probability vs. error} of trained models in Table~\ref{tab:randnet_setups_results} is visualized here  following  Radosavovic~\etal~\cite{cvpr2020regnet}. %
    Across all scales and settings, we observe concatenation-based models outperform those employing additive shortcuts.
    }
    \label{fig:randnet_cumulative_prob}
    \vspace{-1em}
\end{figure}

\vspace{-1.5em}
\subsubsection{RandNet architecture.} Based on \cite{resnet}, we stack random building blocks within the first stage. We generate random networks in the parameter space containing diverse depths, widths, activations, normalizations, and kernel sizes to provide flexibility under constrained costs (see Table~\ref{tab:randnet_setups_results}). 
Additionally, we diversify building blocks across all search spaces to conduct more extensive experiments. Three distinct architectural blocks - dubbed PreNorm, PostNorm, and PostNorm (w/o act) - are differentiated by the use of pre-activation and shortcut positions. PreNorm block adopts the pre-normalization~\cite{preresnet,densenet} precedes a skip connection. In contrast, two PostNorms enjoy post-normalization~\cite{resnet,cvpr2022convnext}. PostNorm varies from PostNorm (w/o act) based on the activation function post-skip connection.%

\vspace{-1em}

\vspace{-.5em}
\subsubsection{Result interpretation.}
Table~\ref{tab:randnet_setups_results} (right) exhibits that concatenation consistently outperforms additive shortcuts across all configurations. Furthermore, Fig.~\ref{fig:randnet_cumulative_prob} demonstrates the superior capability of concatenation-based architectures. %

\vspace{-1em}
\subsection{Impact of Input Size on Performance}
\begin{figure}[t!]
    \centering
    \begin{subfigure}{.32\linewidth}
        \includegraphics[width=\textwidth]{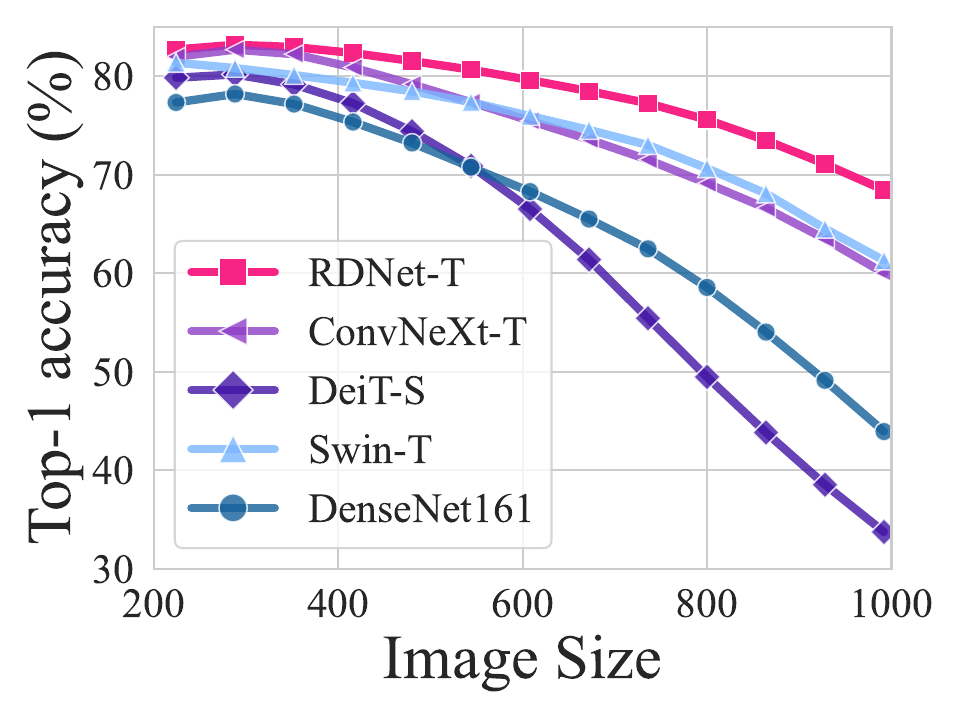}
    \end{subfigure}    
    \begin{subfigure}{.32\linewidth}
        \includegraphics[width=\textwidth]{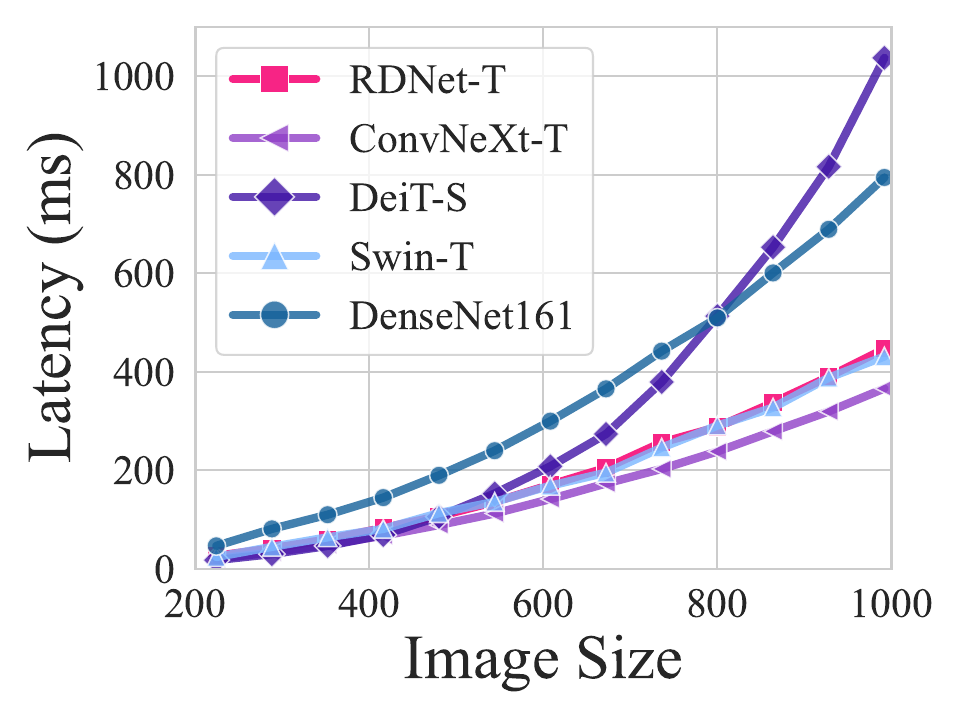}
    \end{subfigure}  
    \begin{subfigure}{.32\linewidth}
        \includegraphics[width=\textwidth]{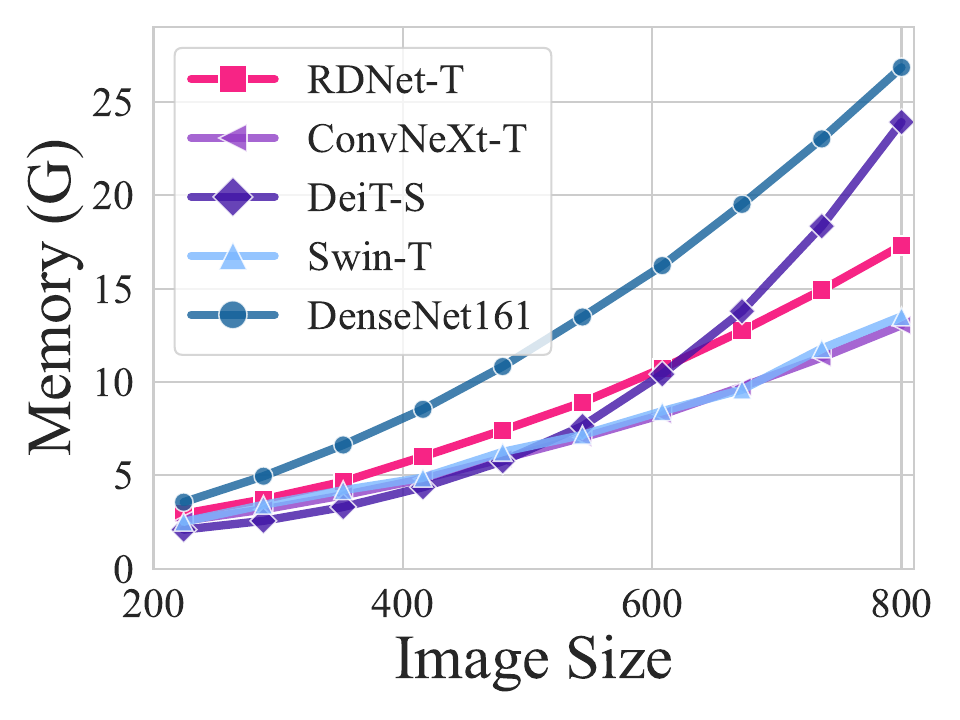}
    \end{subfigure}  
    \vspace{-1.25em}
\caption{\textbf{Accuracy/latency/memory vs. resolution.}  \ours enjoys resolution-robustness against various input image sizes to maintain accuracy. Furthermore, \ours exhibits a similar latency/memory trend to ConvNeXt and Swin Transformer, maintaining minimal increase with larger images compared to DeiT-S and DenseNet161. }
    \label{fig:img_scale_robustness}
    \vspace{-1.5em}
\end{figure}
We provide compelling findings regarding versus input size. First, Fig.~\ref{fig:img_scale_robustness} (left) shows \ours enjoys strong adaptability to input size variations. Intriguingly, DenseNet161, even trained without strong data augmentations, still enjoys adaptability, surpassing DeiT-S trained with strong data augmentations. We attribute this to the effectiveness of dense connections. 

Our finding further shows that, unlike width-oriented networks that slow with larger input sizes (due to the large intermediate tensors), our model's optimized width avoids latency/memory loss. Fig.~\ref{fig:img_scale_robustness} (middle, right) illustrates that \ours compete with ConvNeXt and Swin Transformer, diverging from DenseNet161~\cite{densenet} that gets slower and consumes more memory as image size grows. We note that larger scales (\eg, -S, -B, and -L) all follow the same trend.

\vspace{-1em}
\subsection{CKA analysis}
\vspace{-0.5em}

\begin{figure}[t]
    \centering
    \small
    \begin{subfigure}{.25\linewidth}
        \includegraphics[width=\textwidth]{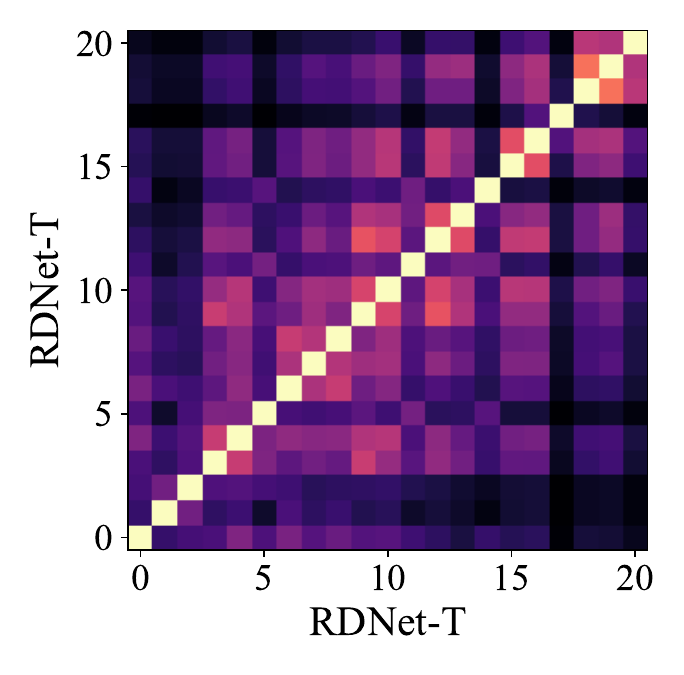} 
        \vspace{-1.75em}
        \caption{\ours-T}
    \end{subfigure}
    \begin{subfigure}{.25\linewidth} \quad
        \includegraphics[width=\textwidth]{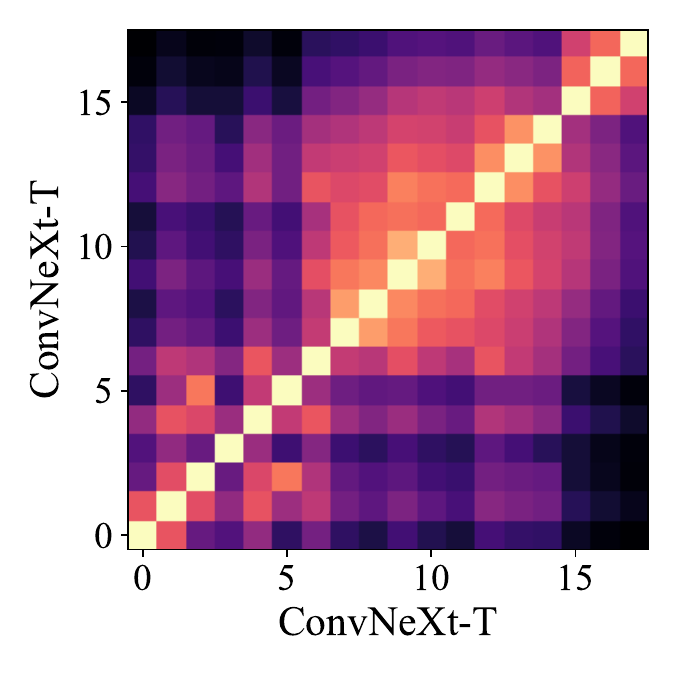}
        \vspace{-1.75em}
        \caption{ConvNeXt-T}
    \end{subfigure}
    \begin{subfigure}{.27\linewidth} \quad
        \includegraphics[width=\textwidth]{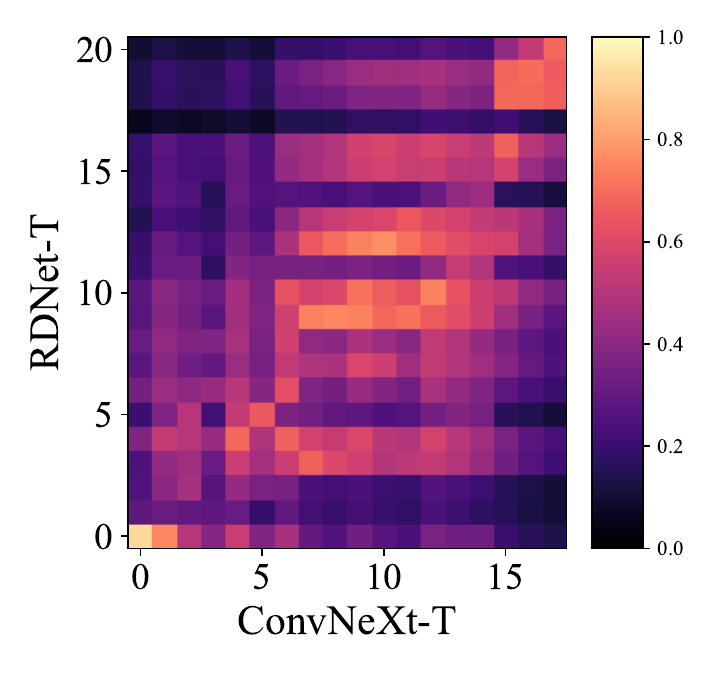} 
        \vspace{-1.75em}
        \caption{vs. ConvNeXt-T}
    \end{subfigure}  
    \vspace{-1em}
    \caption{\textbf{CKA analysis.} We compute CKA using features before passing through shortcuts (either concatenation or addition). This result suggests \ours generates diverse features across all layers in comparison to ConvNeXt. The third column presents a direct comparison between \ours and ConvNeXt. Overall, \ours learns distinct features from what ConvNeXt does and more varied features.}%
    \label{fig:cka}
    \vspace{-1em}
\end{figure}

We analyze the layer-specific features of \ours compared to ConvNeXt using Centered Kernel Alignment (CKA)~\cite{icml19cka}. %
Fig.~\ref{fig:cka} displays \ours learns distinct features at every layer, showcasing different patterns compared to ConvNeXt. In the third column, %
ConvNeXt and \ours astonishingly learn different features when compared, highlighting the unique learning dynamics of each model.

\begin{table*}[ht]
\centering
\small
\caption{\textbf{Ablation study results} are reported here with each ImageNet-1K accuracy (\%) with parameter count (M) and FLOPS (G). %
The best options are marked in \colorbox{baselinecolor}{gray}.}
\label{tab:ablation_study}
\vspace{-2em}
\subfloat[
\textbf{Depth/width scaling}
\label{tab:ablation_depth_vs_wide}
]{
\begin{minipage}{0.3\linewidth}
\centering
    \tabcolsep=0.5em
    \resizebox{0.915\linewidth}{!}{
    \begin{tabular}{@{}c|ccc@{}}
    \toprule
    Depth   & Param & FLOPs & {Top-1}   \\ \midrule
    3, 3, \:\:9, 3      & 23.3  & 5.0    & 82.5     \\  %
    \gr 3, 3, 12, 3     & 23.9  & 5.0    & 82.8     \\ 
    3, 3, 15, 3         & 23.5  & 5.0    & 82.8     \\   %
    \midrule
    3, 3, 18, 6        & 50.3  & 8.7    & 83.5     \\  %
    \gr 3, 3, 21, 6    & 50.4  & 8.7    & 83.7     \\ %
    3, 3, 24, 6        & 49.9  & 8.7    & 83.6     \\   %
    \midrule
    3, 3, 18, 6        & 89.2  & 15.4    & 84.2      \\  %
    \gr 3, 3, 21, 6    & 86.2  & 15.4    & 84.4    \\ %
    3, 3, 24, 6        & 87.4  & 15.4    & 84.2     \\   %
    \bottomrule
    \end{tabular}    
    }
\end{minipage}
} 
\hspace{-1em}
\subfloat[
\textbf{Block configuration}
\label{tab:ablation_featuremixer}
]{
\begin{minipage}{0.33\linewidth}
\centering
 \tabcolsep=0.1em
    \resizebox{1\linewidth}{!}{
    \begin{tabular}{@{}l|ccc@{}}
    \toprule
    Block conf  &  Param &  FLOPs  &  {Top-1} \\ \midrule
    \gr (a) \ours-T       & 23.9      & 5.0     & 82.8   \\
    \quad (a) + more act    & 23.9 & 5.0 & 81.9 \\ %
    \quad (a) $\leftrightarrow$ 3$\stimes$3 dwconv           & 23.5      & 4.9     & 82.4   \\
    \quad (a) $\leftrightarrow$  5$\stimes$5 dwconv           & 23.7      & 5.0     & 82.5   \\
    \quad (a) $\leftrightarrow$  9$\stimes$9 dwconv           & 23.9      & 5.1     & 82.8   \\
    \quad (a) $\leftrightarrow$  11$\stimes$11 dwconv          & 24.4      & 5.2     & 82.7   \\ 
    \quad (a) $\leftrightarrow$  13$\stimes$13 dwconv         & 24.8      & 5.3     & 82.7   \\ 
        
    \midrule    
    (b) (a) $\leftrightarrow$  dwconv at last    & 23.6 & 5.0 & 82.3 \\ %
    \quad (b) + more act/norm & 23.6 & 5.0 & 81.1 \\ %
    \quad (b) LN $\leftrightarrow$ BN     & 23.6 & 5.0 & 82.2 \\  %
    \bottomrule
    \end{tabular}
    }
\end{minipage}
} 
\hspace{-1em}
\subfloat[
\textbf{Expansion ratio (ER)}
\label{tab:ablation_expansion_ratio}
]{
\begin{minipage}{0.32\linewidth}
\centering
\tabcolsep=0.8em
    \resizebox{00.9\linewidth}{!}{    
    \begin{tabular}{@{}c|cccc@{}}
    \toprule
       ER  &    Param & FLOPs & {Top-1}   \\ \midrule
      1.0      & 24.4  & 5.0 & 82.1         \\
      2.0      & 23.8  & 5.0 & 82.6      \\
      3.0      & 24.2  & 5.0 & 82.7       \\ 
    \gr 4.0    & 23.9  & 5.0 & 82.8     \\
      6.0      & 24.3  & 5.0 & 82.6     \\
    \bottomrule
    \end{tabular}
    }
\end{minipage}

}

\vspace{-.5em}
\subfloat[
\textbf{Growth rate (GR)}
\label{tab:ablation_same_gr}
]{
\tabcolsep=0.1cm
\centering
\begin{minipage}{0.35\linewidth}
\tabcolsep=0.2em
    \resizebox{0.9\linewidth}{!}{
    \begin{tabular}{@{}c|ccc@{}}
    \toprule
    GR     & Param & FLOPs & {Top-1}   \\ \midrule
    \:\:90, \:\:90, \:\:90, \:\:90    & 13.2  & 5.0  &  81.6     \\
    120, 120, 120, 120 & 23.9  & 8.9 &   83.0     \\
    \gr  \:\:64, 104, 128, 224 & 23.9  & 5.0    & 82.8     \\
    \bottomrule
    \end{tabular}
    }
\end{minipage}
}
\hspace{-2.0em}
\subfloat[
\textbf{Transition layer intervals}
\label{tab:ablation_transition_interval}
]{
\begin{minipage}{0.335\linewidth}
\centering
\tabcolsep=1em
    \resizebox{0.96\linewidth}{!}{
    \begin{tabular}{@{}c|ccc@{}}
    \toprule
    Interval &  Param &  FLOPs &  {Top-1} \\ 
    \midrule
    2        &  24.3  & 5.0    & 82.7   \\
    \gr 3    &  23.9  & 5.0    & 82.8   \\
    4        & 24.1  & 5.0     & 82.6   \\
    6        & 23.7  & 5.0     & 82.1   \\ 
    \bottomrule
    \end{tabular}
    }
\end{minipage}
} 
\hspace{-1em}
\subfloat[
\textbf{Transition ratio}
\label{tab:ablation_transition_ratio}
]{
\begin{minipage}{0.305\linewidth}
\centering
 \tabcolsep=0.8em
    \resizebox{00.9\linewidth}{!}{
    \begin{tabular}{@{}c|ccc@{}}
    \toprule
    Ratio          & Param & FLOPs & {Top-1}   \\  \midrule
    0.3       & 24.4  & 5.0    & 82.6     \\
    0.4       & 23.9  & 5.0    & 82.6     \\
    \gr 0.5   & 23.9  & 5.0    & 82.8     \\
    0.6       & 23.8  & 5.0   & 82.5     \\
    0.7       & 23.6  & 5.0     & 82.3     \\ 
    \bottomrule
    \end{tabular}
    }
\end{minipage}
}
\vspace{-2em}
\end{table*}

\vspace{-1em}
\subsection{Revisiting Stochastic Depth} \begin{wrapfigure}{r}{0.45\columnwidth}
\vspace{-3.7em}
\centering
\hspace{-1em}
\begin{minipage}{0.24\columnwidth}
    \includegraphics[width=1\textwidth]{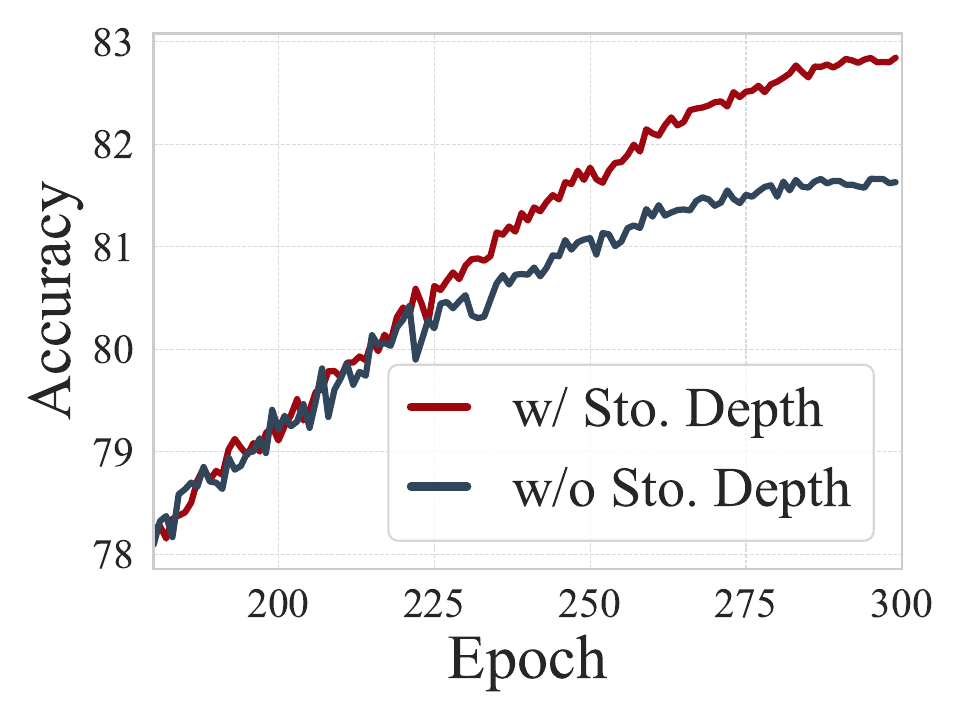}
\end{minipage}
    \hspace{-1em}
\begin{minipage}{0.24\columnwidth}
    \includegraphics[width=1\textwidth]{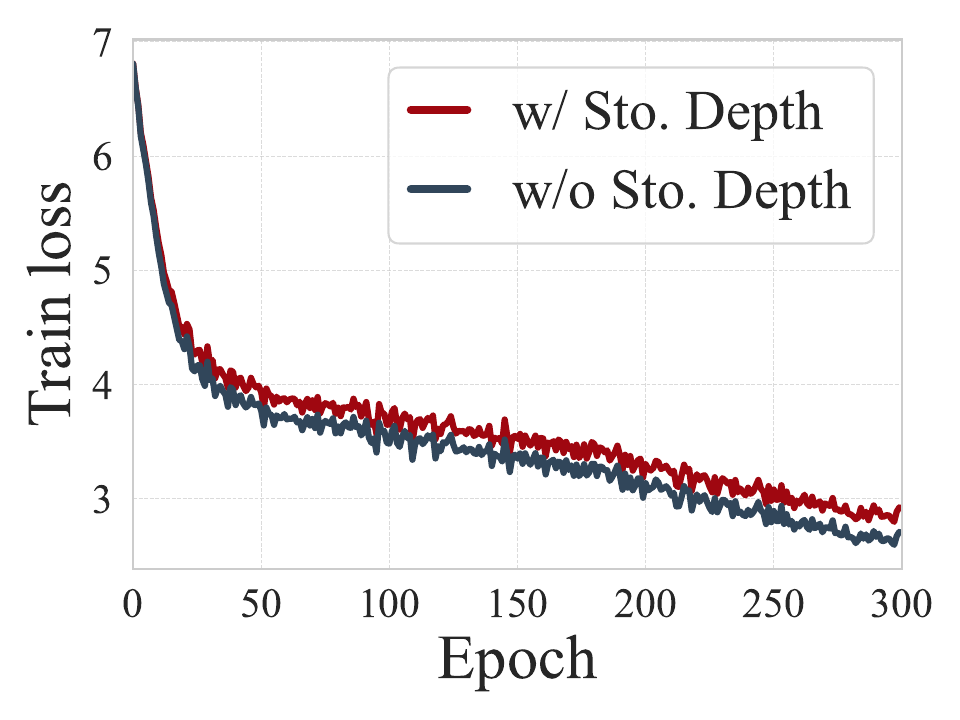}
\end{minipage}
    \vspace{-1.25em}
\caption{\textbf{Stochastic depth proves effective with dense connections.} It still acts as a regularizer.}
\label{fig:stodepth_impact}
\vspace{-2em}
\end{wrapfigure}
 
\vspace{-0.5em}
Notably, DenseNets primitively did not employ Stochastic Depth~\cite{huang2016deep} for model training due to sharing the similarity in connectivity patterns of networks. \begin{wraptable}{r}{12em}
\centering
\vspace{-5.5em}
\small
    \caption{ \textbf{Stochastic depth is compatible with dense connections.}}
    \label{tab:ablation_stodepth}
    \tabcolsep=0.7em
    \resizebox{1\linewidth}{!}{
    \begin{tabular}{c|ccc}
    \toprule
    Ratio  & Param  &  FLOPs  &  Top-1 \\ \midrule
    0     & 23.9      & 5.0        & 81.6   \\
    \midrule
    0.05     & 23.9      & 5.0        & 82.5    \\
    0.10     & 23.9      & 5.0        & 82.6   \\
    \gr 0.15 & 23.9      & 5.0    & 82.8   \\
    0.20     & 23.9      & 5.0    & 82.6    \\
    \bottomrule
    \end{tabular}
    }
\vspace{-5em}
\end{wraptable} 
We posit that Stochastic Depth should not be overlooked; our results demonstrate a noticeable improvement when it is incorporated into our model, as illustrated in Fig.~\ref{fig:stodepth_impact}. We also observe that a small stochastic depth ratio affects profoundly (see Table~\ref{tab:ablation_stodepth}).

\vspace{-1em}
\subsection{Ablation Studies} We gather all ablation studies in Table~\ref{tab:ablation_study}. Each table contains several models that are meticulously adjusted for almost equivalent computational costs with others to ensure a fair comparison of our specific focuses. Our methodology, in \S\ref{sec:method_revital}, methodically referenced each study.

\vspace{-0.5em}
\section{Conclusion}
\vspace{-0.5em}
In this paper, we have revisited the past success of DenseNet, which once outperformed ResNet in this era dominated by models using addition-based shortcuts, such as ResNet, ConvNeXt, and ViT. We have first rediscovered the potential of DenseNet, focusing on the underappreciated fact that DenseNet's concatenation shortcuts surpass the expressivity of the convention of ResNet-style addition-based shortcuts through our pilot study. We then highlight the outdated training setups and classical macro-block designs that diminish DenseNet's effectiveness against modernized architectures. %
By achieving our goal to widen DenseNet with modernized elements, we have proven that DenseNet's foundational principles are competitive in achieving robust modeling performance on their own. Our models exhibit strong performance competitive to the latest modern architectures; the employment of diverse concatenated features has significantly enhanced performance in dense prediction tasks, showcasing an advantage overlooked in models utilizing addition shortcuts. We hope that our work sheds light on the advantages of using concatenations in network design, advocating for the consideration of DenseNet-style architectures alongside ResNet-style ones. \\

\noindent\textbf{Limitations.} Our models have been scaled to a `-large' level, but resource limitation prevents more extensions to upper scales such as ViT-G.
\bibliographystyle{splncs04}
\bibliography{main}

\clearpage
\renewcommand\thefigure{\Alph{figure}}    
\setcounter{figure}{0}  
\renewcommand\thetable{\Alph{table}}
\setcounter{table}{0} 
\appendix

{\large \noindent\textbf{Appendix}}
\vspace{2.5em}

\noindent In this Appendix, we provide additional experiments and details to complement the main paper. The contents are as follows: 
\begin{itemize}
    \item \S\ref{sec:exp_setup} presents our experimental setups for ImageNet and downstream tasks training and evaluation setups;
    \item \S\ref{sec:dataflow} presents detailed layer configuration of \ours-T;
    \item \S\ref{sec:further_speed} evaluates further ImageNet top-accuracy versus latency trade-offs on various testbeds, encompassing both PyTorch and TensorRT A100 inference, as well as CPU inference outcomes;
    \item \S\ref{sec:apples_to_apples} conducts an ablation study by aligning the depth, parameters, and FLOPs with ResNet and ConvNeXt to reaffirm the strengths of dense connections;
    \item \S\ref{sec:more_od} reports further COCO~\cite{2014MicrosoftCOCO} object detection and instance segmentation results with Mask-RCNN~\cite{2017iccvmaskrcnn} and Cascade Mask-RCNN~\cite{tpami2019cascade};
    \item \S\ref{sec:transferability} benchmarks the transferability of ImageNet-1K pre-trained models. We utilize fine-grained classification datasets and long-tailed classification datasets, including CIFAR-10~\cite{09cifar}, CIFAR-100~\cite{09cifar}, Flowers-102~\cite{08flowers}, Stanford-Cars~\cite{13cars}, iNaturalist-2018~\cite{Horn2018INaturalist}, and iNaturalist-2019~\cite{Horn2019INaturalist};
    \item \S\ref{sec:gan} evaluates the effectiveness of dense connections on generative models;%
    \item \S\ref{sec:robustness} benchmarks robustness of the ImageNet-1K~\cite{imagenet} pre-trained models on the out-of-distribution datasets, including ImageNet-V2~\cite{Recht2019imagenetv2}, ObjectNet~\cite{nips19objectnet}, ImageNet-A~\cite{hendrycks2021imagenetA}, ImageNet-Sketch~\cite{wang2019imagenetSketch}, and ImageNet-R~\cite{hendrycks2021imagenetR};
    \item \S\ref{sec:pilot} gives more details of our pilot study described in \S\ref{sec:potential_concat} 
    with overall results, and scaled-up experiments on ImageNet-1K.
\end{itemize}

\newpage

\begin{table}[t!]
\centering
\caption{\textbf{ImageNet-1K training settings}. Most training setups are consistently used except for the multiple stochastic depth rates (e.g., 0.15/0.35/0.4/0.45) that regularize the corresponding models (e.g., \ours-T/S/B/L), respectively.}
\tabcolsep=0.6em
\resizebox{0.6\linewidth}{!}{
\begin{tabular}{l|cc}
\toprule
                                                        & \ours-T/S/B/L                   & \ours-L                         \\
                                                        & (Pre-)Training                   & Fine-Tuning                      \\ \midrule
image size                                              & 224                         & 384                         \\
weight init                                             & kaiming normal                  & pre-trained                      \\
optimizer                                               &  AdamW                           & AdamW                           \\
base learning rate                                      & 1e-3                            & 2e-5                            \\
weight decay                                            & 0.05                            & 1e-8                            \\
optimizer momentum ($\beta_1, \beta_2$)                                   & 0.9, 0.999 & 0.9, 0.999 \\
batch size                                              & 512                             & 512                             \\
training epochs                                         & 300                             & 30                              \\
learning rate schedule                                  & cosine decay                    & cosine decay                    \\
warmup epochs                                           & 20                              & 5                               \\
warmup schedule                                         & linear                          & linear                          \\
layer-wise lr decay \cite{clark2020electra,bao2021beit} & None                            & 0.7                             \\
randaugment \cite{cubuk2020randaugment}                 & (9, 0.5)                        & (9, 0.5)                        \\
mixup \cite{zhang2017mixup}                             & 0.8                             & 0.0                             \\
cutmix \cite{yun2019cutmix}                             & 1.0                             & 0.0                             \\
random erasing \cite{zhong2020random}                   & 0.25                            & 0.25                            \\
label smoothing \cite{cvpr2016inceptionv3}              & 0.1                             & 0.1                             \\
stochastic depth \cite{eccv2016droppath}                & 0.15/0.35/0.4/0.5               & 0.6                             \\
layer scale \cite{Touvron2021GoingDW}                   & 1e-6                            & pre-trained                      \\
head init scale \cite{Touvron2021GoingDW}               & None                            & 1e-3                           \\
gradient clip                                           & None                            & None                            \\
center crop percent                                     & 0.9                             & 1.0                             \\
exp. mov. avg. (EMA) \cite{siam1992ema}                 & None                            & None                            \\ \bottomrule
\end{tabular}
}
\label{suptab:train_settings_in1k}
\end{table}
\section{Experimental Settings}
\label{sec:exp_setup}
\subsection{ImageNet Training}
Table~\ref{suptab:train_settings_in1k} presents the training settings for \ours on ImageNet-1K. Each variant of \ours adheres to these settings, except for the stochastic depth rate~\cite{eccv2016droppath}, which is tailored to each model variant. For fine-tuning, we group three consecutive feature mixer blocks for layer-wise learning rate decay~\cite{bao2021beit, clark2020electra} akin to the approach taken in ConvNeXt. We use the timm package~\cite{timm} for model training.

\subsection{Downstream Tasks}
We adhere to the hyper-parameter sweep protocol outlined in~\cite{cvpr2022convnext} but sweep much lightly. For UperNet~\cite{eccv2018upernet} training on ADE20K, we explore the following hyperparameters: learning rate \{8e-5, 1e-3\}, weight decay \{0.01, 0.03, 0.05\}. For Mask-RCNN~\cite{2017iccvmaskrcnn} training on COCO, we explore hyperparameters such as learning rate \{1e-3, 2e-3, 3e-3\}, weight decay \{0.05, 0.1\}, stochastic depth \{0.1, 0.2\}. For Cascade Mask-RCNN~\cite{tpami2019cascade} training on COCO, we explore hyperparameters such as learning rate \{8e-5, 1e-4\}, weight decay \{0.05, 0.1\}, stochastic depth \{0.4, 0.5, 0.6\}, and layer-wise learning rate decay \{0.7, 0.8\}. We note that our search space is similar to or less extensive than the known search space in ConvNeXt~\cite{cvpr2022convnext} (\eg, 6 (ours) vs. 12 (ConvNeXt) for ADE20K (UperNet) and 8/24 (ours) vs. 48 (ConvNeXt) for COCO (Cascade Mask-RCNN) experiments).

\subsection{Benchmark Setup}
We measure latency and memory on the V100 GPU utilizing PyTorch 1.13.1 and CUDA 11.6. In all measurements, we employ the channels-last memory format~\cite{ChannelLast}. Memory is measured in the training phase with a batch size of 16.

\section{Model Configurations of \ours}
\label{sec:dataflow}
From a practitioner's perspective, it can be challenging to ascertain the number of channels of features of \ours. Therefore, we provide a detailed model configuration in Table~\ref{suptab:dataflow}. The values of resolution and channel are based on the output feature. We ensured that the output dimensions of the transition layers are multiples of 8 for efficiency.

\begin{table}[b!]
\caption{\textbf{Model configurations of RDNet.} The table on the left presents the layer-wise configuration of \ours-T. The table on the right details the configuration of the RDNet model family. As described in Fig.~\ref{fig:rdnet_architecture}, 
each stage of the RDNet comprises $L_N$ mixing blocks.}
\label{suptab:dataflow}
\centering
    \tabcolsep=0.45em
    \resizebox{0.44\linewidth}{!}{
    \begin{tabular}{ccccc}
    \toprule
    Stage                & GR                    & Layer          & Resolution                  & \#Channels \\
    \hline
                         &                       & Patchification & 56 $\times$ 56 & 64         \\ \hline
    \multirow{3}{*}{S1}  & \multirow{3}{*}{64}   & Feature mixer  & 56 $\times$ 56 & 128        \\
                         &                       & Feature mixer  & 56 $\times$ 56 & 192        \\
                         &                       & Feature mixer  & 56 $\times$ 56 & 256        \\ \hline
                         &                       & Transition S2  & 28 $\times$ 28 & 128        \\ \hline
    \multirow{3}{*}{S2}  & \multirow{3}{*}{104}  & Feature mixer  & 28 $\times$ 28 & 232        \\
                         &                       & Feature mixer  & 28 $\times$ 28 & 336        \\
                         &                       & Feature mixer  & 28 $\times$ 28 & 440        \\ \hline
                         &                       & Transition S2  & 14 $\times$ 14 & 216        \\ \hline
    \multirow{15}{*}{S3} & \multirow{15}{*}{128} & Feature mixer  & 14 $\times$ 14 & 344        \\
                         &                       & Feature mixer  & 14 $\times$ 14 & 472        \\
                         &                       & Feature mixer  & 14 $\times$ 14 & 600        \\
                         &                       & Transition S1  & 14 $\times$ 14 & 296        \\
                         &                       & Feature mixer  & 14 $\times$ 14 & 424        \\
                         &                       & Feature mixer  & 14 $\times$ 14 & 552        \\
                         &                       & Feature mixer  & 14 $\times$ 14 & 680        \\
                         &                       & Transition S1  & 14 $\times$ 14 & 336        \\
                         &                       & Feature mixer  & 14 $\times$ 14 & 464        \\
                         &                       & Feature mixer  & 14 $\times$ 14 & 592        \\
                         &                       & Feature mixer  & 14 $\times$ 14 & 720        \\
                         &                       & Transition S1  & 14 $\times$ 14 & 360        \\
                         &                       & Feature mixer  & 14 $\times$ 14 & 488        \\
                         &                       & Feature mixer  & 14 $\times$ 14 & 616        \\
                         &                       & Feature mixer  & 14 $\times$ 14 & 744        \\ \hline
                         &                       & Transition S2  & 7 $\times$ 7   & 368        \\ \hline
    \multirow{3}{*}{S4}  & \multirow{3}{*}{224}  & Feature mixer  & 7 $\times$ 7   & 592        \\
                         &                       & Feature mixer  & 7 $\times$ 7   & 816        \\
                         &                       & Feature mixer  & 7 $\times$ 7   & 1040       \\ \hline
                         &                       & Classifier     & 1 $\times$ 1   & 1000      \\
    \bottomrule
    \end{tabular}
    }
    \tabcolsep=0.55em    
    \resizebox{0.55\linewidth}{!}{
        \begin{tabular}{cc|cccc}
        \toprule
        \multicolumn{1}{c|}{Stage}               & Layer              & \multicolumn{4}{c}{RDNet}                                                                  \\ \cline{3-6} 
        \multicolumn{1}{c|}{}                    & Settings           & \multicolumn{1}{c|}{Tiny} & \multicolumn{1}{c|}{Small} & \multicolumn{1}{c|}{Base} & Large \\ \midrule

        \multicolumn{1}{c|}{\multirow{2}{*}{S1}} & Growth Rate        & \multicolumn{1}{c|}{64}   & \multicolumn{1}{c|}{64}    & \multicolumn{1}{c|}{96}   & 128    \\  
        \multicolumn{1}{c|}{}                    & \# Mixing Block    & \multicolumn{1}{c|}{1}    & \multicolumn{1}{c|}{1}     & \multicolumn{1}{c|}{1}    & 1     \\ \midrule
        \multicolumn{6}{c}{Transition S/2}                                                                        \\ \midrule
        \multicolumn{1}{c|}{\multirow{2}{*}{S2}} & Growth Rate        & \multicolumn{1}{c|}{104}  & \multicolumn{1}{c|}{128}   & \multicolumn{1}{c|}{128}  & 192   \\  
        \multicolumn{1}{c|}{}                    & \# Mixing Block    & \multicolumn{1}{c|}{1}    & \multicolumn{1}{c|}{1}     & \multicolumn{1}{c|}{1}    & 1     \\ \midrule
        \multicolumn{6}{c}{Transition S/2}                                                                        \\ \midrule
        \multicolumn{1}{c|}{\multirow{2}{*}{S3}} & Growth Rate        & \multicolumn{1}{c|}{128}  & \multicolumn{1}{c|}{128}   & \multicolumn{1}{c|}{168}  & 256   \\  
        \multicolumn{1}{c|}{}                    & \# Mixing Block    & \multicolumn{1}{c|}{4}   & \multicolumn{1}{c|}{7}    & \multicolumn{1}{c|}{7}   & 8    \\ \midrule
        \multicolumn{6}{c}{Transition S/2}                                                                        \\ \midrule
        \multicolumn{1}{c|}{\multirow{2}{*}{S4}} & Growth Rate        & \multicolumn{1}{c|}{192}  & \multicolumn{1}{c|}{240}   & \multicolumn{1}{c|}{336}  & 360   \\  
        \multicolumn{1}{c|}{}                    & \# Mixing Block    & \multicolumn{1}{c|}{1}    & \multicolumn{1}{c|}{2}     & \multicolumn{1}{c|}{2}    & 2     \\ \midrule
        \multicolumn{2}{c|}{Classifier}                               & \multicolumn{4}{c}{GAP, Linear}                                         \\ \midrule
        \multicolumn{2}{c|}{Parameters (M)}                           & \multicolumn{1}{c|}{24} & \multicolumn{1}{c|}{50}  & \multicolumn{1}{c|}{87} & 186  \\ \midrule
        \multicolumn{2}{c|}{FLOPs (G)}                                & \multicolumn{1}{c|}{5.0} & \multicolumn{1}{c|}{8.7}  & \multicolumn{1}{c|}{15.4} & 34.7  \\ \bottomrule
        \vspace{4.5em}
        \end{tabular}
    }
\end{table}

\begin{figure}[t!]
    \centering
    \begin{subfigure}{\linewidth}
        \centering
        \begin{subfigure}{.3\linewidth}
            \includegraphics[width=\textwidth]{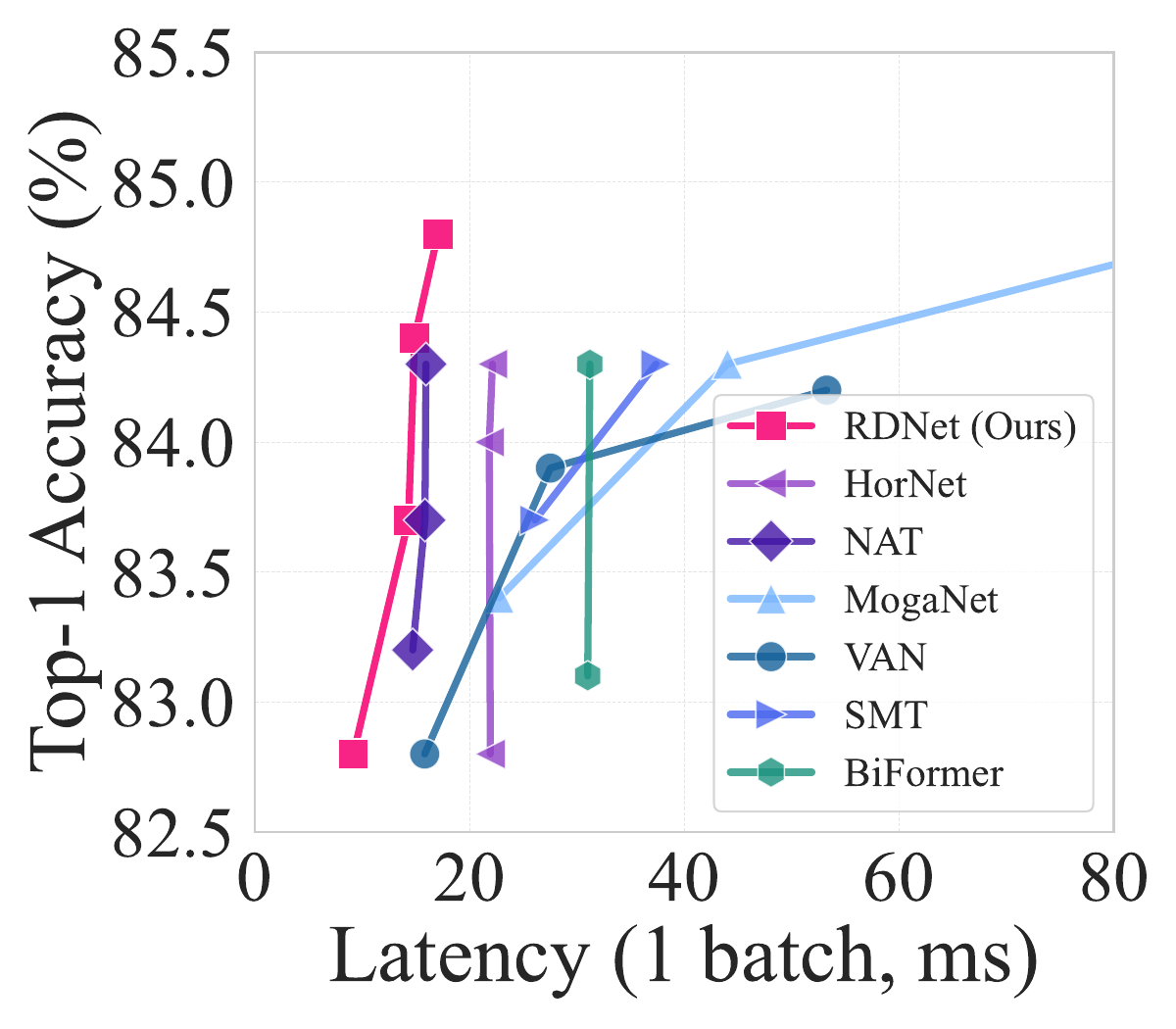} 
        \end{subfigure}
        \begin{subfigure}{.3\linewidth}
            \includegraphics[width=\textwidth]{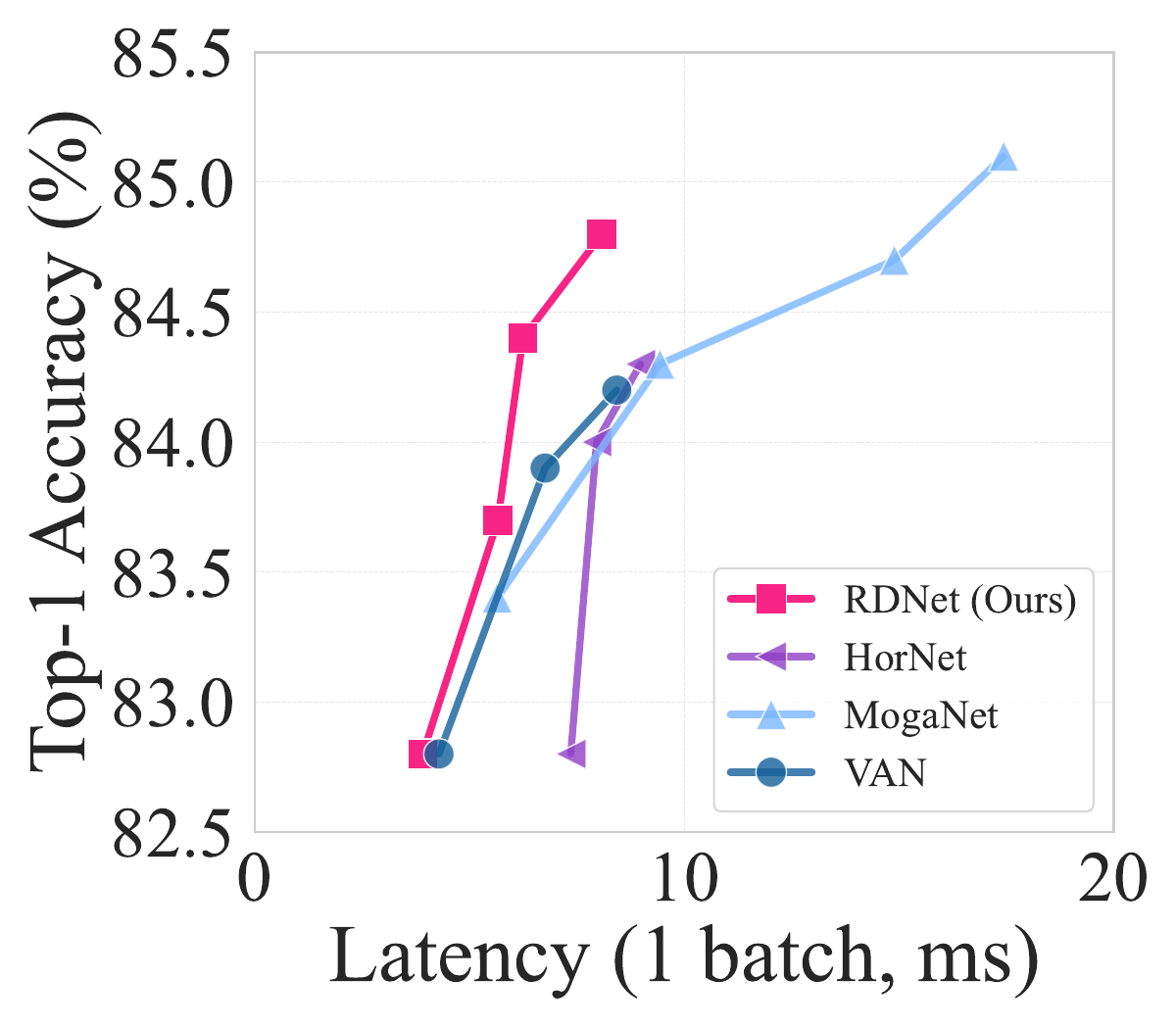}
        \end{subfigure}
        \begin{subfigure}{.3\linewidth}
        \includegraphics[width=\textwidth]{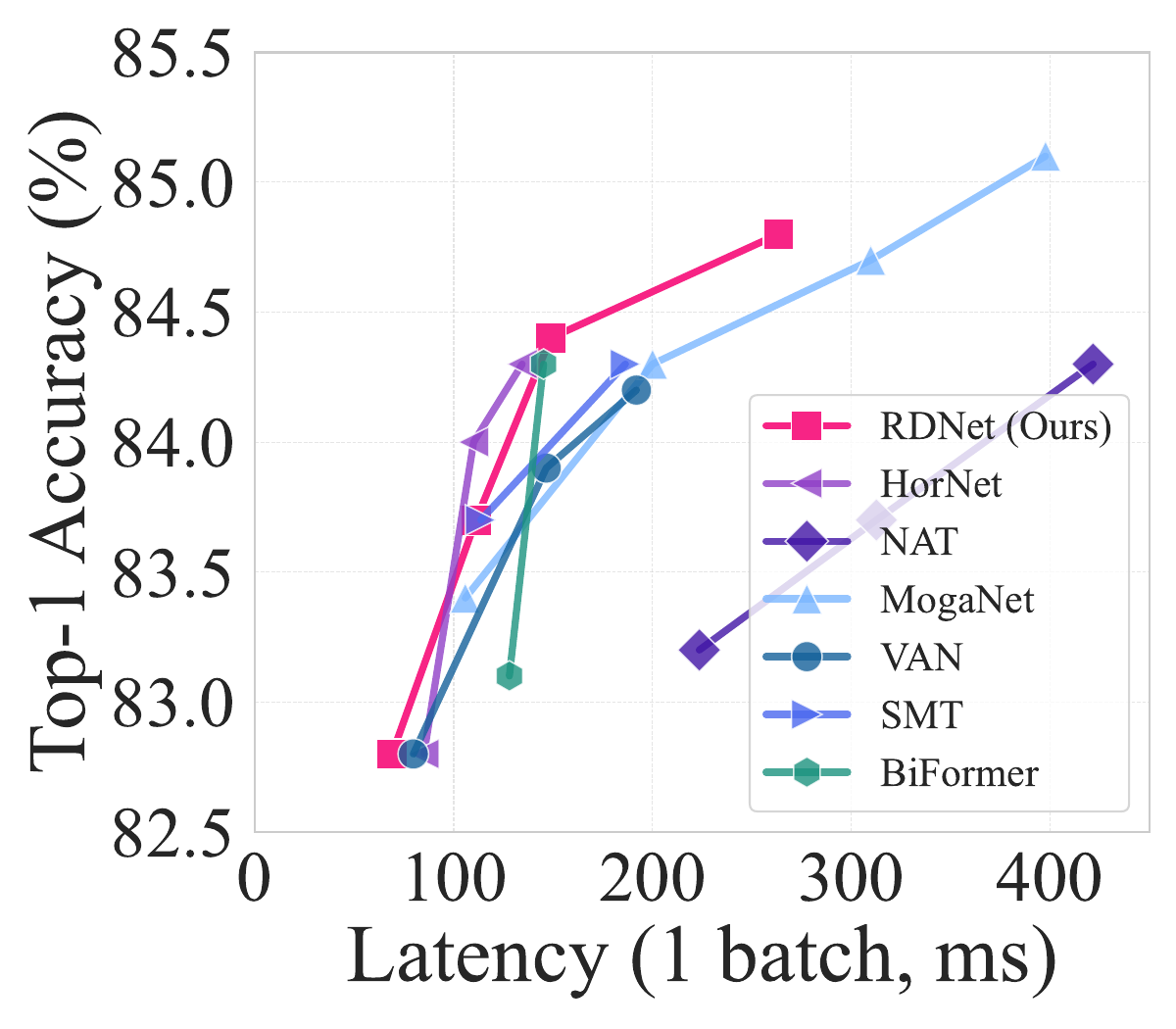}
        \end{subfigure}
        \begin{subfigure}{.3\linewidth}
            \includegraphics[width=\textwidth]{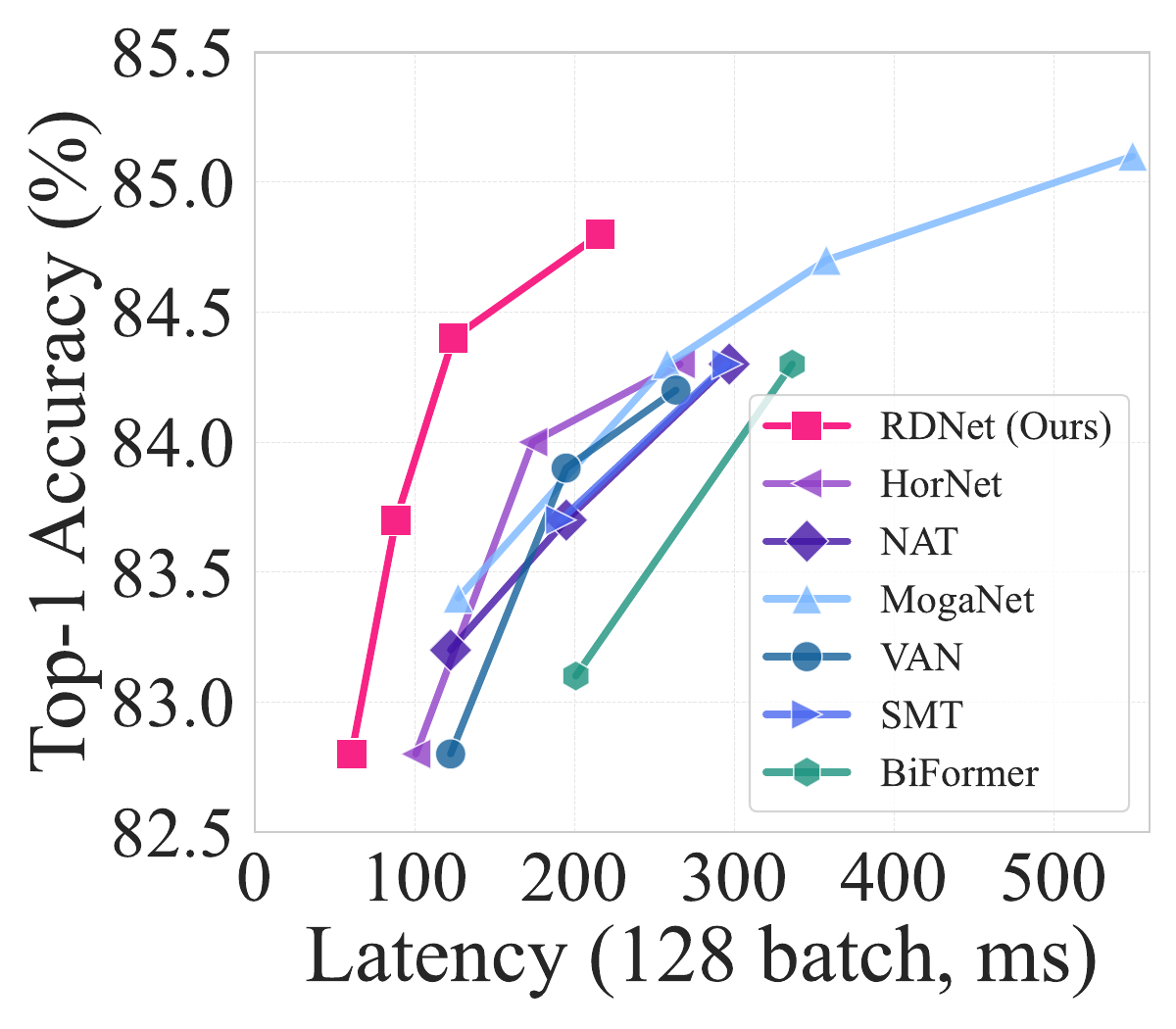}
        \caption{PyTorch {(A100 GPU)}}
        \end{subfigure}
        \begin{subfigure}{.3\linewidth}
            \includegraphics[width=\textwidth]{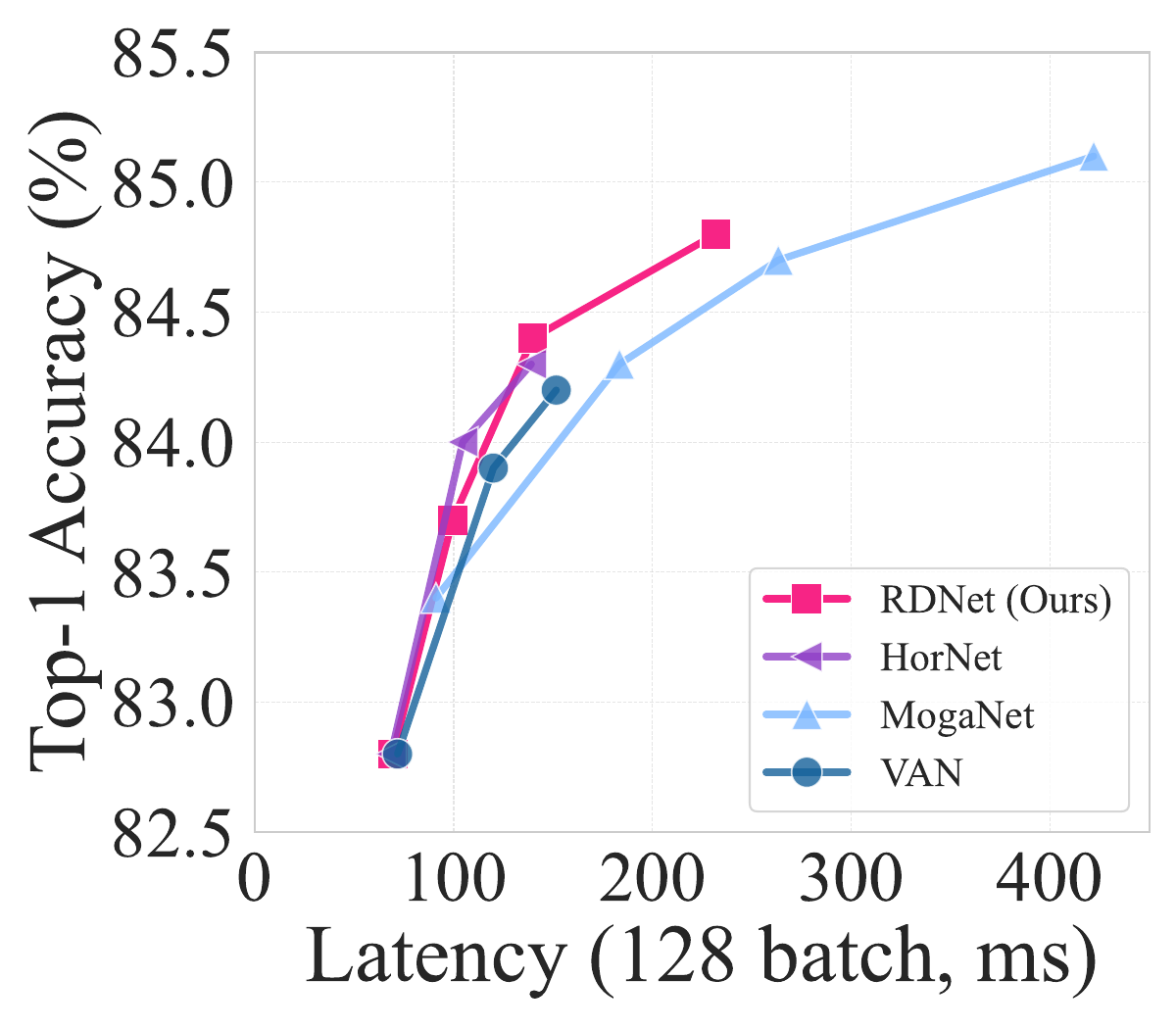}
            \caption{TensorRT {(A100 GPU)}}
        \end{subfigure}
        \begin{subfigure}{.3\linewidth}
            \includegraphics[width=\textwidth]{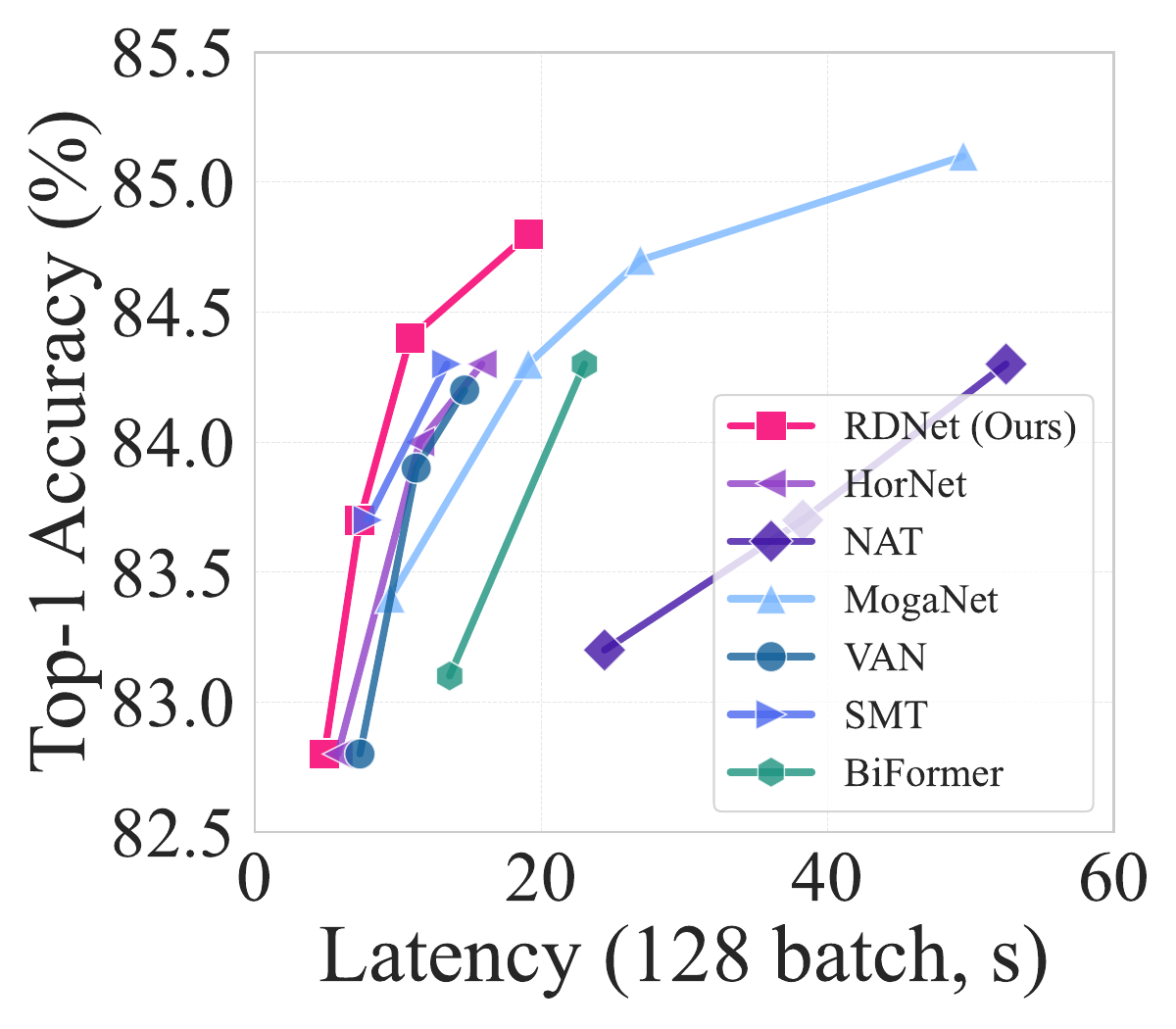}
        \caption{CPU {(Xeon Gold 5120)}}
        \end{subfigure}
    \end{subfigure}        
    \caption{\textbf{Further trade-offs in ImageNet-1K performance.} We provide comparative visualizations with diverse environments (including A100, TesorRT, and CPU) between \textit{state-of-the-art models}, which were known for top-performing models. It turns out that \ours is highly competitive in practice in terms of model speed.}
    \label{supfig:latency_acc_with_sota}
\end{figure}

\section{More ImageNet Accuracy vs. Latency Trade-offs}
\label{sec:further_speed}
To assess the practicality of our models, we measure speeds across diverse testbeds. We precisely measure inference speeds using the PyTorch framework on NVIDIA A100 GPU, Intel Xeon Gold 5120 CPU, and TensorRT Inference Engine on NVIDIA A100 GPU. Our testing environment incorporates PyTorch version 1.13.1, CUDA version 11.6, and TensorRT version 8.5.3 for these experiments. Fig.~\ref{supfig:latency_acc_with_sota} shows that our models consistently show superior accuracy vs. latency trade-offs across all evaluation setups. Note that (b) in Fig.~\ref{supfig:latency_acc_with_sota} includes fewer models because those that are challenging to convert to inference engines due to factors like the use of CUDA custom kernels are not evaluated. We report all the numbers in Table~\ref{suptab:in1k_latest_a100}.

\begin{table}[ht!]
\small
\centering
\caption{\textbf{ImageNet-1K comparison with the latest models.} Fig.~\ref{supfig:latency_acc_with_sota} visualized this table. We thoroughly compare our models against the latest architectures in practical latencies. b$n$ denotes latency, measured with a batch size of $n$. Certain models were excluded from evaluation because their CUDA custom kernels were not compiled. }
\label{suptab:in1k_latest_a100}
\tabcolsep=0.15em
\resizebox{1.0\linewidth}{!}{
\begin{tabular}{lc|ccc|rrrr|rrrr|rrrr}
\toprule
Model                                       & Date        &Param & FLOPs & Top-1  & \multicolumn{4}{c|}{PyTorch (A100, ms)}  & \multicolumn{4}{c|}{TensorRT (A100, ms)} & \multicolumn{4}{c}{PyTorch (Xeon 5120, s)}\\
                                            &             & (M)  & (G)   & (\%)   & b1   & b8    & b32   & b128  & b1   & b8    & b32 & b128 & b1   & b8    & b32 & b128 \\ 
\midrule
\gr \ours-T                                 & Ours          & 24    & 5.0   & 82.8  & 9.2 & 9.2 & 17.8 & 60.5 & 3.9 & 7.5 & 19.7 & 69.3 & 0.07 & 0.25 & 1.09 & 4.85 \\ 
HorNet-T$_{7\times 7}$~\cite{nips2022hornet}& NeurIPS'2022  & 22    & 4.0   & 82.8  & 21.9 & 21.9 & 27.8 & 100.7 & 7.4 & 10.6 & 22.8 & 68.4 & 0.09 & 0.21 & 0.89 & 5.79 \\ 
VAN-B2~\cite{cvmj2023VAN}                   & CVMJ'2023     & 27    & 5.0   & 82.8  & 15.8 & 16.4 & 32.9 & 122.5 & 4.1 & 8.6 & 21.2 & 71.8 & 0.08 & 0.38 & 1.61 & 7.33 \\ 
BiFormer-S~\cite{zhu2023biformer}           & CVPR'2023     & 26    & 4.5   & 83.8  & 31.0 & 35.3 & 54.2 & 200.9 & - & - & - & - & 0.13 & 0.39 & 2.17 & 13.60 \\ 
NAT-T~\cite{hassani2023NAT}                 & CVPR'2023     & 28    & 4.3   & 83.2  & 14.7 & 14.7 & 32.9 & 122.4 & - & - & - & - & 0.22 & 1.43 & 5.73 & 24.42 \\ 
SMT-S~\cite{Weifeng2023smt}                 & ICCV'2023     & 21    & 4.7   & 83.7  & 26.1 & 26.4 & 53.0 & 191.4 & - & - & - & - & 0.11 & 0.32 & 1.16 & 7.91 \\ 
MogaNet-S~\cite{anonymous2024moganet}       & ICLR'2024     & 25    & 5.0   & 83.4  & 22.7 & 22.7 & 34.9 & 127.2 & 5.6 & 10.3 & 27.6 & 91.0 & 0.11 & 0.42 & 1.97 & 9.46  \\ 
\midrule
\gr \ours-S                                 & Ours          & 50    & 8.7   & 83.7  & 14.3 & 14.4 & 26.4 & 88.3 & 5.7 & 10.8 & 29.3 & 99.4 & 0.11 & 0.38 & 1.73 & 7.34    \\ 
HorNet-S$_{7\times 7}$~\cite{nips2022hornet}& NeurIPS'2022  & 50    & 8.8   & 84.0  & 21.8 & 22.2 & 46.9 & 173.9 & 8.0 & 14.1 & 33.6 & 104.5 & 0.11 & 0.38 & 2.43 & 11.51 \\ 
VAN-B3~\cite{cvmj2023VAN}                   & CVMJ'2023     & 45    & 9.0   & 83.9  & 27.5 & 32.8 & 50.4 & 194.8 & 6.8 & 13.7 & 34.5 & 119.9 & 0.15 & 0.59 & 2.50 & 11.27 \\ 
BiFormer-B~\cite{zhu2023biformer}           & CVPR'2023     & 57    & 9.8   & 84.3  & 31.2 & 31.1 & 89.3 & 336.3 & - & - & - & -  & 0.15 & 0.78 & 4.90 & 23.03 \\ 
NAT-S~\cite{hassani2023NAT}                 & CVPR'2023     & 51    & 7.8   & 83.7  & 15.8 & 20.6 & 51.9 & 194.9 & - & - & - & -  & 0.31 & 2.10 & 8.78 & 38.27 \\ 
SMT-B~\cite{Weifeng2023smt}                 & ICCV'2023     & 32    & 7.7   & 84.3  & 37.4 & 39.1 & 81.2 & 295.6 & - & - & - & -  & 0.19 & 0.60 & 2.24 & 13.41 \\ 
MogaNet-B~\cite{anonymous2024moganet}       & ICLR'2024     & 44    & 9.9   & 84.3  & 44.0 & 47.4 & 70.6 & 258.0 & 9.4 & 19.0 & 53.4 & 183.4 & 0.20 & 0.77 & 4.08 & 19.10 \\ 
\midrule
\gr \ours-B                                 & Ours          & 87     & 15.4  & 84.4  & 14.9 & 15.1 & 36.0 & 124.2 & 6.2 & 13.6 & 39.5 & 139.8 & 0.15 & 0.58 & 2.52 & 10.84 \\ 
HorNet-B$_{7\times 7}$~\cite{nips2022hornet}& NeurIPS'2022  & 87     & 15.6  & 84.3  & 22.1 & 22.3 & 68.9 & 266.1 & 9.0 & 16.5 & 41.8 & 139.1 & 0.13 & 0.57 & 3.43 & 15.86 \\ 
VAN-B4~\cite{cvmj2023VAN}                   & CVMJ'2023     & 60     & 12.2  & 84.2  & 53.3 & 53.6 & 80.7 & 263.6 & 8.4 & 17.0 & 45.1 & 151.6 & 0.19 & 0.70 & 3.26 & 14.65 \\ 
NAT-B~\cite{hassani2023NAT}                 & CVPR'2023     & 90     & 13.7  & 84.3  & 15.9 & 22.5 & 88.6 & 296.9 & - & - & - & - & 0.42 & 3.03 & 12.33 & 52.47 \\ 
MogaNet-L~\cite{anonymous2024moganet}       & ICLR'2024     & 83     & 15.9  & 84.7  & 81.6 & 90.3 & 98.8 & 357.7 & 14.9 & 28.7 & 77.4 & 263.2 & 0.31 & 1.08 & 51.56 & 26.93\\ 
\midrule
\gr \ours-L                                 & Ours          & 186    & 34.7  & 84.8  & 17.0 & 17.3 & 59.8 & 216.1 & 8.1 & 20.4 & 62.9 & 231.8 & 0.26 & 1.15 & 4.72 & 19.13 \\ 
MogaNet-XL~\cite{anonymous2024moganet}      & ICLR'2024     & 181    & 34.5  & 85.1  & 82.3 & 93.3 & 146.3 & 549.5 & 17.4 & 38.6 & 115.4 & 421.9 & 0.40 & 2.57 & 10.27 & 49.49 \\ 
\bottomrule
\end{tabular}
}
\end{table}

\section{More Comparison with ResNets and ConvNeXts}
\label{sec:apples_to_apples}
We presented the performance of RDNets matched with ResNets and ConvNeXts in terms of the number of parameters and FLOPs in Table~\ref{tab:ablation_depth_vs_wide}. 
Here, we further report some RDNets' performances, which are more closely aligned with ResNets and ConvNeXts than our previous models by 1) having identical depth; 2) having the same number of building blocks in each stage; 3) aligning the parameters and FLOPs as closely as possible with those of ResNet and ConvNext. Therefore, the comparison is more like an apples-to-apples comparison. As shown in Table~\ref{suptab:apples_to_apples}, despite \ours not having the optimal width and depth for dense connections, \ours demonstrates superior performance compared to the competing models. 
\begin{table}[t!]
\centering
\caption{\textbf{Apples-to-apples comparison with ResNet and ConvNeXt.} We align the parameters, FLOPs, and depth of ResNet and ConvNext as closely as possible to compare their top-1 accuracy (\%).}
\label{suptab:apples_to_apples}
    \tabcolsep=.8em
    \resizebox{0.65\linewidth}{!}{
    \begin{tabular}{@{}c|ccc|c@{}}
    \toprule
    Model       & Params & FLOPs & Depth & Top-1  \\ 
    \midrule
    ResNet-50   & 25.6M   & 4.1G  & [3,4,6,3] & 78.8        \\
    RDNet       & 25.5M & 4.1G & [3,4,6,3] &\textbf{82.1}        \\
    \midrule
    ResNet-152  & 60.2M   & 11.5G & [3,8,36,3] & 80.8        \\
    RDNet       & 59.7M & 11.5G & [3,8,36,3] &\textbf{83.7}        \\ 
    \midrule
    ConvNeXt-T  & 28.6M   & 4.5G & [3,3,9,3] & 82.1        \\
    RDNet       & 27.1M & 4.5G & [3,3,9,3] &\textbf{82.4}        \\
    \midrule
    ConvNeXt-B  & 88.6M   & 15.4G & [3,3,27,3] & 83.8       \\
    RDNet       & 88.3M & 15.3G & [3,3,27,3] &\textbf{84.1}        \\
    \bottomrule
    \end{tabular}
    }
\end{table}

\begin{table}[t]
\centering
\caption{\textbf{COCO object detection and segmentation results - Mask-RCNN 1x schedule.} FLOPs (G) are calculated with image size (1280, 800). }
\vspace{-0.5em}
\tabcolsep=0.3em
\resizebox{.9\linewidth}{!}{
\begin{tabular}{@{}lcccccccc@{}}
\toprule
Backbone & Param & FLOPs & $\text{AP}^{\text{box}}$ & $\text{AP}^{\text{box}}_{50}$ & $\text{AP}^{\text{box}}_{75}$ & $\text{AP}^{\text{mask}}$ & $\text{AP}^{\text{mask}}_{\text{50}}$ & $\text{AP}^{\text{mask}}_{75}$  \\
    \midrule
    PVT-S~\cite{iccv2021PVT} & 44M & 245G & 40.4 & 62.9 & 43.8 & 37.8 & 60.1 & 40.3 \\  %
    Swin-T~\cite{liu2021swin} & 48M & 264G & 43.6 & 66.2 & 47.7 & 39.6 & 62.9 & 42.2 \\  %
    PVTv2-B2~\cite{cvmj2022PVTv2} & 45M & 309G & 45.3 & 67.1 & 49.6 & 41.2 & 64.2 & 44.4 \\
    ConvNeXt-T~\cite{cvpr2022convnext} & 48M & 262G & 43.5 & 65.6 & 48.0 & 39.7 & 62.5 & 42.7 \\  %
    CSwin-T~\cite{cvpr2022CSWin} & 42M & 279G & 46.7 & 68.6 & 51.3 & 42.2 & 65.6 & 45.4 \\ 
    FocalNet-S (SRF)~\cite{nips2022focalnet} & 49M & 267G & 45.9 & 68.3 & 50.1 & 41.3 & 65.0 & 44.3 \\
    \gr RDNet-T  & 43M & 278G & 46.1 & 68.0 & 50.8 & 41.4 & 65.1 & 44.2  \\
    \midrule
    PVT-M~\cite{iccv2021PVT} & 64M & 302G & 42.0 & 64.4 & 45.6 & 39.0 & 61.6 & 42.1 \\  %
    Swin-S~\cite{liu2021swin} & 69M & 354G & 46.5 & 68.7 & 51.3 & 42.1 & 65.8 & 45.2 \\  %
    PVTv2-B3~\cite{cvmj2022PVTv2} & 65M & 397G & 47.0 & 68.1 & 51.7 & 42.5 & 65.7 & 45.7 \\
    ConvNeXt-S~\cite{cvpr2022convnext} & 70M & 348G & 46.8 & 69.0 & 51.5 & 42.1 & 65.8 & 45.2 \\  %
    CSwin-S~\cite{cvpr2022CSWin} & 54M & 342G & 47.9 & 70.1 & 52.6 & 43.2 & 67.1 & 46.2 \\ 
    FocalNet-S (SRF)~\cite{nips2022focalnet} & 71M & 356G & 48.0 & 69.9 & 52.7 & 42.7 & 66.7 & 45.7 \\
    \gr RDNet-S  & 70M & 354G & 48.2 & 69.9 & 53.0 & 43.0 & 66.9 & 46.3  \\
    \midrule
    Swin-B~\cite{liu2021swin} & 107M & 496G & 46.9 & 69.2 & 51.6 & 42.3 & 66.0 & 45.5 \\  %
    PVTv2-B5~\cite{cvmj2022PVTv2} & 102M & 557G & 47.4 & 68.6 & 51.9 & 42.5 & 65.7 & 46.0 \\
    ConvNeXt-B~\cite{cvpr2022convnext} & 108M & 486G & 47.5 & 69.9 & 51.9 & 42.5 & 66.8 & 45.7 \\  %
    CSwin-B~\cite{cvpr2022CSWin} & 97M & 526G & 48.7 & 70.4 & 53.9 & 43.9 & 67.8 & 47.3 \\ 
    FocalNet-B (SRF)~\cite{nips2022focalnet} & 109M & 496G & 48.8 & 70.7 & 53.5 & 43.3 & 67.5 & 46.5 \\
    \gr RDNet-B  & 107M & 493G & 48.8 & 70.4 & 53.5 & 43.4 & 67.5 & 46.6  \\
\bottomrule
\end{tabular}
 }
\label{suptab:coco_1x}
\end{table}

\begin{table}[ht!]
\centering
\caption{\textbf{COCO object detection and segmentation results - Cascade Mask-RCNN 3x schedule.} FLOPs (G) are calculated with image size (1280, 800). The result of Swin-T is from the official repository~\cite{github_swin_detection}. }
\vspace{-0.5em}
\tabcolsep=0.35em
\resizebox{.85\linewidth}{!}{
\begin{tabular}{@{}lcccccccc@{}}
\toprule
Backbone & Param & FLOPs & $\text{AP}^{\text{box}}$ & $\text{AP}^{\text{box}}_{50}$ & $\text{AP}^{\text{box}}_{75}$ & $\text{AP}^{\text{mask}}$ & $\text{AP}^{\text{mask}}_{\text{50}}$ & $\text{AP}^{\text{mask}}_{75}$  \\ \hline
Swin-T~\cite{liu2021swin}                    & 86M  & 745G & 50.4 & 69.2 & 54.7 & 43.7 & 66.6 & 47.3 \\
PVTv2-B2~\cite{cvmj2022PVTv2}                & 83M  & 788G & 51.1 & 69.8 & 55.3 & -    & -    & -    \\
FocalNet-T~\cite{nips2022focalnet}           & 86M  & 746G & 51.5 & 70.1 & 55.8 & -    & -    & -    \\
ConvNeXt-T~\cite{cvpr2022convnext}           & 86M  & 741G & 50.4 & 69.1 & 54.8 & 43.7 & 66.5 & 47.3 \\
NAT-T~\cite{hassani2023NAT}                  & 85M  & 737G & 51.4 & 70.0 & 55.9 & 44.5 & 67.6 & 47.9 \\

\gr \ours-T                                  & 81M  & 757G & 51.6 & 70.5 & 56.0 & 44.6 & 67.9 & 48.3 \\ \midrule
Swin-S~\cite{liu2021swin}                    & 107M & 838G & 51.9 & 70.7 & 56.3 & 45.0 & 68.2 & 48.8 \\
ConvNeXt-S~\cite{cvpr2022convnext}           & 108M & 827G & 51.9 & 70.8 & 56.5 & 45.0 & 68.4 & 49.1 \\
NAT-S~\cite{hassani2023NAT}                  & 108M & 809G & 52.0 & 70.4 & 56.3 & 44.9 & 68.1 & 48.6 \\
\gr \ours-S                                  & 108M & 832G & 52.3 & 70.8 & 56.6 & 45.4 & 68.5 & 49.3 \\ \midrule
Swin-B~\cite{liu2021swin}                    & 145M & 982G & 51.9 & 70.5 & 56.4 & 45.0 & 68.1 & 48.9 \\
ConvNeXt-B~\cite{cvpr2022convnext}           & 146M & 964G & 52.7 & 71.3 & 57.2 & 45.6 & 68.9 & 49.5 \\
NAT-B~\cite{hassani2023NAT}                  & 147M & 931G & 52.5 & 71.1 & 57.1 & 45.2 & 68.6 & 49.0 \\
\gr \ours-B                                  & 144M & 971G & 52.9 & 71.5 & 57.2 & 46.0 & 69.1 & 50.0 \\
\bottomrule
\end{tabular}
 }
\label{suptab:coco}
\vspace{-1em}
\end{table}

\section{More Object Detection/Instance Segmentation results}
\label{sec:more_od}
An extension to the Mask-RCNN~\cite{2017iccvmaskrcnn} with 3x schedule~\cite{he2019rethinking} results in Table~\ref{tab:coco} 
in the main paper, we additionally train Mask-RCNN using a 1x schedule~\cite{he2019rethinking} to perform a more comprehensive and fair comparison with a broader range of models. As shown in Table~\ref{suptab:coco_1x}, pre-trained \ours models demonstrate competitive performance. Furthermore, we employ the Cascade Mask-RCNN head~\cite{tpami2019cascade} to evaluate our pre-trained models. As demonstrated in Table~\ref{suptab:coco}, \ours exhibits competitive performance. Note that our models do not experience exhaustive fine-tuning of training hyperparameters for maximum precisions compared with ConvNeXt's, which indicates additional potential for achieving higher accuracy.

\section{Transferability Evaluation}
\label{sec:transferability}
We benchmark the transferability of ImageNet-1K pre-trained models. We utilize fine-grained classification datasets - CIFAR-10~\cite{09cifar}, CIFAR-100~\cite{09cifar}, Flowers-102~\cite{08flowers}, and Stanford-Cars~\cite{13cars}. Furthermore, to verify the ability of long-tailed classification, we utilize iNaturalist-2018~\cite{Horn2018INaturalist} and iNaturalist-2019~\cite{Horn2019INaturalist}.  We follow the training recipe of DeiT~\cite{icml2021deit}. As demonstrated in Table~\ref{suptab:fgvc}, \ours exhibits strong transferability. Notably, \ours delivers impressive performance on long-tailed classification tasks. \ours-T surpasses DeiT-III-L on iNaturalist-2018 and iNaturalist-2019.

\begin{table*}[t!]
\scriptsize
    \caption{Top-1 accuracy (\%), parameter count (M), and FLOPs (G) on various classification tasks with ImageNet-1k pre-trained models.}
    \label{suptab:fgvc}
    \centering
    \begin{tabular}{l|cc|cccc|cc}
    \toprule
    Model    & Param & FLOPs & CIFAR10 & CIFAR100  & Flowers & Cars & iNat18 & iNat19  \\
    \midrule                                                                          
    Grafit ResNet-50~\cite{Touvron2020GrafitLF} & 26 & 4.1 & -   & - & 98.2 & 92.5 & 69.8 & 75.9 \\
    DeiT-III-S~\cite{eccv2022deit3} & 22 & 4.6 & 98.9 & 90.6 & 96.4 & 89.9 & 67.1 & 72.7 \\
    \gr RDNet-T & 24 & 5.0 & 98.9 & 90.4 & 98.6 & 93.9 & 77.0 & 81.2 \\ 
    \midrule
    ResNet-152~\cite{Chu2020FeatureSA}           & 60 & 11.6 & - & - & - & - & 69.1 & - \\
    \gr RDNet-S & 50 & 8.7 & 99.0 & 91.1 & 98.4 & 94.2 & 79.1 & 82.4 \\ 
    \midrule
    ViT-B/16~\cite{vit}         & 86 & 35.1 & 98.1 & 87.1 & 89.5 & -   & - & -  \\
    ViT-B/16~\cite{steiner2021train}         & 86 & 35.1 & - & 87.8 & 96.0 & -   & - & -  \\
    DeiT-B~\cite{icml2021deit}  & 87 & 17.5 & 99.1 & 90.8 & 98.4  & 92.1 & 73.2 &  77.7 \\
    DeiT-III-B~\cite{eccv2022deit3} & 87 & 17.5& 99.3 & 92.5 & 98.6 & 93.4 & 73.6 & 78.0\\ 
    \gr RDNet-B & 87 & 15.4 & 99.3 & 91.4 & 98.6 & 94.1 & 80.5 & 83.5 \\ 
    \midrule
    \gr RDNet-L & 186 & 34.7 & 99.3 & 91.5 & 98.6 & 94.2 & 81.5 & 83.7 \\ 
    \midrule
    ViT-L/16~\cite{vit}         & 303 & 122.9 & 97.9 & 86.4 & 89.7 & -   & - & -  \\
    ViT-L/16~\cite{steiner2021train}         & 303 & 122.9 & - & 86.2 & 91.4 & -   & - & -  \\
    DeiT-III-L~\cite{eccv2022deit3} & 304 & 61.6 & 99.3 & 93.4 & 98.9 & 94.5 & 75.6 & 79.3 \\ 
    \bottomrule
    \end{tabular}
\end{table*}

\begin{table}[b]
\centering
\caption{\textbf{WGAN experiments}. 
We demonstrate the capability of dense connections in \ours for generation tasks. We utilize the WGAN architecture, redesigning only the generator to showcase whether the generation ability could improve.}
\label{suptab:gan}
    \tabcolsep=0.5em
    \begin{tabular}{@{}c|c|cc@{}}
    \toprule
    Model    & Param & IS↑   & FID↓       \\ \midrule
    ResBlock &  5.4M & 7.27±0.26 & 27.79±1.06 \\
    Ours & 5.6M & \textbf{7.52±0.10} & \textbf{25.37±0.21}  \\
    \bottomrule
    \end{tabular}
\end{table} 

\section{Dense Connections for Generative Models}
\label{sec:gan}
We test a generative model, WGAN~\cite{wgan}, by replacing ResNet-GAN in the generator with our concatenation-based model. We maintained identical discriminator architecture, training setups, and similar model sizes. We focus on a proof of concept on CIFAR-10 to showcase whether our models could outperform with faster convergence. Table~\ref{suptab:gan} results confirmed that our model works across multiple runs. We plan to extend our research to larger models in future work.

\begin{table}[ht!]
\centering
\caption[caption]{\textbf{Robustness evaluation.} We compare the models evaluating the out-of-distribution (OOD)
metrics ImageNet-V2/A/Sketch/ObjNet/R. We further average the OOD scores to show the averaged distribution shifts denoted by \textbf{Avg Shift}. Interestingly, even when RDNet demonstrates lower accuracy on ImageNet-1K compared to other networks, it consistently attains high OOD scores compared with other datasets.}
\label{suptab:robustness}
\tabcolsep=0.4em
\resizebox{0.85\linewidth}{!}{
\begin{tabular}{l|cc|cc|ccccc}
    \toprule
    Model       & Param & FLOPs  & IN & Avg Shift & V2 & Obj & A & Sketch & R \\
    \midrule
    Swin-T~\cite{liu2021swin}       & 28     & 4.5   & 81.3 & 38.9 & 69.7 & 33.1 & 21.1 & 29.3 & 41.5 \\
    ConvNeXt-T~\cite{cvpr2022convnext}   & 29     & 4.5   & 82.1 & 42.7 & 72.5 & 35.6 & 24.2 & 33.8 & 47.2  \\
    HorNet-T~\cite{nips2022hornet}    & 22      & 4.0   & 82.8 & 43.4 & 72.3 & 37.5 & 26.6 & 34.1 & 46.6 \\
    SLaK-T~\cite{Liu2022SLak}      & 30      & 5.0   & 82.5 & 43.3 & 72.0 & 36.6 & 30.0 & 32.4 & 45.3 \\
    NAT-T~\cite{hassani2023NAT}       & 28      & 4.3   & 83.2 & 44.0 & 72.2 & 37.8 & 33.0 & 31.9 & 44.9 \\
    \gr \ours-T & 24      & 5.0   & 82.8 & {44.7} & 72.9 & 36.9 & 27.7 & 37.0 & 49.0 \\ \midrule
    Swin-S~\cite{liu2021swin}      & 50      &  8.7  & 83.0 & 43.8 & 72.0 & 36.8 & 32.5 & 32.3 & 45.2 \\
    ConvNeXt-S~\cite{cvpr2022convnext}  & 50     & 8.7    & 83.1 & 45.7 & 72.5 & 38.0 & 31.3 & 37.1 & 49.6 \\
    HorNet-S~\cite{nips2022hornet}    &  50    & 8.8    & 84.0 & 47.3 & 73.6 & 39.9 & 36.2 & 36.9 & 49.7 \\
    SLaK-S~\cite{Liu2022SLak}      & 55    & 9.8     & 83.8 & 48.2 & 73.6 & 39.6 & 39.3 & 37.5 & 50.9 \\
    NAT-S~\cite{hassani2023NAT}       &  51    & 7.8    & 83.7 & 46.4 & 73.2 & 39.9 & 37.4 & 34.3 & 47.3 \\
    \gr \ours-S & 50   &  8.7     & 83.7 & {47.8} & 73.8 & 39.3 & 33.5 & 39.8 & 52.8 \\ \midrule
    Swin-B~\cite{liu2021swin}      & 88      & 15.4  & 83.5 & 44.9 & 72.4 & 37.6 & 35.4 & 32.7 & 46.5 \\
    ConvNeXt-B~\cite{cvpr2022convnext}  & 89     &  15.4  & 83.8 & 47.9 & 73.7 & 39.9 & 36.7 & 38.2 & 51.2 \\
    HorNet-B~\cite{nips2022hornet}    & 87     & 15.6   & 84.3 & 48.8 & 73.9 & 41.0 & 39.9 & 38.1 & 51.2 \\
    SLaK-B~\cite{Liu2022SLak}      & 95     & 17.1   & 84.0 & 48.9 & 74.0 & 39.7 & 41.6 & 38.5 & 50.8 \\
    NAT-B~\cite{hassani2023NAT}       & 90     & 13.7   & 84.3 & 48.5 & 74.1 & 40.7 & 41.4 & 36.6 & 49.7 \\
    \gr \ours-B &  87     & 15.4  & 84.4 & {49.0} & 74.2 & 39.7 & 38.1 & 40.1 & 52.7 \\  \midrule
    ConvNeXt-L~\cite{cvpr2022convnext}  & 198    & 34.4   & 84.3 & 49.9 & 74.2 & 40.6 & 41.3 & 40.1 & 53.5 \\
    \gr \ours-L & 186    & 34.7   & 84.8 & {52.2} & 75.0 & 42.1 & 42.9 & 44.5 & 56.5 \\
    \bottomrule
\end{tabular}
}
\vspace{-1em}
\end{table}

\section{Robustness Evaluation}
\label{sec:robustness}

We further evaluate the robustness of our models using the ImageNet out-of-distribution (OOD) benchmarks - ImageNet-V2~\cite{Recht2019imagenetv2}, ImageNet-A~\cite{hendrycks2021imagenetA}, ImageNet-Sketch~\cite{wang2019imagenetSketch}, ImageNet-R~\cite{hendrycks2021imagenetR}, and ObjectNet~\cite{nips19objectnet}. Table~\ref{suptab:robustness} shows our \ours demonstrates superior robustness in comparison to the other models. We specifically select models (HorNet, SLaK, and NAT) having comparable ImageNet-1K accuracies to demonstrate the superior out-of-distribution (OOD) performance of our models compared with those. Nobaly, even when \ours demonstrates lower accuracy on ImageNet-1K than competing models, \ours achieves high OOD scores across various benchmarks.

\section{More Details and Extended Pilot Study}
\label{sec:pilot}

\begin{figure}[t!]
    \centering
    \small
    \includegraphics[width=0.7\textwidth]{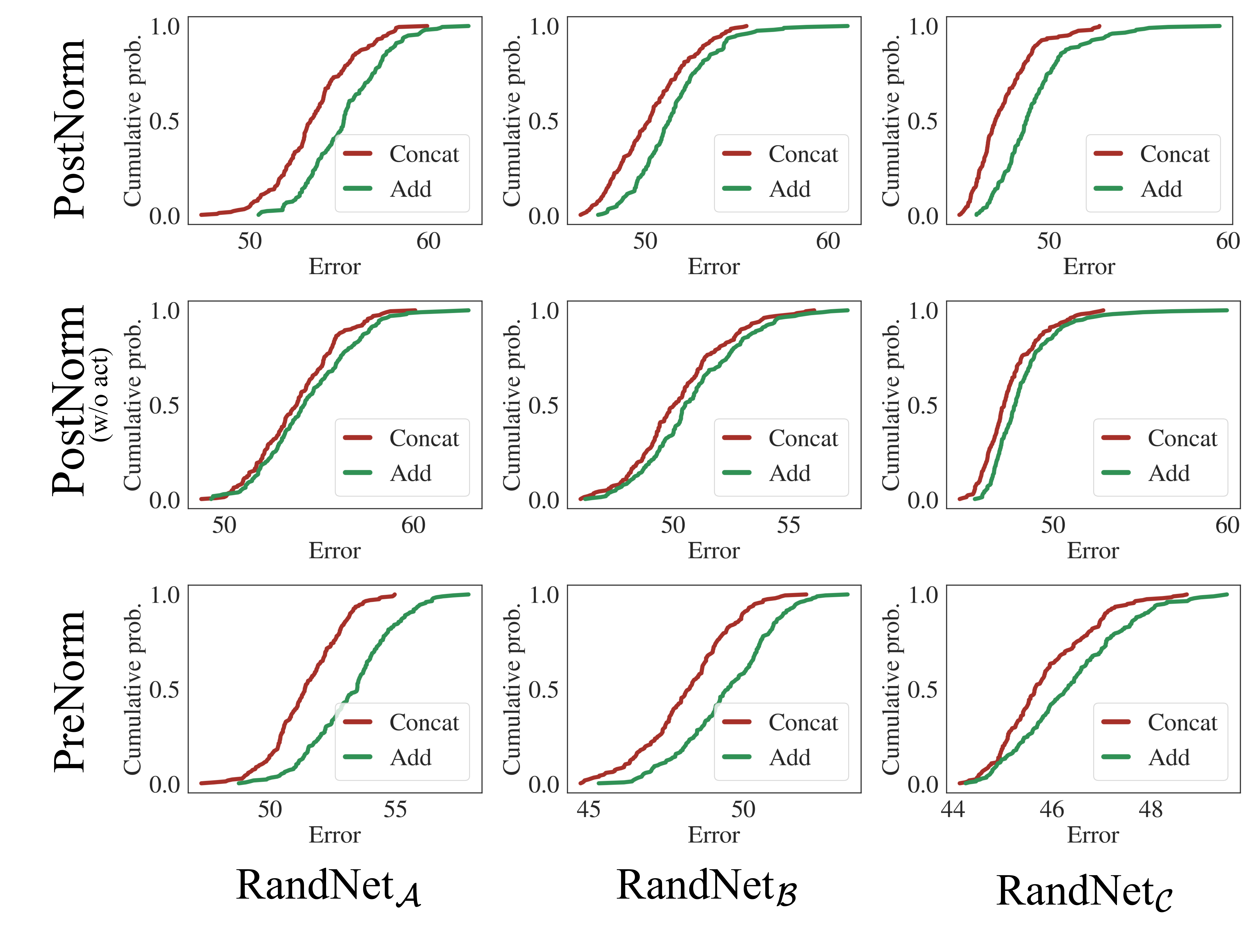} 
    \vspace{-0.5cm}
    \caption{\textbf{Cumulative prabability vs. error} of trained models in Table~\ref{suptab:randnet_results_basic} is visualized here  following  Radosavovic~\etal~\cite{cvpr2020regnet}. Each row demonstrates three block types, and each column exhibits parameter spaces $\mathcal{A}$, $\mathcal{B}$, and \textbf{$\mathcal{C}$}.}
    \label{supfig:randnet_cumulative_prob_basic}
\end{figure}

\begin{figure}[t!]
    \centering
    \small
    \includegraphics[width=\textwidth]{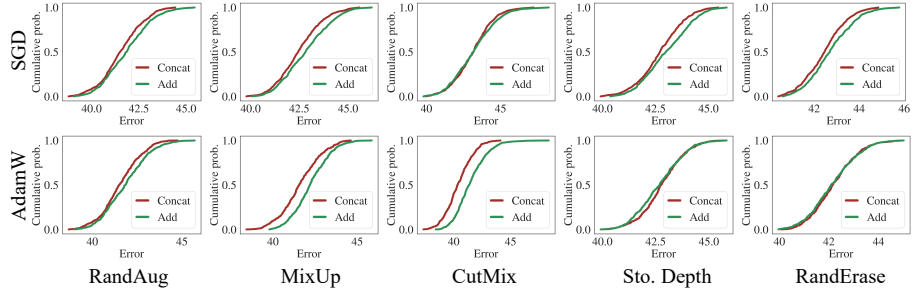}
    \vspace{-0.5cm}
    \caption{\textbf{Cumulative prabability vs. error} of trained models in Table~\ref{suptab:randnet_results_aug} is visualized here following  Radosavovic~\etal~\cite{cvpr2020regnet}. The figure corresponds to the manuscript's parameter spaces $\mathcal{D}$ and $\mathcal{E}$. Each row demonstrates optimizers, and each column exhibits augmentations.}
    \label{supfig:randnet_cumulative_prob_aug}
    \vspace{-1.5em}
\end{figure}

\subsection{More Details of Pilot Study}
We present individual RandNet experimental results performed under controlled setups. We conduct 200 experiments for each configuration of budget, block type, and skip connection type. Fig.~\ref{supfig:randnet_cumulative_prob_basic} and Table~\ref{suptab:randnet_results_basic}, concatenation outperforms addition across parameter spaces $\mathcal{A}$, $\mathcal{B}$, and $\mathcal{C}$. 
For the parameter spaces $\mathcal{D}$ and $\mathcal{E}$ using data augmentations, we use the following hyper-parameters for experiments. For RandAugment~\cite{cubuk2020randaugment}, we limit the magnitude values of \{3, 5, 7\}; for MixUp~\cite{zhang2017mixup} and CutMix~\cite{yun2019cutmix}, we employ alpha values of \{0.1, 0.3, 0.5\} respectively; Stochastic Depth~\cite{eccv2016droppath} ratio are set to \{0.05, 0.1\}; Random Erasing~\cite{zhong2020random} is fixed to 0.25. 

Due to the diverse setups in each data argumentation, we conduct 600 experiments for each element. Additionally, even when switching the optimizer to AdamW, we train 600 random networks for every augmentation as well. Fig.~\ref{supfig:randnet_cumulative_prob_aug} and Table~\ref{suptab:randnet_results_aug} suggest that that data augmentation reduces the performance gap between additive and concatenative shortcuts, particularly noting that the element (\ie, trained using stochastic depth with AdamW) benefit more from an additive shortcut. Nevertheless, we continue to observe that concatenation-based models dominate over advantage over additive shortcuts, even in the AdamW training configurations.

\begin{table}[t]
\centering
\small
\caption{\textbf{Concatenation vs. addition.} The table corresponds to the manuscript's parameter spaces $\mathcal{A}$, $\mathcal{B}$, and $\mathcal{C}$. We sample 200 random networks within each parameter space, ensuring similar computational costs of FLOPs, the number of parameters (Param), and memory consumption (Mem), and individually train them on Tiny-ImageNet. All results are averaged along with the standard deviation (± std). Experimental results with higher accuracy are shaded in \colorbox{baselinecolor}{gray}.}
\vspace{-1em}
\tabcolsep=0.3em
\resizebox{0.85\linewidth}{!}{
\begin{tabular}{lccccc|l}
    \toprule
    Model                       & Skip type  & Block type & FLOPs (G)  & Param (M) & Mem (GB)   & Top-1 (\%)               \\ \midrule
    RandNet$_{\mathcal{A}}$     & Add        & PostNorm  & 2.24±0.14&2.20±0.13&0.66±0.03&45.6±2.3            \\
    \gr RandNet$_{\mathcal{A}}$ & Concat     & PostNorm  & 2.24±0.14&2.25±0.14&0.71±0.05&\textbf{46.9±2.2}    \\ \midrule
    RandNet$_{\mathcal{A}}$     & Add        & PostNorm (w/o act) & 2.24±0.13&2.20±0.13&0.66±0.02&45.1±2.1         \\
    \gr RandNet$_{\mathcal{A}}$ & Concat     & PostNorm (w/o act) & 2.23±0.14&2.24±0.14&0.76±0.06&\textbf{46.8±2.2}    \\ \midrule
    RandNet$_{\mathcal{A}}$     & Add        & PreNorm    & 2.24±0.13&2.2±0.13&0.66±0.02&46.8±1.67         \\
    \gr RandNet$_{\mathcal{A}}$ & Concat     & PreNorm    & 2.25±0.14&2.25±0.14&0.82±0.08&\textbf{48.7±1.6}  \\ \midrule
    RandNet$_{\mathcal{B}}$     & Add        & PostNorm  & 4.51±0.27&4.42±0.26& 0.78±0.05& 49.9±2.2       \\
    \gr RandNet$_{\mathcal{B}}$ & Concat     & PostNorm  & 4.53±0.28&4.52±0.28& 0.88±0.09& \textbf{50.7±2.2}          \\ \midrule
    RandNet$_{\mathcal{B}}$     & Add        & PostNorm (w/o act)   & 4.52±0.27&4.43±0.26& 0.78±0.05& 49.4±2.3         \\
    \gr RandNet$_{\mathcal{B}}$ & Concat     & PostNorm (w/o act)   & 4.51±0.28&4.50±0.28& 0.84±0.08& \textbf{50.7±2.1}    \\ \midrule
    RandNet$_{\mathcal{B}}$     & Add        & PreNorm    & 4.51±0.26&4.42±0.26& 0.78±0.05& 51.1±1.6         \\
    \gr RandNet$_{\mathcal{B}}$ & Concat     & PreNorm    & 4.50±0.30&4.48±0.29& 0.97±0.12& \textbf{52.2±1.5}    \\  \midrule
        RandNet$_{\mathcal{C}}$ & Add        & PostNorm  & 9.58±0.23&9.37±0.23& 1.00±0.09& 52.8±2.2                       \\
    \gr RandNet$_{\mathcal{C}}$ & Concat     & PostNorm  & 9.55±0.27&9.46±0.27& 1.05±0.11& \textbf{53.8±2.1}                  \\ \midrule
        RandNet$_{\mathcal{C}}$ & Add        & PostNorm (w/o act)   & 9.60±0.24&9.39±0.23& 1.02±0.09& 52.5±2.2              \\
    \gr RandNet$_{\mathcal{C}}$ & Concat     & PostNorm (w/o act)   & 9.56±0.27&9.46±0.26& 1.11±0.13& \textbf{53.9±2.1}    \\ \midrule
        RandNet$_{\mathcal{C}}$ & Add        & PreNorm    & 9.58±0.22&9.37±0.21& 1.02±0.10& 54.4±1.6         \\
    \gr RandNet$_{\mathcal{C}}$ & Concat     & PreNorm    & 9.56±0.26&9.46±0.25& 1.24±0.17& \textbf{55.1±1.3}       \\
    \bottomrule
\end{tabular}
}
\label{suptab:randnet_results_basic}
\end{table}

\begin{table}[t!]
\centering
\small
\caption{\textbf{Concatenation vs. addition with data augmentations.} The table corresponds to the manuscript's parameter spaces $\mathcal{D}$ and $\mathcal{E}$. We utilize the PreNorm block for augmentation experiments. We sample 600 random networks due to diverse degrees of data augmentations and report identically in Table~\ref{suptab:randnet_results_basic}. Experimental results with higher accuracy are shaded in \colorbox{baselinecolor}{gray}.}
\label{suptab:randnet_results_aug}
\vspace{-1em}
\tabcolsep=0.3em
\resizebox{0.85\linewidth}{!}{
\begin{tabular}{lcccccc|l}
    \toprule
         Model                       & Skip type  &  Augmentation            & AdamW     & FLOPs (G)  & Param (M) & Mem (GB)   & Top-1 (\%)               \\ \midrule
        RandNet$_{\mathcal{D}}$ & Add     &  RandAug                 &           & 9.60±0.23&9.40±0.23 & 1.02±0.09 & 57.8±1.3    \\ 
    \gr RandNet$_{\mathcal{D}}$ & Concat        &  RandAug                 &           & 9.54±0.26&9.44±0.26 & 1.25±0.17 & \textbf{58.7±1.2}    \\ \midrule
        RandNet$_{\mathcal{D}}$ & Add     &  MixUp                   &           & 9.59±0.23&9.39±0.22 & 1.02±0.09 & 57.3±1.4    \\ 
    \gr RandNet$_{\mathcal{D}}$ & Concat        &  MixUp                   &           & 9.55±0.26&9.45±0.26 & 1.24±0.16 & \textbf{57.9±1.2}  \\ \midrule
        RandNet$_{\mathcal{D}}$ & Add     &  CutMix                  &           & 9.60±0.23&9.39±0.23 & 1.03±0.09 & 57.0±1.7    \\ 
    \gr RandNet$_{\mathcal{D}}$ & Concat        &  CutMix                  &           & 9.55±0.27&9.45±0.26 & 1.23±0.16 & \textbf{57.5±1.5}  \\ \midrule
        RandNet$_{\mathcal{D}}$ & Add     &  Sto. Depth                &           & 9.59±0.23&9.39±0.23 & 1.02±0.09 & 57.3±1.4    \\ 
    \gr RandNet$_{\mathcal{D}}$ & Concat        &  Sto. Depth                &           & 9.55±0.27&9.45±0.26 & 1.24±0.17 & \textbf{57.8±1.2}  \\ \midrule
        RandNet$_{\mathcal{D}}$ & Add     &  RandErase               &           & 9.59±0.23&9.38±0.23 & 1.02±0.09 & 57.8±1.3    \\ 
    \gr RandNet$_{\mathcal{D}}$ & Concat        &  RandErase               &           & 9.56±0.26&9.45±0.26 & 1.24±0.16 & \textbf{58.2±1.2}  \\ \midrule
        RandNet$_{\mathcal{E}}$ & Add     &  RandAug                 & \ding{51} & 9.60±0.23&9.40±0.23 & 1.02±0.09 & 58.5±1.4    \\  
    \gr RandNet$_{\mathcal{E}}$ & Concat        &  RandAug                 & \ding{51} & 9.54±0.26&9.44±0.26 & 1.25±0.17 & \textbf{59.2±1.3}  \\ \midrule 
        RandNet$_{\mathcal{E}}$ & Add     &  MixUp                   & \ding{51} & 9.60±0.23&9.39±0.22 & 1.02±0.09 & 58.2±1.3  \\ 
    \gr RandNet$_{\mathcal{E}}$ & Concat        &  MixUp                   & \ding{51} & 9.55±0.25&9.44±0.25 & 1.24±0.16 & \textbf{58.9±1.3}  \\ \midrule
        RandNet$_{\mathcal{E}}$ & Add     &  CutMix                  & \ding{51} & 9.59±0.23&9.38±0.23 & 1.02±0.09 & 58.9±1.6    \\ 
    \gr RandNet$_{\mathcal{E}}$ & Concat        &  CutMix                  & \ding{51} & 9.54±0.26&9.44±0.26 & 1.25±0.17 & \textbf{60.2±1.4}  \\ \midrule
    \gr RandNet$_{\mathcal{E}}$ & Add     &  Sto. Depth                & \ding{51} & 9.60±0.23&9.39±0.23 & 1.02±0.09 & \textbf{57.5±1.3}  \\ 
        RandNet$_{\mathcal{E}}$ & Concat        &  Sto. Depth                & \ding{51} & 9.55±0.26&9.44±0.26 & 1.25±0.16 & 57.5±1.1    \\ \midrule
    \gr RandNet$_{\mathcal{E}}$ & Add     &  RandErase               & \ding{51} & 9.58±0.23&9.37±0.23 & 1.02±0.08 & \textbf{58.1±1.1}    \\
        RandNet$_{\mathcal{E}}$ & Concat        &  RandErase               & \ding{51} & 9.56±0.26&9.46±0.26 & 1.24±0.18 & 58.1±1.0  \\
    \bottomrule
\end{tabular}
}
\vspace{-2em}
\end{table}

\subsection{Scaled-up Pilot Study}
To further extend our pilot study, we conduct ImageNet-1K experiments across the parameter spaces $\mathcal{C}$, $\mathcal{D}$, and $\mathcal{E}$, as described in \S\ref{sec:potential_concat}. 
For the parameter space $\mathcal{C}$, we train the model for 60 epochs. For $\mathcal{D}$ and $\mathcal{E}$, we extend the training epochs to 90 to assess the impact of augmentation. In the parameter space $\mathcal{C}$, we carry out 150 experiments for each type of skip connection. In $\mathcal{D}$ and $\mathcal{E}$, we conduct 60 experiments for each combination of augmentation and skip connection type. As shown in Table~\ref{suptab:randnet_results_in1k}, even on the ImageNet-1K dataset, the trained models using concatenation statistically surpass those using addition.

\begin{table}[ht!]
\centering
\caption{\textbf{Scaled-up pilot study on ImageNet-1K - concatenation vs. addition experiments.} The table corresponds to our extended pilot study on the same parameter spaces $\mathcal{C}$, $\mathcal{D}$ and $\mathcal{E}$. Each row in parameter space $\mathcal{C}$ contains statistics derived from 150 experiments, while each row in parameter spaces $\mathcal{D}$ and $\mathcal{E}$ encompasses statistics from 300 experiments. We utilize the PreNorm block only for scaled-up experiments. Experimental results with higher accuracy are shaded in \colorbox{baselinecolor}{gray}.}
\label{suptab:randnet_results_in1k}
    \tabcolsep=0.35em
    \resizebox{0.8\linewidth}{!}{
    \begin{tabular}{@{}cc|ccccc@{}}
    \toprule
    Model & Skip type & FLOPs (G) & Param (M) & Mem (GB)   & Top-1 (\%)              \\ \midrule
        RandNet$_{\mathcal{C}}$ & Add       & 9.51±0.28 & 9.41±0.28& 1.02±0.09 & 54.9±2.2            \\
    \gr RandNet$_{\mathcal{C}}$ & Concat    & 9.45±0.22 & 9.47±0.20 & 1.24±0.17 & \textbf{55.1±2.0}    \\ \midrule
        RandNet$_{\mathcal{D}}$ & Add       & 9.60±0.24& 9.46±0.20 & 1.02±0.09& 54.7±2.6            \\
    \gr RandNet$_{\mathcal{D}}$ & Concat    & 9.55±0.26& 9.45±0.26 & 1.25±0.16& \textbf{55.1±2.7}    \\ \midrule
        RandNet$_{\mathcal{E}}$ & Add       & 9.60±0.27& 9.38±0.26& 1.02±0.09& 55.8±2.8            \\
    \gr RandNet$_{\mathcal{E}}$ & Concat    & 9.55±0.27& 9.44±0.26 & 1.25±0.16& \textbf{56.7±2.7}    \\
    \bottomrule
    \end{tabular}
    }
\end{table}

\clearpage

\end{document}